\documentclass{article}
\usepackage{ijcai21}

\usepackage{times}
\usepackage{soul}
\usepackage{url}
\usepackage[hidelinks]{hyperref}
\usepackage[utf8]{inputenc}
\usepackage[small]{caption}
\usepackage{graphicx}
\usepackage{amsmath}
\usepackage{amsthm}
\usepackage{booktabs}
\usepackage{algorithm}
\usepackage{algorithmic}
\usepackage{enumitem} 
\usepackage{subfigure} 
\usepackage{bm} 

\usepackage{multirow,tikz}
\usetikzlibrary{arrows,fit,positioning,shapes,snakes}
\newdimen\arrowsize
\pgfarrowsdeclare{arcsq}{arcsq}
{
	\arrowsize=0.5pt
	\advance\arrowsize by .5\pgflinewidth
	\pgfarrowsleftextend{-4\arrowsize-.5\pgflinewidth}
	\pgfarrowsrightextend{.5\pgflinewidth}
}
{
	\arrowsize=1.5pt
	\advance\arrowsize by .5\pgflinewidth
	\pgfsetdash{}{0pt} 
	\pgfsetroundjoin   
	\pgfsetroundcap    
	\pgfpathmoveto{\pgfpoint{0\arrowsize}{0\arrowsize}}
	\pgfpatharc{-90}{-140}{4\arrowsize}
	\pgfusepathqstroke
	\pgfpathmoveto{\pgfpointorigin}
	\pgfpatharc{90}{140}{4\arrowsize}
	\pgfusepathqstroke
}
\tikzset{snake it/.style={decorate, decoration=snake}}
\tikzstyle{line} = [draw, -latex']

\makeatletter
\newtheorem*{rep@theorem}{\rep@title}
\newcommand{\newreptheorem}[2]{%
\newenvironment{rep#1}[1]{%
 \def\rep@title{#2 \ref{##1}}%
 \begin{rep@theorem}}%
 {\end{rep@theorem}}}
\makeatother

\newtheorem{lemma}{Lemma}
\newtheorem{definition}{Definition}
\newreptheorem{lemma}{Lemma}
\newreptheorem{theorem}{Theorem}
\newreptheorem{definition}{Definition}
\newtheorem{corollary}{Corollary}

\newenvironment{proofsketch}{%
  \proof}{\endproof}

\DeclareMathOperator{\sign}{sign}
\usepackage{amssymb} 

\newtheorem{theorem}{Theorem}

\title{Causal Discovery with Multi-Domain LiNGAM for Latent Factors}

\author{
Yan Zeng$^{1,2}$
\and
Shohei Shimizu$^{2,3}$\and
Ruichu Cai$^{1,4}$\and
Feng Xie$^5$\and \\
Michio Yamamoto$^{2,6}$\And
Zhifeng Hao$^{1,7}$
\affiliations
$^1$Guangdong University of Technology\\
$^2$RIKEN, $^3$Shiga University\\
$^4$Pazhou Lab, $^5$Peking University\\
$^6$Okayama University, $^7$Foshan University
\emails
yanazeng013@gmail.com, shohei-shimizu@biwako.shiga-u.ac.jp, 
\{cairuichu, xiefeng009\}@gmail.com,
m.yamamoto@okayama-u.ac.jp,
zfhao@gdut.edu.cn
}

\begin{document}

\maketitle

\begin{abstract} 
Discovering causal structures among latent factors from observed data is a particularly challenging problem. Despite some efforts for this problem, existing methods focus on the single-domain data only. In this paper, we propose \underline{M}ulti-\underline{D}omain \underline{Li}near \underline{N}on-Gaussian Acyclic Models for L\underline{A}tent Factors (MD-LiNA), where the causal structure among latent factors of interest is shared for all domains, and we provide its identification results. The model enriches the causal representation for multi-domain data.
We propose an integrated two-phase algorithm to estimate the model.
In particular, we first locate the latent factors and estimate the factor loading matrix. Then to uncover the causal structure among shared latent factors of interest, we derive a score function based on the characterization of independence relations between external influences and the dependence relations between multi-domain latent factors and latent factors of interest.
 We show that the proposed method provides locally consistent estimators.
Experimental results on both synthetic and real-world data demonstrate the efficacy and robustness of our approach.
\end{abstract}

\section{Introduction}

Learning causal relationships from observed data, termed as causal discovery, has been developed rapidly over the past decades~\cite{pearl2009causality,spirtes2016causal,peters2017elements}.
In many scenarios, including sociology, psychology, and educational research, the underlying causal relations are usually embedded between latent variables (or factors) that cannot be directly measured, e.g., anxiety, depression, or coping, etc~\cite{silva2006learning,Bartholomewbook}, in which scientists are often interested. 

Some approaches have been developed to identify the causal structure among latent factors, which can be categorized into covariance-based and non-Gaussianity-based ones.
Covariance-based methods employ the covariance structure of data alone, e.g., 
BuildPureClusters
algorithm~\cite{silva2006learning}, or FindOneFactorClusters algorithm~\cite{kummerfeld2016causal}, to ascertain how many latent factors as well as the structure of latent factors.
However, these algorithms can only output structures up to the Markov equivalence class for latent factors.
Non-Gaussianity-based methods address this indistinguishable identification problem by taking the best of the non-Gaussianity of data. 
Specifically,~\citeauthor{shimizu2009estimation}~\shortcite{shimizu2009estimation} leveraged non-Gaussianity and firstly achieved identifying a unique causal structure between latent factors based on the Linear, Non-Gaussian, Acyclic Models (LiNGAM)~\cite{shimizu2006linear}. 
They transformed the problem into the Noisy Independent Component Analysis (NICA). 
Recently, to avoid the local optima of the NICA,~\citeauthor{cai2019triad}~\shortcite{cai2019triad} designed the so-called Triad constraints and~\citeauthor{xie2020generalized}~\shortcite{xie2020generalized} developed the GIN condition. They both proposed a two-phase method to learn the structure or causal orderings among latent factors.

It is noteworthy that the above-mentioned methods all focus on the data which are originated from the same domain, i.e., single-domain data.
However, in many real-world applications, data are often collected under distinct conditions. They may be originated from different domains, resulting in distinct distributions and/or various causal effects. 
For instance, in neuroinformatics, functional
Magnetic Resonance Imaging (fMRI) signals are frequently extracted from multiple subjects or over time~\cite{smith2011network}; in biology, a particular disease is measured by distinct medical equipment~\cite{dhir2020integrating}, 
etc. 
Existing methods to handle multi-domain data in causal discovery are flourishing, e.g.,~\citeauthor{danks2009integrating}~\shortcite{danks2009integrating},
~\citeauthor{tillman2011learning}~\shortcite{tillman2011learning},
~\citeauthor{ghassami2018multi}~\shortcite{ghassami2018multi},~\citeauthor{kocaoglu2019characterization}~\shortcite{kocaoglu2019characterization},~\citeauthor{dhir2020integrating}~\shortcite{dhir2020integrating}, ~\citeauthor{huang2020causal}~\shortcite{huang2020causal},~\citeauthor{Jaber2020soft}~\shortcite{Jaber2020soft}, etc.
Though there are some methods to handle multi-domain data that allow the existence of latent variables or confounders
, no such method is yet proposed in the literature when learning the causal structure among latent factors with only observed data, to our best knowledge. Thus, it is desirable to perform causal discovery from multi-domain data to uncover the structure among latent factors. 

When considering multi-domain instead of single-domain data in latent factor models, there may exist different causal structures among latent factors or with different causal effects in different domains. An important question then naturally raises, i.e.,  how to guarantee that factors in different domains are represented by the same factors of interest so that the underlying structure among latent factors of interest is uncovered.
A solution may be to naively concatenate the multi-domain observed data, such that the multi-domain model can be regarded as a single-domain latent factor model. However, it may cause serious bias in estimating the causal structure among latent factors of interest~\cite{shimizu2012joint}. 
In this paper, we propose \underline{M}ulti-\underline{D}omain \underline{Li}near \underline{N}on-Gaussian Acyclic Models for L\underline{A}tent Factors (MD-LiNA) to represent the causal mechanism of latent factors,
which tackles not only single-domain data but multi-domain ones.
In addition, we propose an integrated two-phase approach to uniquely identify the underlying causal structure among latent factors (of interest).
In particular, in the first phase, we locate the latent factors and estimate the factor loading matrix (relating the observed variables to its latent factors)  for all domains, leveraging the ideas from
Triad constraints and factor analysis~\cite{cai2019triad,reilly1996identification}. 
In the second phase, we derive a score function to characterize the independence relations between external variables, and interestingly, we unify this function to characterize the dependence relations between latent factors from different domains and latent factors of interest. Then such unified function is enforced with acyclicity, sparsity, and elastic net constraints, with which our task is formulated as a purely continuous optimization problem. The method is (locally) consistent to produce feasible solutions. 

Our contributions are mainly three-folded:
\begin{itemize} 
\item This is the first effort to study the causal discovery problem of identifying causal structures between latent factors for multi-domain observed data.
\item We propose an MD-LiNA model, 
which is a causal representation for both single and multi-domain data in latent factor models and offers
a deep interpretation of dependencies between observed variables across domains, and show its identifiability results.
\item We propose an integrated two-phase approach to uniquely estimate the underlying causal structure among latent factors, which simultaneously identifies causal directions and effects. It is capable of handling cases when the sample size is small or the latent factors are highly correlated. And the local consistency is also provided.
\end{itemize}

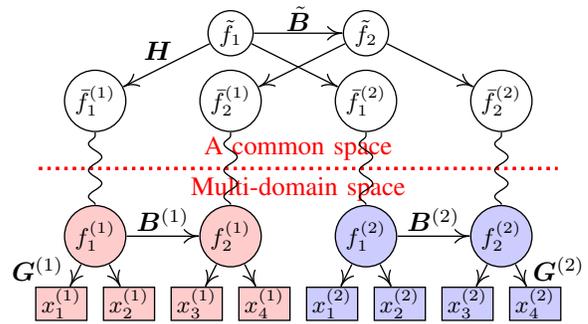
\begin{figure}[t]    
	\centering
    \begin{tikzpicture}[scale=0.6,line width=0.5pt,inner sep=0.2mm, shorten >=.1pt, shorten <=.1pt, mycirc0/.style={fill=blue!20}, mycirc2/.style={fill=red!20}]
        \draw (2.25, 3) node(13) [circle, minimum size=2mm, draw] {{\footnotesize\,$\Tilde{{f}_1}$\,}};
        \draw (5.25, 3) node(14) [circle, minimum size=2mm, draw] {{\footnotesize\,$\Tilde{{f}_2}$\,}};
         \draw (-0.75, 1.5) node(15) [circle, minimum size=4mm, draw] {{\footnotesize\,$\Bar{{f}}_1^{(1)}$\,}};
        \draw (2.25, 1.5) node(16) [circle, minimum size=4mm, draw] {{\footnotesize\,$\Bar{f}_2^{(1)}$\,}};
        \draw (5.25, 1.5) node(17) [circle, minimum size=4mm, draw] {{\footnotesize\,$\Bar{f}_1^{(2)}$\,}};
        \draw (8.25, 1.5) node(18) [circle, minimum size=4mm, draw] {{\footnotesize\,$\Bar{f}_2^{(2)}$\,}};
        \draw (-0.75, -1.5) node(1) [circle, mycirc2, minimum size=5mm, draw] {{\footnotesize\,${f}_1^{(1)}$\,}};
        \draw (2.25, -1.5) node(2) [circle, mycirc2, minimum size=4mm, draw] {{\footnotesize\,${f}_2^{(1)}$\,}};
        \draw (5.25, -1.5) node(11) [circle, mycirc0, minimum size=4mm, draw] {{\footnotesize\,${f}_1^{(2)}$\,}};
        \draw (8.25, -1.5) node(12) [circle, mycirc0, minimum size=4mm, draw] {{\footnotesize\,${f}_2^{(2)}$\,}};

        \draw (4.5, -3) node(3) [rectangle, mycirc0, minimum size=1mm, draw] {{\footnotesize\,${x}_1^{(2)}$\,}};
        \draw (6, -3) node(4) [rectangle, mycirc0, minimum size=1mm, draw] {{\footnotesize\,${x}_2^{(2)}$\,}};
        \draw (7.5, -3) node(5) [rectangle, mycirc0, minimum size=1mm, draw] {{\footnotesize\,${x}_3^{(2)}$\,}};
        \draw (9, -3) node(6) [rectangle, mycirc0, minimum size=1mm, draw] {{\footnotesize\,${x}_4^{(2)}$\,}};;
        \draw (-1.5, -3) node(7) [rectangle, mycirc2, minimum size=1mm, draw] {{\footnotesize\,${x}_1^{(1)}$\,}};
        \draw (0, -3) node(8) [rectangle, mycirc2, minimum size=1mm, draw] {{\footnotesize\,${x}_2^{(1)}$\,}};
        \draw (1.5, -3) node(9) [rectangle, mycirc2, minimum size=1mm, draw] {{\footnotesize\,${x}_3^{(1)}$\,}};
        \draw (3, -3) node(10) [rectangle, mycirc2, minimum size=1mm, draw] {{\footnotesize\,${x}_4^{(1)}$\,}};
        
        \draw [dotted,red,very thick] (-2,0) -- node [text width=5.5cm,above=0.2em,align=center ]  {A common space}  (9.5,0);
        \draw [dotted,red,very thick] (-2,0) -- node [text width=5.5cm,below=0.2em,align=center ]  {Multi-domain space}  (9.5,0);
        \draw[-arcsq] (1) -- (2) node[pos=0.5,sloped,above=1pt] {$\bm{B}^{(1)}$};
        \draw[-arcsq] (11) -- (12) node[pos=0.5,sloped,above=1pt] {$\bm{B}^{(2)}$};
        \draw[-arcsq] (11) -- (3) node[pos=0.5,sloped,above] {};
        \draw[-arcsq] (11) -- (4) node[pos=0.5,sloped,above] {};
        \draw[-arcsq] (1) -- (7) node[pos=0.2,left=5pt] {$\bm{G}^{(1)}$};
        \draw[-arcsq] (1) -- (8) node[pos=0.5,sloped,above] {}; 
        \draw[-arcsq] (12) -- (5) node[pos=0.5,sloped,above] {};
        \draw[-arcsq] (12) -- (6) node[pos=0.2,right=5pt] {$\bm{G}^{(2)}$}; 
        \draw[-arcsq] (2) -- (9) node[pos=0.5,sloped,above] {};
        \draw[-arcsq] (2) -- (10) node[pos=0.5,sloped,above] {};
        \draw[-arcsq] (13) -- (14) node[pos=0.5,sloped,above=0pt] {$\tilde{\bm{B}}$}; 
        \draw[-arcsq] (13) -- (15) node[pos=0.6,above=3pt] {$\bm{H}$};
        \draw[-arcsq] (13) -- (17) node[pos=0.5,sloped,above] {}; 
        \draw[-arcsq] (14) -- (16) node[pos=0.5,sloped,above] {};
        \draw[-arcsq] (14) -- (18) node[pos=0.5,sloped,above] {};
        \draw[snake it] (1) -- (15) node[pos=0.5,sloped,above] {};
        \draw[snake it] (2) -- (16) node[pos=0.5,sloped,above] {}; 
        \draw[snake it] (11) -- (17) node[pos=0.5,sloped,above] {};
        \draw[snake it] (12) -- (18) node[pos=0.5,sloped,above] {};
	\end{tikzpicture} 
	\caption{An MD-LiNA model. Variables in the same color (light red and light blue) are in the same domain. Observed variables $\bm{x}^{(m)}$ in domain $m$ entail its latent factors $\bm{f}^{(m)}$. Augmented latent factors $\Bar{\bm{f}}$ are obtained using the coding representation method, which are signified by the curved waved lines\protect\footnotemark. $\Tilde{\bm{f}}$ are shared latent factors of interest, whose structure is shared by ${\bm{f}}$ from different domains.}
	\label{fig:MD-LiNA}
\end{figure}
\footnotetext{These waved lines are used to emphasize the common space and they can be simply replaced by directed edges.}

\section{Problem Formalization} 
Suppose we have data from $M$ domains. Let $\bm{x}^{(m)}$ and $\bm{f}^{(m)}$ ($m=1,...,M$) be the random vectors that collect $p_m$ observed variables and $q_m$ latent factors in domain $m$, respectively.
The total number of observed variables for all domains is $p=\sum_{m=1}^M p_m$ while that of latent factors is $q=\sum_{m=1}^M q_m$. In this study, we focus on linear models,
\begin{equation}\label{eq:MD_LiNA}
\begin{aligned}
\bm{f}^{(m)}&=\bm{B}^{(m)}\bm{f}^{(m)}+\bm{\varepsilon}^{(m)},\\
\bm{x}^{(m)}&=\bm{G}^{(m)}\bm{f}^{(m)}+\bm{e}^{(m)},
\end{aligned}
\end{equation}
where $\bm{\varepsilon}^{(m)}$, and $\bm{e}^{(m)}$ are random vectors that collect external influences, and errors, respectively, and they are independent with each other. $\bm{B}^{(m)}$ is a matrix that collects causal effects $b_{ij}^{(m)}$ between ${f}_i^{(m)}$ and ${f}_j^{(m)}$, while $\bm{G}^{(m)}$ collects factor loadings $g_{ij}^{(m)}$ between ${f}_j^{(m)}$ and ${x}_i^{(m)}$. ${f}_i^{(m)}$ is assumed to have zero means and unit variances. Note that in a specific domain $m$, data are generated with the same $b_{ij}^{(m)}$ and $g_{ij}^{(m)}$. ${\varepsilon}_i^{(m)}$ and ${e}_i^{(m)}$ are sampled independently from the identical distributions. In different domains, $\bm{B}^{(m_1)}$ ($\bm{G}^{(m_1)}$) and $\bm{B}^{(m_2)}$ ($\bm{G}^{(m_2)}$) may be different, but they have a shared causal structure.

Thereafter, we encounter two problems: how to integrate the multi-domain data effectively
; and how to guarantee factors $\bm{f}$ in different domains are represented by the same concepts (factors) of interest, with which we identify the underlying causal structure among latent factors of interest. 
To address the first problem, we leverage an idea of simple coding representation~\cite{shimodaira2016cross}, i.e., an observation is represented as an augmented one with $p$ dimensions, where only $p_m$ dimensions come from the original domain while the other $(p - p_m)$ dimensions are padded with zeros. Any augmented data are expressed with a bar, e.g., $\Bar{\bm{x}}$ and $\Bar{\bm{f}}$. With such representations, we obtain, 
\begin{equation} \label{eq:augment}
\begin{aligned}
\bar{\bm{f}}&=\bar{\bm{B}}\bar{\bm{f}}+\bar{\bm{\varepsilon}},\\
\bar{\bm{x}}&=\bar{\bm{G}} \bar{\bm{f}}+ \bar{\bm{e}},
\end{aligned}
\end{equation}
where $\bar{\bm{B}}=\mathrm{Diag} (\bm{B}^{(1)},...,\bm{B}^{(M)})$, $\bar{\bm{\varepsilon}}$ and $\bar{\bm{e}}$ are independent, and $\bar{\bm{G}}= \mathrm{Diag}( \bm{G}^{(1)},...,\bm{G}^{(M)})$.
A detailed explanation is presented in Supplementary Materials A (SM A). To address the second problem, we introduce factors of interest $\Tilde{\bm{f}}$, which are embedded as different concepts with causal relations. As depicted in Figure~\ref{fig:MD-LiNA}, suppose $\bar{\bm{f}}$ are linearly generated by $\Tilde{\bm{f}}$, 
\begin{gather} \label{eq:fandf}
\bar{\bm{f}}= \bm{H}\tilde{\bm{f}},
\end{gather}
where $\bm{H} \in \mathbb{R}^{{q}\times \tilde{q}} (\Tilde{q} \leq q$) is a transformation matrix, and $\Tilde{q}$ is the number of $\Tilde{\bm{f}}$. The whole model is defined as follows.
\begin{definition}[\underline{M}ulti-\underline{D}omain \underline{LiN}GAM for L\underline{A}tent Factors (MD-LiNA)] \label{def:lina}
    An MD-LiNA model satisfies the following assumptions:
    \begin{itemize}[noitemsep,topsep=-3pt,leftmargin=20pt]
    \item [A1.] $\bm{f}^{(m)}$ are generated linearly from a Directed Acyclic Graph (DAG) with non-Gaussian distributed external variables $\bm{\varepsilon}^{(m)}$, as in~Eq.\eqref{eq:MD_LiNA};
    \item [A2.] $\bm{x}^{(m)}$ are generated linearly from $\bm{f}^{(m)}$ plus Gaussian distributed errors $\bm{e}^{(m)}$, as in~Eq.\eqref{eq:MD_LiNA};
    \item [A3.] Each ${f}_i$ has at least 2 pure measurement variables~\footnote{Pure measurement variables are those which have only one single latent factor parent~\cite{silva2006learning}.};
    \item [A4.] Each $\Bar{{f}_i}^{(m)}$ is linearly generated by only one latent in $\Tilde{\bm{f}}$ and each $\Tilde{{f}}_i$ generates at least one latent in $\Bar{\bm{f}}$, as in~Eq.\eqref{eq:fandf}.
\end{itemize}
\end{definition}
Note that assumption A4 implies that each row of $\bm{H}$ has only one non-zero element and each column has at least one non-zero element. It is reasonable since it is equal to the most interpretable structure in factor analysis. More plausibility of assumptions is discussed in SM B. 

Once given multi-domain data, our goal is to estimate $\bar{\bm{G}}$ and $\tilde{\bm{B}}$, where $\tilde{\bm{B}}$ is a matrix that collects causal effects between shared latent factors of interest $\Tilde{\bm{f}}$. $\tilde{\bm{B}}$ reflects the underlying causal structure shared by different $\bm{B}^{(m)}$. If there is only one single domain in the data, one just needs to estimate $\bar{\bm{G}}=\bm{G}^{(1)}$ and ${\bm{B}}^{(1)}$, where assumption A4 can be neglected and our model is simply called \underline{LiN}GAM for L\underline{A}tent Factors (LiNA).
For simplicity, we use the \textbf{measurement model} to relate the structure from $\bm{x}^{(m)}$ to $\bm{f}^{(m)}$ while the \textbf{structure model} to record the causal relations among $\tilde{\bm{f}}$ or $\bm{f}^{(m)}$~\cite{silva2006learning}.

\section{Model Identification}
We state our identifiability results here. Note that the identifiability of LiNA has been provided by~\citeauthor{shimizu2009estimation}~\shortcite{shimizu2009estimation}, but with the assumption that each latent factor has at least 3 pure measurement variables. Below we show this identifiability can be strengthened to 2 pure measurement variables for each factor inspired by Triad constraints~\cite{cai2019triad}.
\begin{lemma} \label{theo:triad} 
    Assume that the input data $\bm{X}$ strictly follow the LiNA model.
    Then the factor loading matrix ${\bm{G}}$ is identifiable up to permutation and scaling of columns and the causal effects matrix ${{\bm{B}}}$ is fully identifiable.
\end{lemma}
The proof is in SM C. It relies on the corollary that Triad constraints also hold for our models, which helps find pure measurement variables and achieve LiNA's identifiability.
Next, we show the identifiability of MD-LiNA in Theorem~\ref{theo:MD-LiNA}.


\begin{theorem} \label{theo:MD-LiNA}
    Assume that the input multi-domain data $\bm{X}$ with ${\bm{X}}^{(m)}$ of domain $m$, strictly follow the MD-LiNA model.
	Then the underlying factor loading matrix $\bar{\bm{G}}$ is identifiable up to permutation and scaling of columns and the causal effects matrix ${\tilde{\bm{B}}}$ is fully identifiable.
\end{theorem}

We give a sketch of the proof below. For complete proofs of all theoretical results, please see SM C.
\begin{proofsketch}
Firstly in~Eq.\eqref{eq:augment} and due to Lemma 1, $\bar{\bm{G}}$ is identifiable up to permutation and scaling of columns, since we estimate one measurement model for all domains simultaneously (mentioned in Section 4).
Furthermore, combining~Eqs.\eqref{eq:augment} and~\eqref{eq:fandf}, we obtain $\Tilde{\bm{f}}=\Tilde{\bm{B}} \Tilde{\bm{f}}+\Tilde{\bm{\varepsilon}}$, where $\Tilde{\bm{B}} = (\bm{H}^T\bm{H})^{-1}\bm{H}^T\bar{\bm{B}}\bm{H} \in \mathbb{R}^{\tilde{q} \times \tilde{q}}$ and $\Tilde{\bm{\varepsilon}}= (\bm{H}^T\bm{H})^{-1}\bm{H}^T\bar{\bm{\varepsilon}}\in \mathbb{R}^{\tilde{q}}$. Note that the inverse matrix of $\bm{H}^T\bm{H}$ always exists since $\bm{H}$ is full column rank due to the assumption A4. To prove $\Tilde{\bm{B}}$ is identifiable, we have to additionally ensure $\Tilde{\bm{B}}$ can be permuted to a strictly lower triangular matrix and $\Tilde{\bm{\varepsilon}}$ are independent with each other. Fortunately, due to assumption A1, $\Tilde{\bm{B}}$ satisfies the condition. Due to assumption A4, by virtue of the independence between $\bar{\bm{\varepsilon}}$, its non-Gaussianity and the Darmois-Skitovich theorem~\cite{kagan1973characterization}, $\Tilde{\bm{\varepsilon}}$ are also independent with each other. Thus, ${\tilde{\bm{B}}}$ is fully identifiable, which implies the theorem is proved.
\end{proofsketch}

\section{Model Estimation}
We exhibit a two-phase framework (measurement-model and structure-model phases) to estimate causal structures under latent factors in Algorithm~\ref{alg:Framwork_b}, and provide its consistency.

\subsection{MD-LiNA Algorithm}

To learn measurement models, we have several approaches.
Firstly, we can use the Confirmatory Factor Analysis (CFA) \cite{reilly1996identification}, after employing Triad\footnote{Triad constraints help locate latent factors and learn the causal structure between them, but they focus on single-domain data.} to yield the structure between latent factors $\bm{f}^{(m)}$ and observed variables $\bm{x}^{(m)}$~\cite{cai2019triad}.
Secondly, more exploratory approaches are advocated, e.g., Exploratory Structural Equation Modeling (ESEM)~\cite{asparouhov2009exploratory}, which enables us to use fewer restrictions on estimating factor loadings. Please see SM D for details. In our paper, we take the first approach, as illustrated in lines 1 to 3 of Algorithm~\ref{alg:Framwork_b}, but we can use the second one as well in our framework. 

To learn structure models, we introduce the log-likelihood function of LiNA, then unify it to MD-LiNA.
For brevity, we omit the superscripts of all notations for LiNA.
The log-likelihood function of LiNA is derived by characterizing the independence relations between $\bm{\varepsilon}$ from NICA models,
\begin{equation} \label{eq:likelihood_nor1}
\begin{aligned}
\mathcal{L}(\bm{B},\hat{\bm{G}}) &= \sum_{t=1}^n \bigg [ \frac{1}{2} \left \| \bm{X}(t)-\hat{\bm{G}}\Tilde{\hat{\bm{G}}}^T\bm{X}(t)\right \| ^2_{\Sigma^{-1}} \\ 
&+ \sum_{i=1}^{q} \log \hat{p}_i(\bm{g}_i^T\bm{X}(t)-\bm{b}_i^T\Tilde{\hat{\bm{G}}}^T\bm{X}(t)) \bigg ] + C,
\end{aligned}
\end{equation}
where $\left \| \bm{x} \right \|^2_{\Sigma^{-1}} = \bm{x}^T \Sigma^{-1}\bm{x} $, $\bm{X}(t)$ is the $t^{th}$ column (observation) of data $\bm{X}$. $\hat{\bm{G}}$ is the estimate of $\bm{G}$. $\Tilde{\hat{\bm{G}}}^T = (\hat{\bm{G}}^T \hat{\bm{G}})^{-1}\hat{\bm{G}}^{T}$ relates to $\hat{\bm{G}}$ and $\bm{g}_i$ is the $i^{th}$ column of $\Tilde{\hat{\bm{G}}}$. The inverse matrix of $\hat{\bm{G}}^T \hat{\bm{G}}$ always exists due to assumption A3. $\bm{b}_i$ denotes the $i^{th}$ column of $\bm{B}^T$. $n$ is the sample size. $C$ is a constant and $\hat{p}_i$ is their corresponding density function, which is specified to be Laplace distribution in estimation. Please see SM E.1 for detailed derivations.

\begin{algorithm}[tb]
\caption{MD-LiNA Algorithm} 
\label{alg:Framwork_b} 
\begin{algorithmic}[1] 
\REQUIRE 
Data $\bm{X}^{(1)},...,\bm{X}^{(M)}$; $M$.
\ENSURE Factor loadings $\bar{\bm{G}}$ ($\bm{G}$); effects matrix ${\tilde{\bm{B}}}$ (${\bm{B}}$).\\
{\it{Phase I: Measurement models}}
\STATE Find the number of latent factors $q_m$ and locate latent factors for each domain $m$ by Triad constraints;
\STATE Get augmented data $\bar{\bm{X}}=\mathrm{Diag}({\bm{X}}^{(1)},...,{\bm{X}}^{(M)})$ and estimate $\bar{\bm{G}}$ by CFA;
\STATE Estimate $\bar{\bm{f}} \doteq (\bar{\bm{G}}^{T} \bar{\bm{G}})^{-1}\bar{\bm{G}}^{T} \bar{\bm{X}}$;\label{algo:f}\\
{\it{Phase II: Structure models}}
\IF{$M>1$}
    \STATE Optimize~Eq.\eqref{eq:optimal_multi} iteratively for ${\bm{H}}$ and ${\tilde{\bm{B}}}$ until convergence using QPM (or ALM);\label{algo:G_b}
    \STATE Update ${\tilde{\bm{B}}}$ with regard to ${\bm{H}}$; \label{algo:H}
\ELSE
    \STATE Optimize~Eq.\eqref{eq:optimal_elastic} to get ${{\bm{B}}}$ with QPM (or ALM);
\ENDIF
\RETURN $\bar{\bm{G}}$ and ${\tilde{\bm{B}}}$ ($M>1$ ); or ${\bm{G}}=\bar{\bm{G}}$ and ${{\bm{B}}}$ ($M=1$).
\end{algorithmic}
\end{algorithm}
Further, to strengthen the learning power in different cases, e.g., with small sample sizes or multicollinearity problem, 
we render $\bm{B}$ to satisfy an acyclicity constraint, adaptive $\ell_1$ as well as $\ell_2$ regularizations~\cite{zheng2018dags,hyvarinen2010estimation,zou2005regularization},
\begin{equation} 
\label{eq:optimal_elastic}
\begin{aligned}
\min_{\bm{B}} \  \mathcal{F}(\bm{B},\hat{\bm{G}})  & , \quad \mathrm{s.t.}\quad h(\bm{B}) = 0, \\
\mathrm{where} \quad \mathcal{F}(\bm{B},\hat{\bm{G}}) = -\mathcal{L}  & (\bm{B},\hat{\bm{G}}) + \lambda_1 \Arrowvert \bm{B}\Arrowvert_{1*} + \lambda_2 \Arrowvert \bm{B}\Arrowvert^2,
\end{aligned}
\end{equation}
$h(\bm{B})=\mathrm{tr}(e^{\bm{B} \circ \bm{B}}) - q$ is the needed acyclicity constraint. 
$\circ$ is the Hadamard product, and $e^{\bm{B}}$ is the matrix exponential of $\bm{B}$. $\Arrowvert \bm{B}\Arrowvert_{1*} = \sum_{i=1}^q \sum_{j=1}^q \lvert b_{ij} \rvert/\lvert \hat{b}_{ij} \rvert$ represents the sparsity constraint where $\hat{b}_{ij}$ in $\hat{\bm{B}}$ is estimated by maximizing $\mathcal{L}(\bm{B},\hat{\bm{G}})$. $\Arrowvert \bm{B}\Arrowvert^2$ is the $\ell_2$ regularization. $\lambda_1$ and $\lambda_2$ are regularization parameters. This optimization function facilitates the simultaneous estimation of causal directions and effects between latent factors, without additional steps of permutation and rescaling, as required in~\citeauthor{shimizu2009estimation}~\shortcite{shimizu2009estimation}.

Thus, with estimated $\hat{\bm{G}}$, we leverage the Quadratic Penalty Method (QPM) (or Augmented Lagrangian Method, ALM) to optimize ${\bm{B}}$, transforming Eq.~\eqref{eq:optimal_elastic} into an unconstrained one,
\begin{equation} 
\label{eq:unconstrained}
\begin{aligned}
\min_{\bm{B}} \  \mathcal{S}(\bm{B}),
\end{aligned}
\end{equation}
in which $\mathcal{S}(\bm{B}) = \mathcal{F}(\bm{B},\hat{\bm{G}}) + \frac{\rho}{2}h(\bm{B})^2$ is the quadratic penalty function. $\rho$ is a regularization parameter. (For ALM, $\mathcal{S}(\bm{B}) = \mathcal{F}(\bm{B},\hat{\bm{G}}) + \frac{\rho}{2}h(\bm{B})^2 + \alpha h(\bm{B})$, where $\alpha$ is a Lagrange multiplier.) Then~Eq.(\ref{eq:unconstrained}) is solved by L-BFGS-B~\cite{zhu1997algorithm}. In case of avoiding numerical false discoveries from estimation, edges whose estimated effects are under a small threshold $\epsilon$ are ruled out~\cite{zheng2018dags}.

Next, we show how Eq.\eqref{eq:optimal_elastic} is unified to handle multi-domain cases ($M>1$). 
Firstly, all ${\bm{X}}^{(m)}$ are projected into $\bar {\bm{X}}$ in a single common space so that $\bar {\bm{f}}$ are estimated. We then introduce the dependence relations between $\tilde {\bm{f}}$ and $\bar{\bm{f}}$ in~Eq.\eqref{eq:optimal_elastic}, i.e.,
reconstruction errors $\mathcal{E}(\bm{H})$ and an adaptive sparsity $ \Arrowvert \bm{H}\Arrowvert_{1*}$,
\begin{equation} 
\label{eq:optimal_multi}
\begin{aligned}
\min_{\tilde{\bm{B}},\bm{H}} \  \bar{\mathcal{F}}(\tilde{\bm{B}},\bm{H})  & , \quad \mathrm{s.t.}\quad h(\tilde{\bm{B}}) = 0, \\
\mathrm{where} \quad \bar{\mathcal{F}}(\tilde{\bm{B}},\bm{H}) &= \mathcal{F}(\tilde{\bm{B}},{\bm{H}})
+ \mathcal{E}(\bm{H}) + \lambda_3 \Arrowvert \bm{H}\Arrowvert_{1*},
\end{aligned}
\end{equation}
$\mathcal{E}(\bm{H}) = \Arrowvert \bar {\bm{f}}-\bm{H}\tilde{\bm{f}} \Arrowvert^2 
= \Arrowvert \bar {\bm{f}}-\bm{P}_{\bm{H}}\bar{\bm{f}} \Arrowvert^2 $, and $\lambda_3$ is a regularization parameter. $\bm{P}_{\bm{H}}=\bm{H}(\bm{H}^T\bm{H})^{-1}\bm{H}^T$ is a projection matrix onto the column space of $\bm{H}$. $\mathcal{F}(\tilde{\bm{B}},{\bm{H}})$ is the log-likelihood of MD-LiNA,
\begin{equation} 
\label{eq:optimal_multi_log}
\begin{aligned}
&\mathcal{F}(\tilde{\bm{B}},{\bm{H}}) = -\mathcal{L} (\tilde{\bm{B}},{\bm{H}}) + \lambda_1 \Arrowvert \tilde{\bm{B}} \Arrowvert_{1*} + \lambda_2 \Arrowvert \tilde{\bm{B}} \Arrowvert^2,\\
&=- \sum_{t=1}^n \bigg [ \frac{1}{2} \left \| \bar {\bm{X}}(t)-\bar{\bm{G}}\Tilde{\bar{\bm{G}}}^{T}\bar {\bm{X}}(t)\right \| ^2_{\Sigma^{-1}} \\ 
&+ \sum_{i=1}^{\tilde{q}} \log \hat{p}_i(\bm{h}_i^T\bar{\bm{f}}(t)-\tilde{\bm{b}}_i^T\Tilde{\hat{\bm{H}}}^T\bar{\bm{f}}(t)) \bigg ] - C \\
&+ \lambda_1 \Arrowvert \tilde{\bm{B}} \Arrowvert_{1*} + \lambda_2 \Arrowvert \tilde{\bm{B}} \Arrowvert^2,
\end{aligned}
\end{equation}
where $\Tilde{\bar{\bm{G}}}^{T} = (\bar{\bm{G}}^{T} \bar{\bm{G}})^{-1}\bar{\bm{G}}^{T}$, $\Tilde{\hat{\bm{H}}}^T = ({\bm{H}}^T {\bm{H}})^{-1}{\bm{H}}^{T}$, the inverse matrices of ${\bm{H}}^T{\bm{H}}$ and $\bar{\bm{G}}^{T} \bar{\bm{G}}$ always exist due to assumptions A3 and A4. $\bm{h}_i$ is the $i^{th}$ column of $\Tilde{\hat{\bm{H}}}$. $\tilde{\bm{b}}_i$ denotes the $i^{th}$ column of $\tilde{\bm{B}}^T$. See SM E.2 for detailed derivations.

We iteratively optimize $\bm{H}$ and $\tilde{\bm{B}}$ using QPM (or ALM) until convergence. Specifically for QPM, i) to optimize $\bm{H}$, we find its descent direction to derive the next iteration for a given ${\tilde{\bm{B}}}$. Since there is no constraints for $\bm{H}$, it is an unconstrained problem. 
ii) to optimize ${\tilde{\bm{B}}}$, we compute its descent direction for the next iteration given ${{\bm{H}}}$, and update the penalty parameter $\rho$ until the acyclicity constraint $h(\tilde{\bm{B}})=0$ is satisfied. 
Finally, we repeat steps i) and ii) until ${\bm{H}}$ and ${\tilde{\bm{B}}}$ are convergent. With ${\bm{H}}$, we can link the factors from multiple domains with high weights together to symbolize the same concepts so that we can decide which factors from different domains are represented by which factors of interest. Specifically, for $\tilde{f}_i$, those factors $\bar{\bm{f}}$ from different domains with the largest weights are considered to be represented by this $\tilde{f}_i$, where $\tilde{f}_i$ can also be named according to its corresponding observed measurement variables. Then according to these represented factors $\bar{\bm{f}}$ for each $\tilde{f}_i$, we obtain the causal ordering among $\tilde{\bm{f}}$, with which ${\tilde{\bm{B}}}$ can be updated. For the computational complexity, please see SM F.

\subsection{Consistency Proofs}
Here we prove that our methods could provide locally consistent estimators for MD-LiNA, including LiNA.
\begin{theorem}\label{theo:consistent1}
    Assume the input single-domain data $\bm{X}$ strictly follow the LiNA model. Given that the sample size $n$, the number of observed variables $p$ and the penalty coefficient $\rho$ satisfy $n,p,\rho \to \infty$, then under conditions given in C0 \& C1 (see SM C.3), our method using QPM with
    ~Eq.\eqref{eq:optimal_elastic}, is consistent and locally consistent to learn ${{\bm{G}}}$ and ${{\bm{B}}}$, respectively. 
\end{theorem}

\begin{theorem}\label{theo:consistent2} 
    Assume the input multi-domain data $\bm{X}$ with ${\bm{X}}^{(m)}$ strictly follow the MD-LiNA model.
    Given that the sample size $n_m$, the number of observed variables $p_m$ of each domain $m$ and the penalty coefficient $\rho$ satisfy
    $n_m, p_m, \rho \to \infty$, then under conditions given in C0-C5 (see SM C.4),
    our method using QPM with
    ~Eq.\eqref{eq:optimal_multi}, is consistent to learn $\bar{\bm{G}}$ and ${{\bm{H}}}$, and locally consistent to learn ${\tilde{\bm{B}}}$.  
\end{theorem}

Having the identification results of LiNA/MD-LiNA, we present Theorems~\ref{theo:consistent1} and~\ref{theo:consistent2} to ensure that with our methods, asymptotically the resulting estimators $\bar{\bm{G}}$ and $\bm{H}$ will be consistent to the true ones, while ${\tilde{\bm{B}}}$ will be locally consistent to its DAG solution.
The proofs are shown in SM C.3 and C.4.

\section{Experiments}
We performed experiments on synthetic and real data,
including multi-domain and single-domain ones. Due to the unstable performance of Triad, we assume the structure in measurement models is known a priori for all methods.

\subsection{Synthetic Data}
We generated the data according to~Eq.(\ref{eq:MD_LiNA}). Unless specified, each structure in a domain has 5 latent factors and each factor has 2 pure measurement variables with sample size 1000. See SM G.1 for the details. We compared our LiNA with NICA~\cite{shimizu2009estimation} and Triad~\cite{cai2019triad} for single-domain data. Since NICA and Triad do not focus on multi-domain data, we used our method that did not conduct line~\ref{algo:H} of Algorithm 1 as the comparison (MD*).
We did experiments with \textbf{i) different sample sizes}; \textbf{ii) highly-correlated latent variables}; \textbf{iii) different numbers of latent factors} and \textbf{iv) multi-domain data}. For other robustness performances, please see SM G.2.

\begin{figure}[t!]
\centering
\subfigure[Re. with Sa.]{
\includegraphics[width=0.144\textwidth]{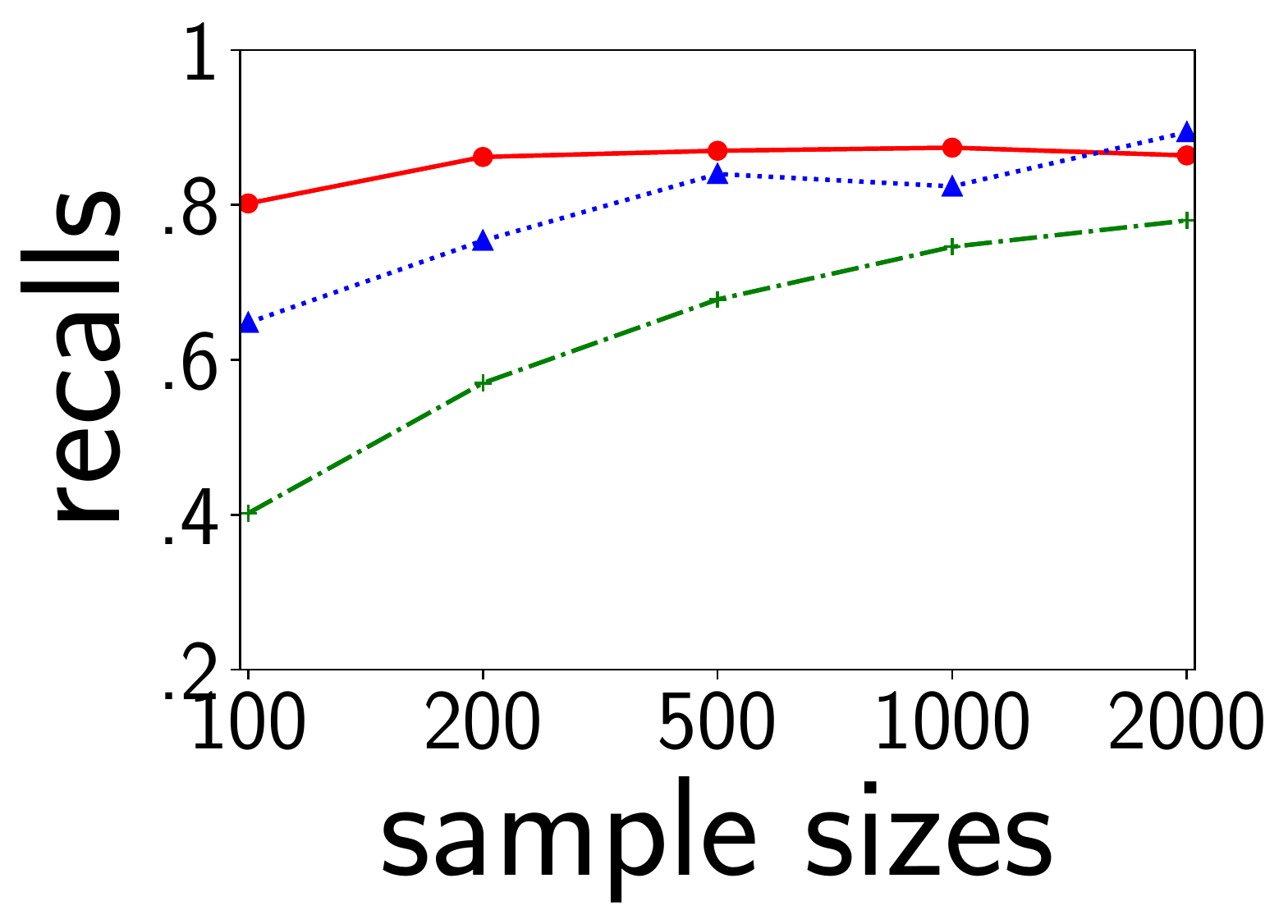}} 
\subfigure[Pre. with Sa.]{
\includegraphics[width=0.144\textwidth]{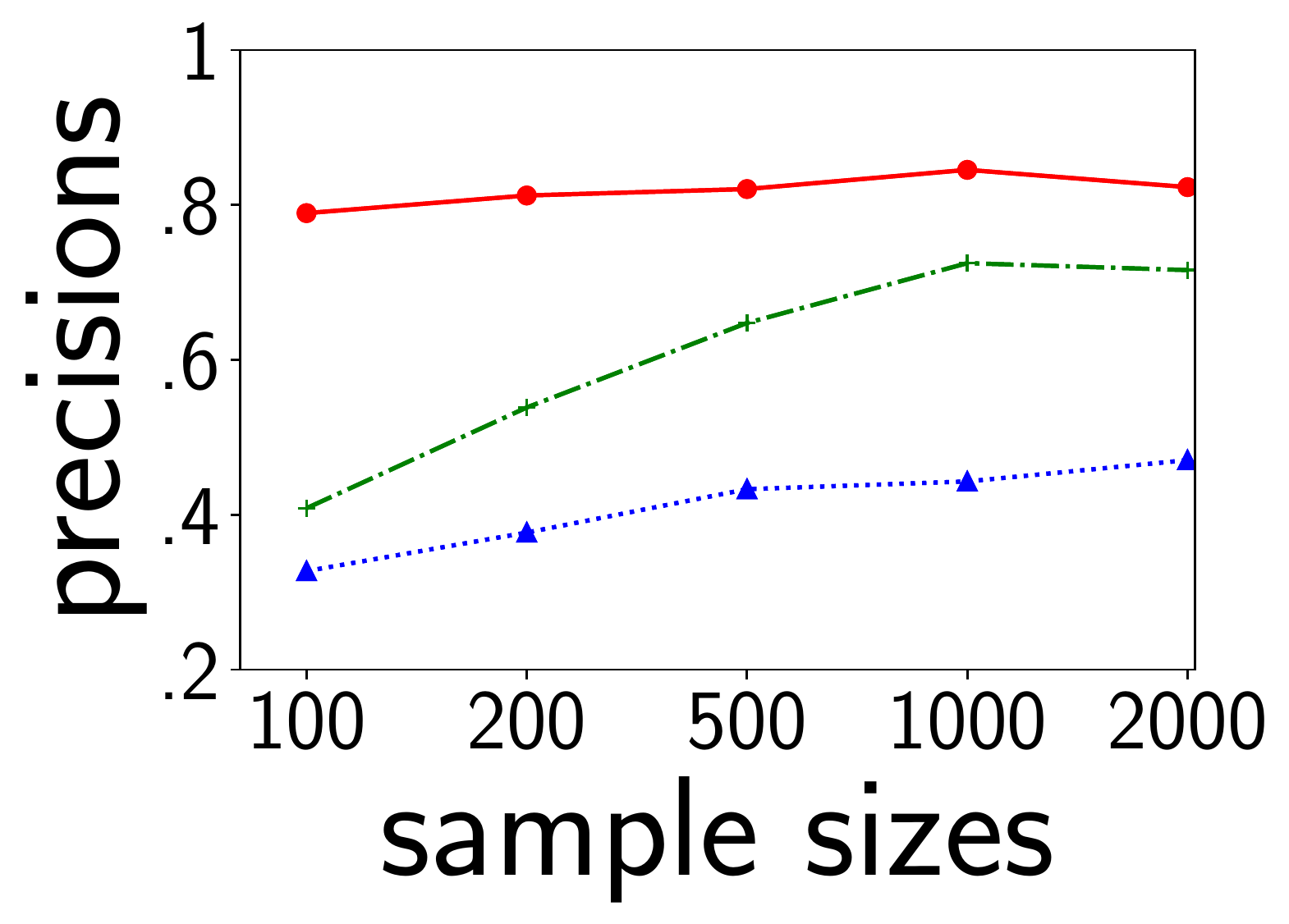}}
\subfigure[F1. with Sa.]{
\includegraphics[width=0.17\textwidth]{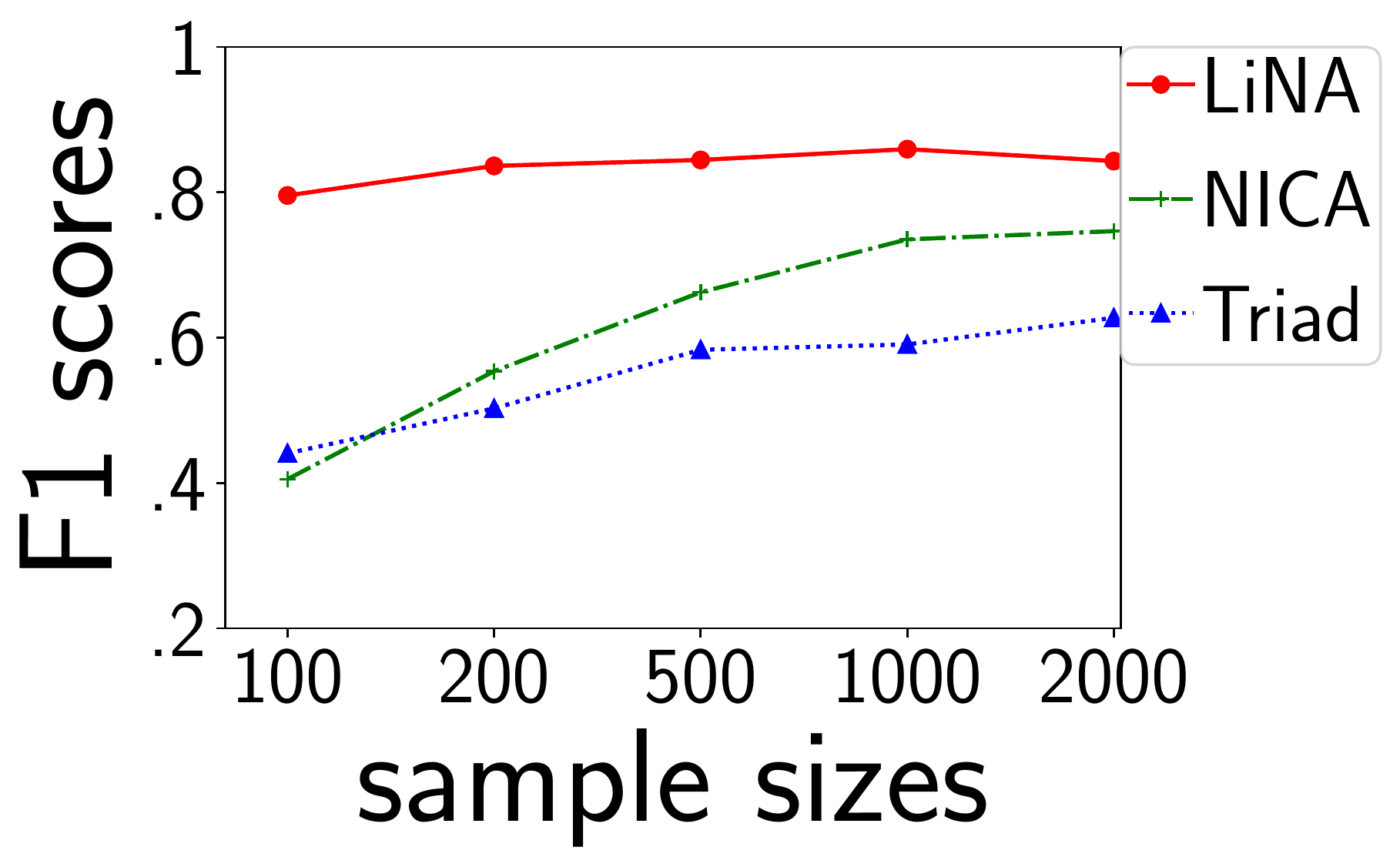}}
\subfigure[Re. with Co.]{
\includegraphics[width=0.14\textwidth]{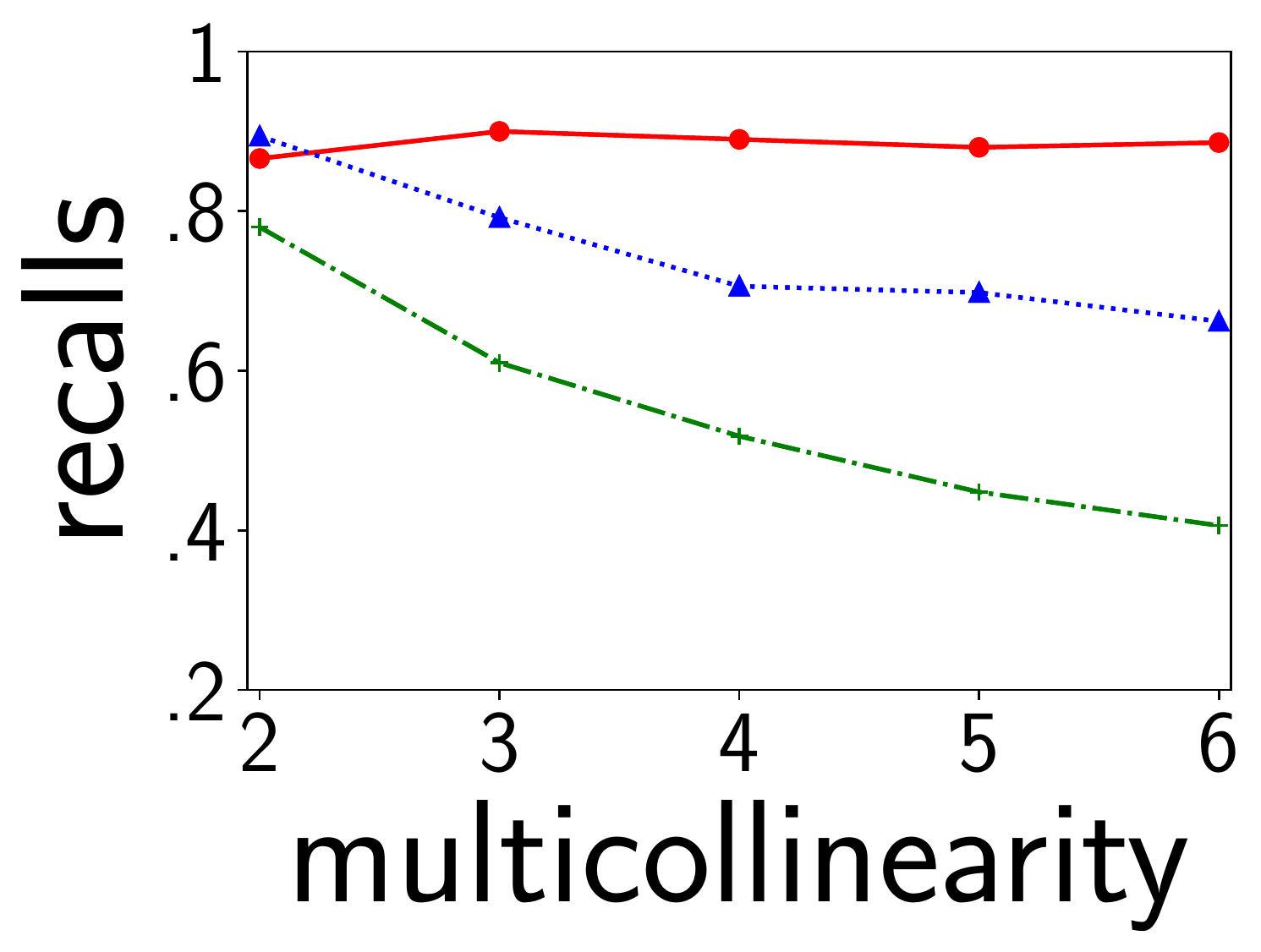}}
\subfigure[Pre. with Co.]{
\includegraphics[width=0.14\textwidth]{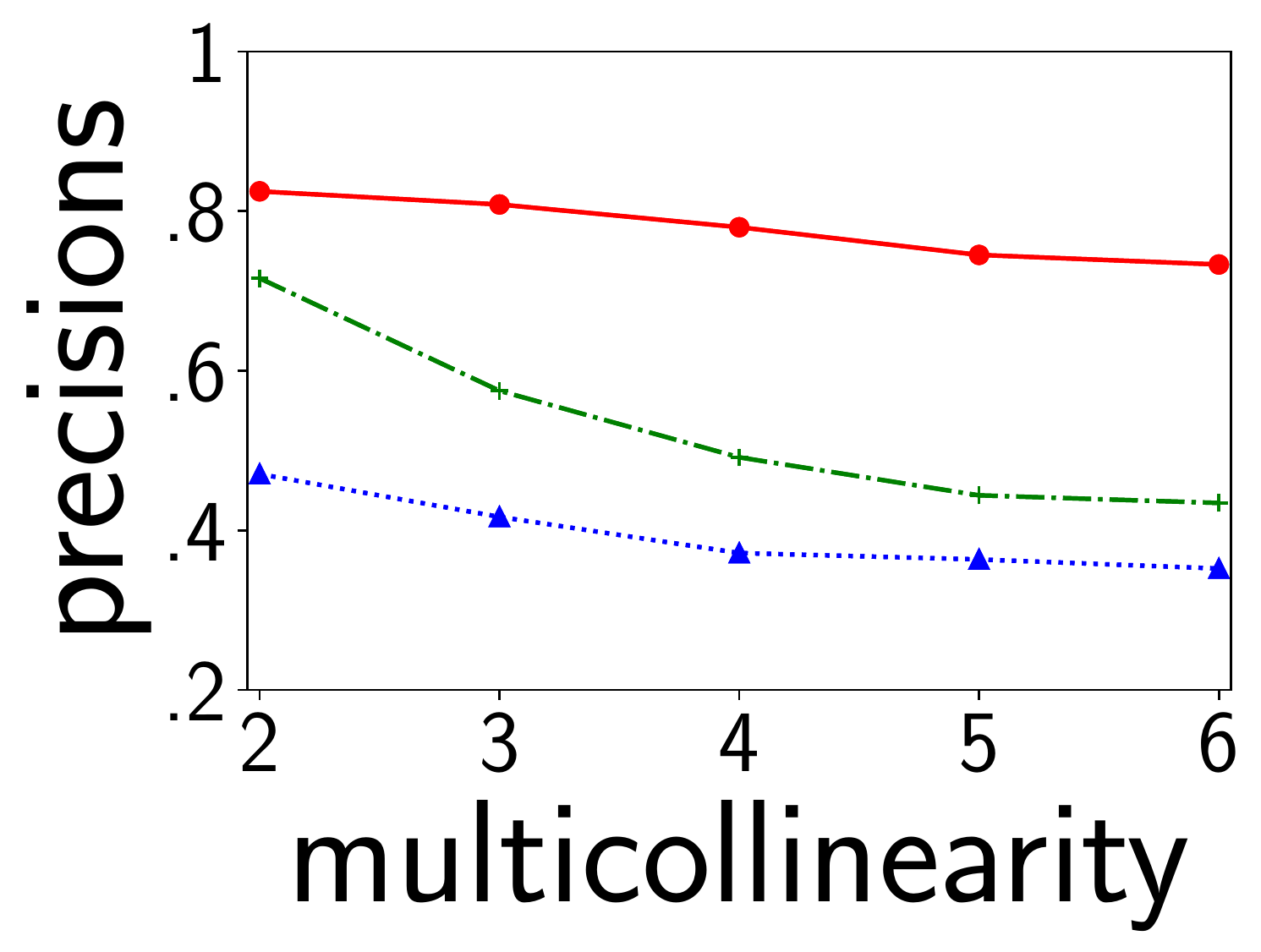}}
\subfigure[F1. with Co.]{
\includegraphics[width=0.17\textwidth]{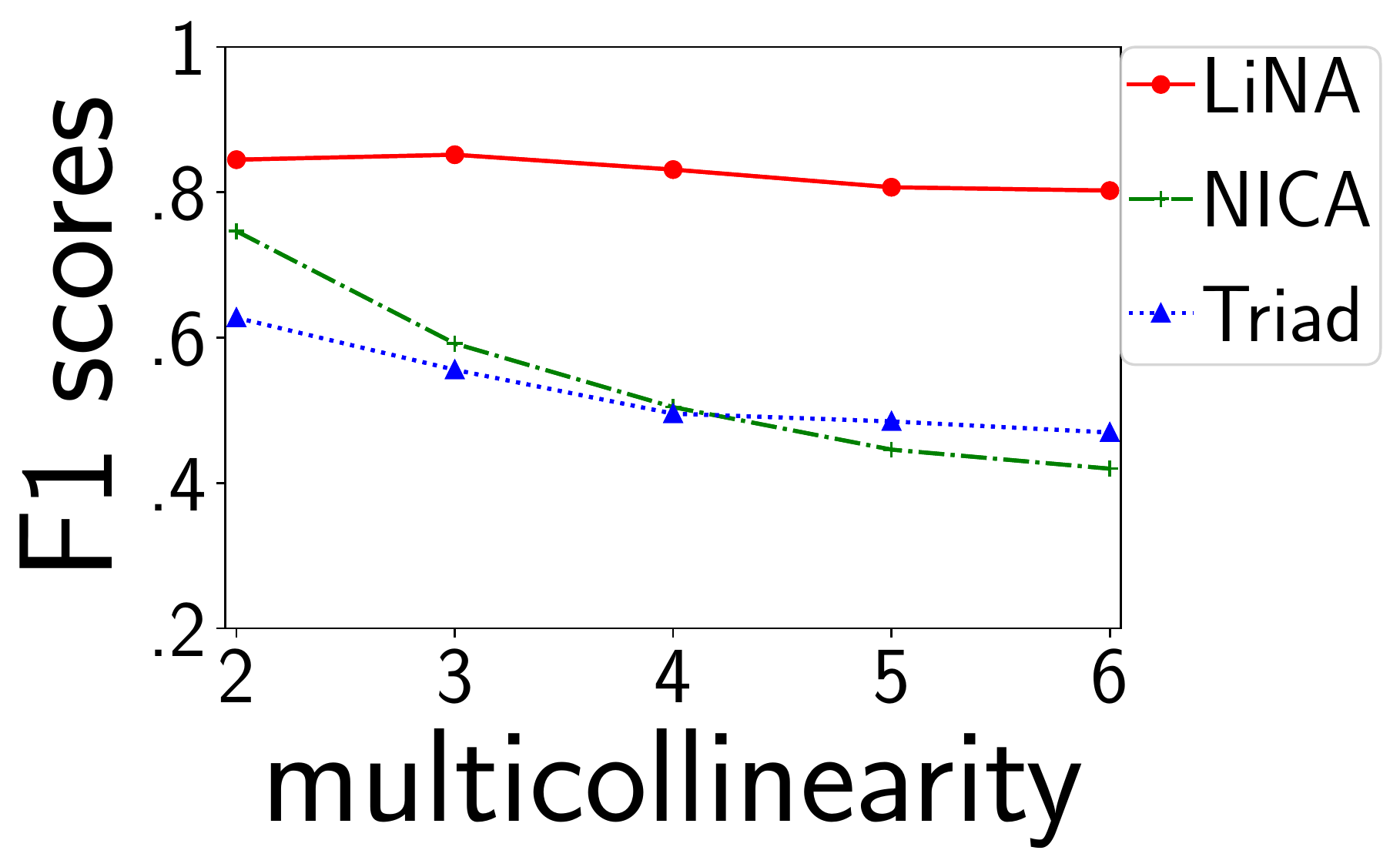}}
\subfigure[Re. with Mul.]{
\includegraphics[width=0.14\textwidth]{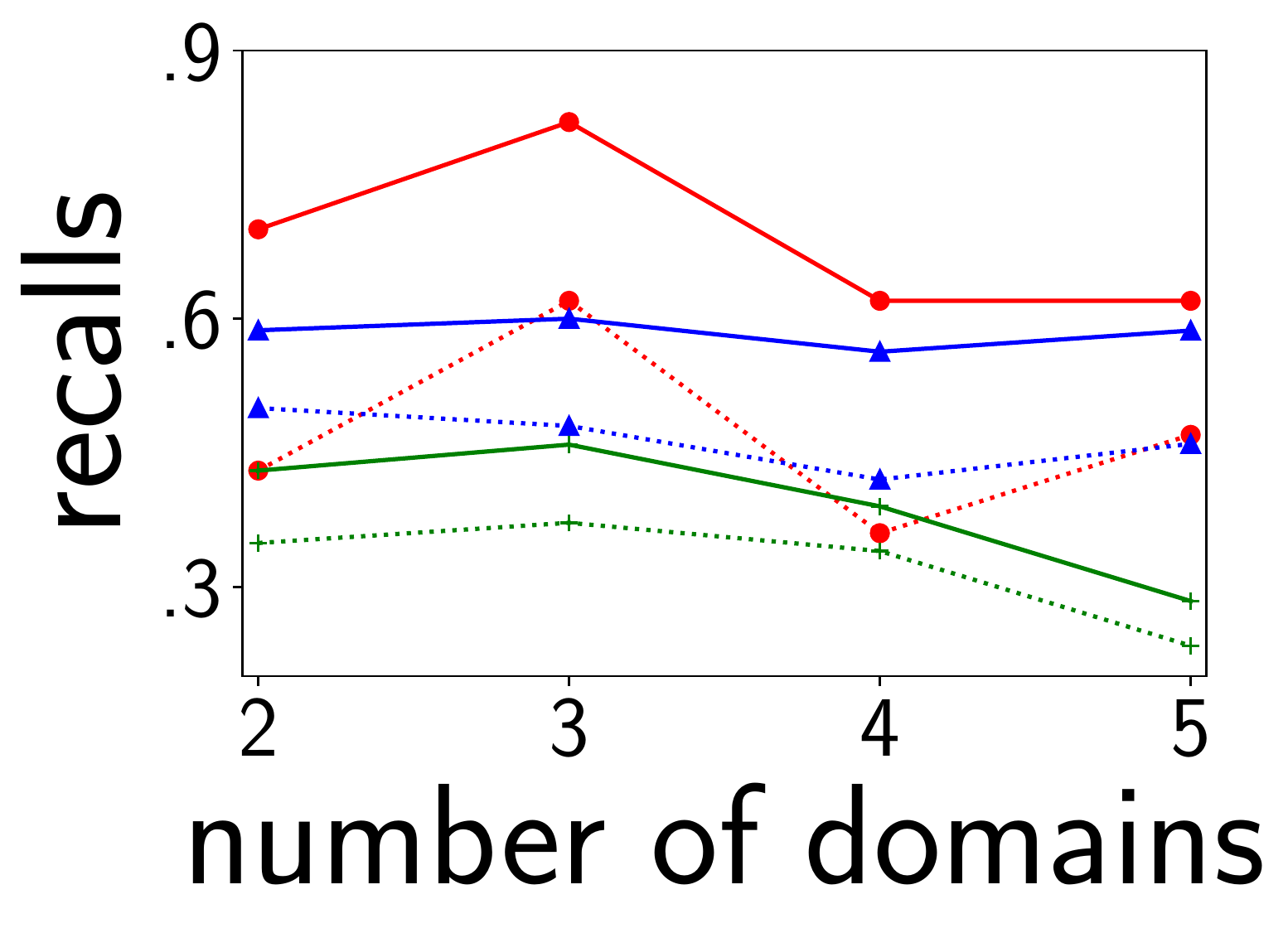}}
\subfigure[Pre. with Mul.]{
\includegraphics[width=0.14\textwidth]{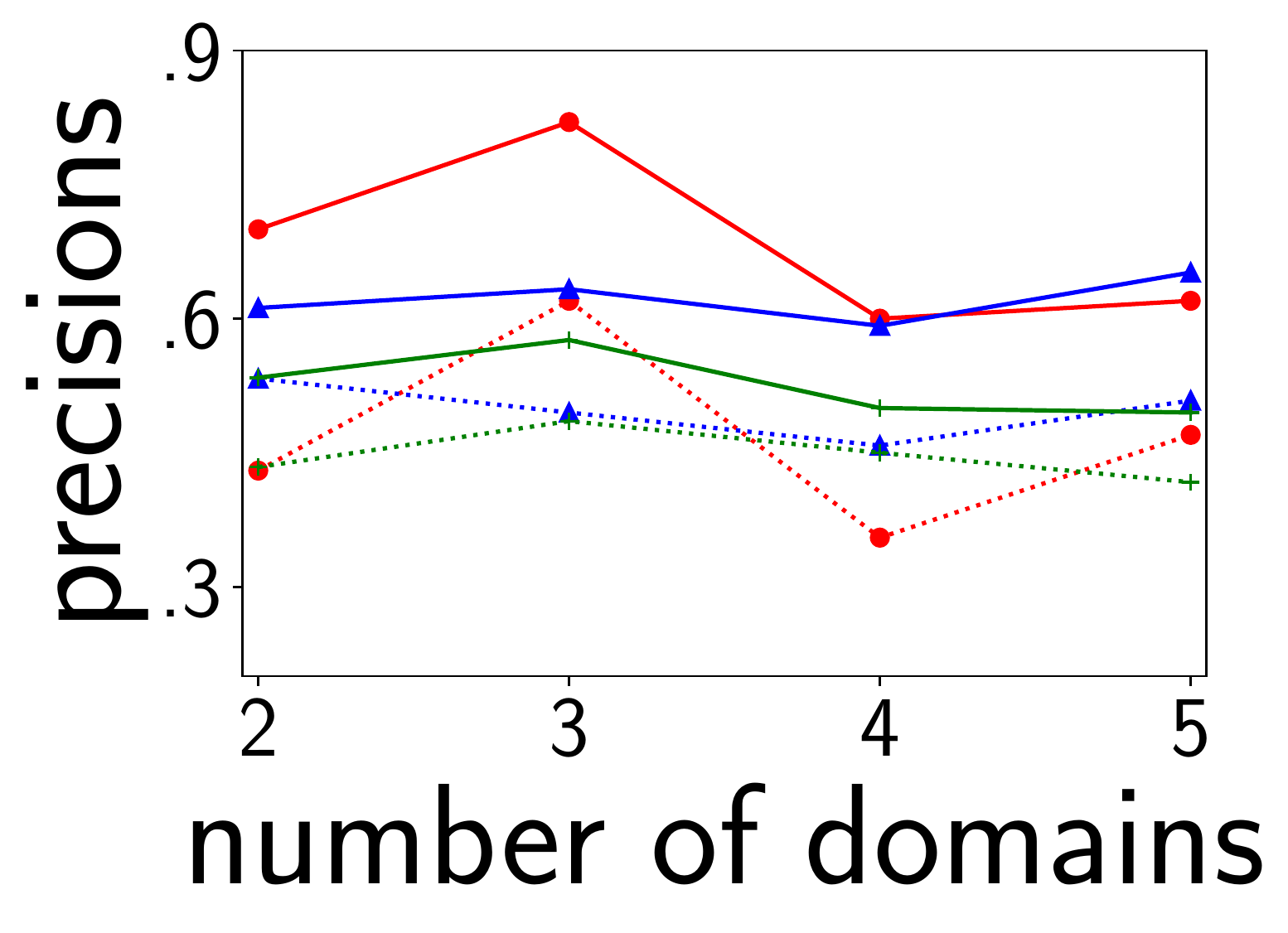}}
\subfigure[F1. with Mul.]{
\includegraphics[width=0.17\textwidth]{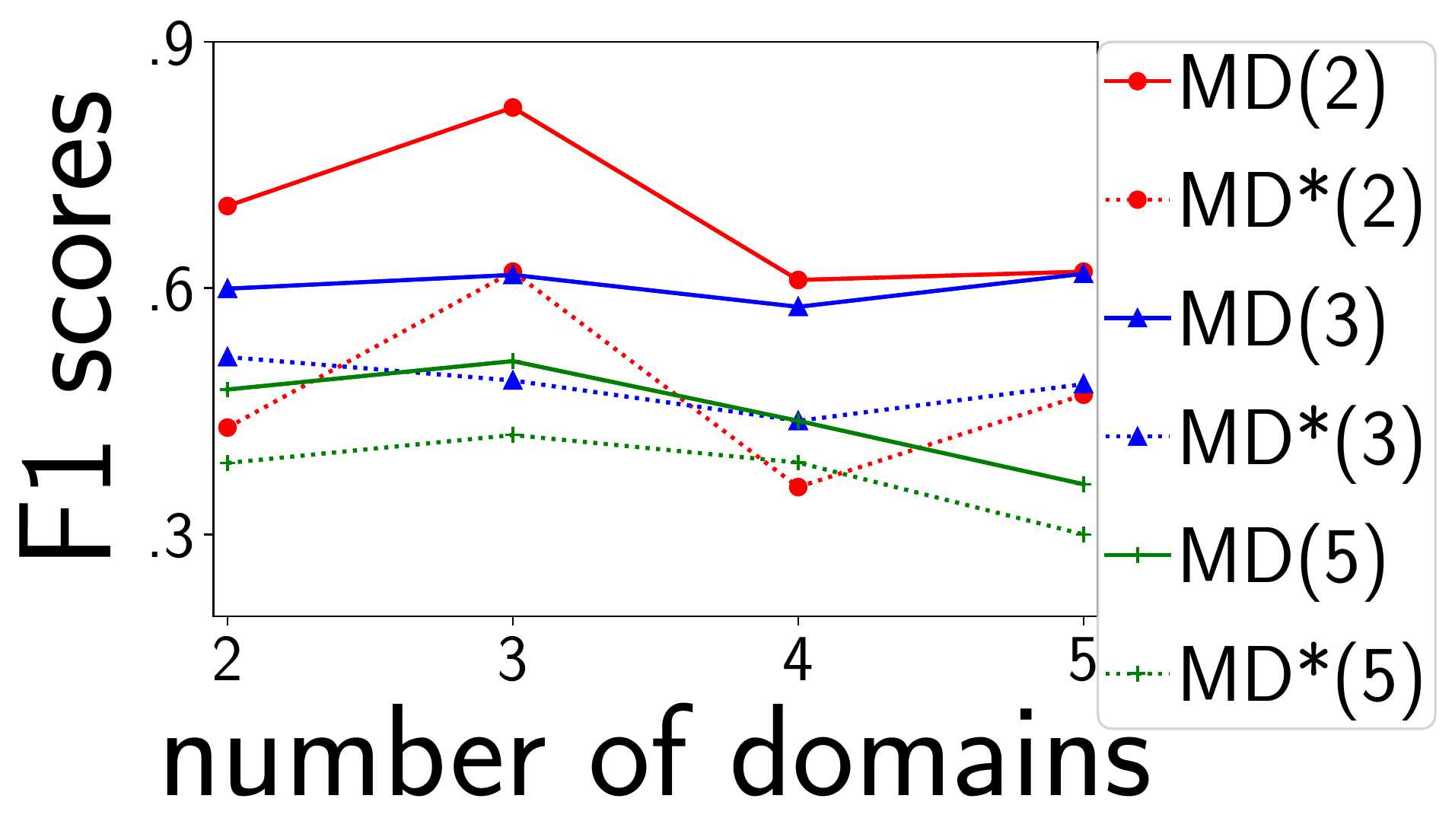}}
\caption{The recall (Re.), precision (Pre.) and F1 scores (F1.) of the recovered causal graphs between latent factors with different sample sizes (Sa.) i.e., $n=$ 100, 200, 500, 1000, 2000 in (a), (b) and (c), whose noises of latent factors follow Laplace distributions; with different levels of multicollinearities (Co.) in (d), (e) and (f). In particular, in the x-axis, levels of multicollinearities of $i=2,...,6$ are 22\%, 47\%, 69\%, 78\% and 84\%, repectively, in the average VIF; and with different numbers of domains (Mul.) in (g), (h) and (i). Solid lines are from our MD-LiNA while dotted lines are from the comparison MD*. Higher F1 score represents higher accuracy.}
\label{fig:diff_sam}
\end{figure}

\textbf{i) Different sample sizes} $n=$ 100, 200, 500, 1000, 2000, to verify the capability in small-sample-size schemes. In Figure~\ref{fig:diff_sam}(a)-(c),
we found our LiNA has the best performance, especially with smaller sample sizes. As sample sizes decrease, performances of other methods decrease whereas LiNA remains incredibly preponderant. Although Triad's recall is comparable to ours with enough sample sizes, its precision is the worst. The reason may be the sample sizes are not adequate to prune the directions, producing redundant edges.

\textbf{ii) Highly-correlated variables} through different effects of latent factors $[-i,-0.5] \cup [0.5,i]$, $i=$ 2,..., 6. Their corresponding average Variable Inflation Factors (VIF)~\footnote{Average Variable Inflation Factors (VIF) are defined as the average VIF for all 100 independent trials of each range of weights.} are 22\%, 47\%, 69\%, 78\% and 84\%, respectively, which measure the multicollinearity of variables and higher VIFs mean the heavier multicollinearity. In Figure~\ref{fig:diff_sam}(d)-(f), we found that as VIF increases, the accuracy of all methods declines in different degrees, but our method still outperforms the other comparisons, due to the employment of the elastic net regularization.

\begin{figure}[t!]
\centering
\subfigure[$\bm{G}$ ($q$=2)]{
\includegraphics[width=0.145\textwidth]{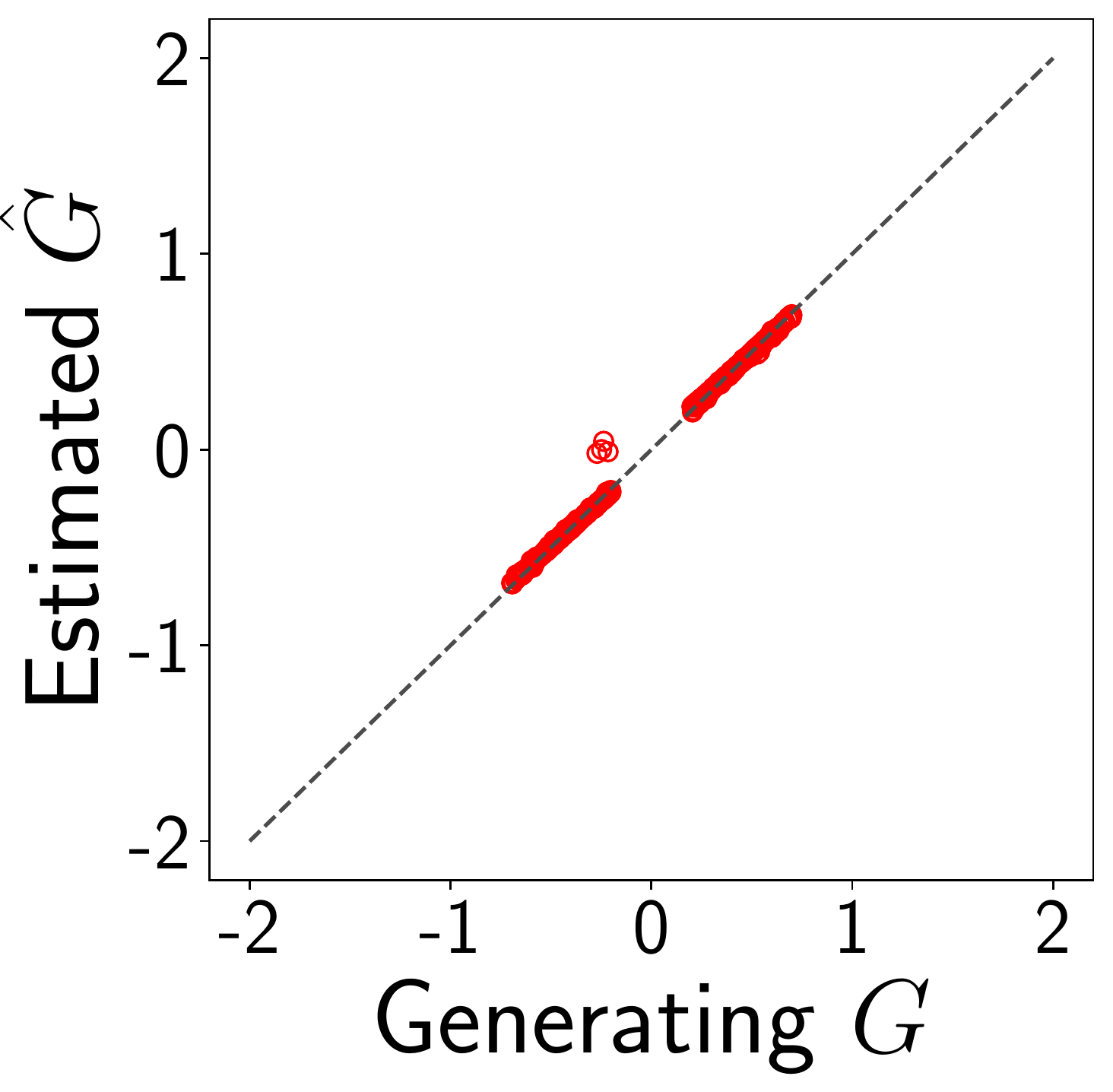}} 
\subfigure[$\bm{B}_{\mathrm{LiNA}}$ ($q$=2)]{
\includegraphics[width=0.145\textwidth]{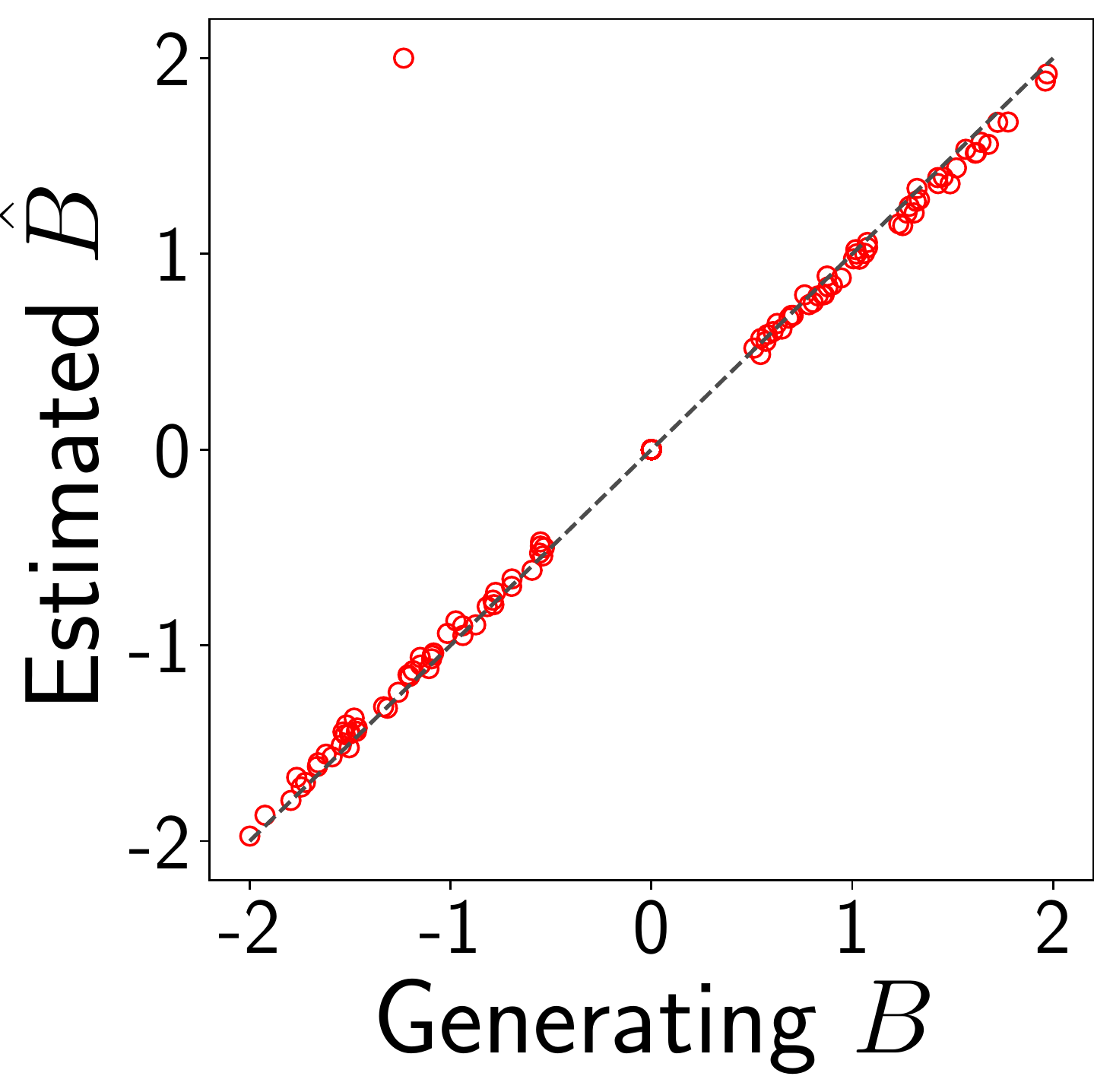}}
\subfigure[$\bm{B}_{\mathrm{NICA}}$ ($q$=2)]{
\includegraphics[width=0.145\textwidth]{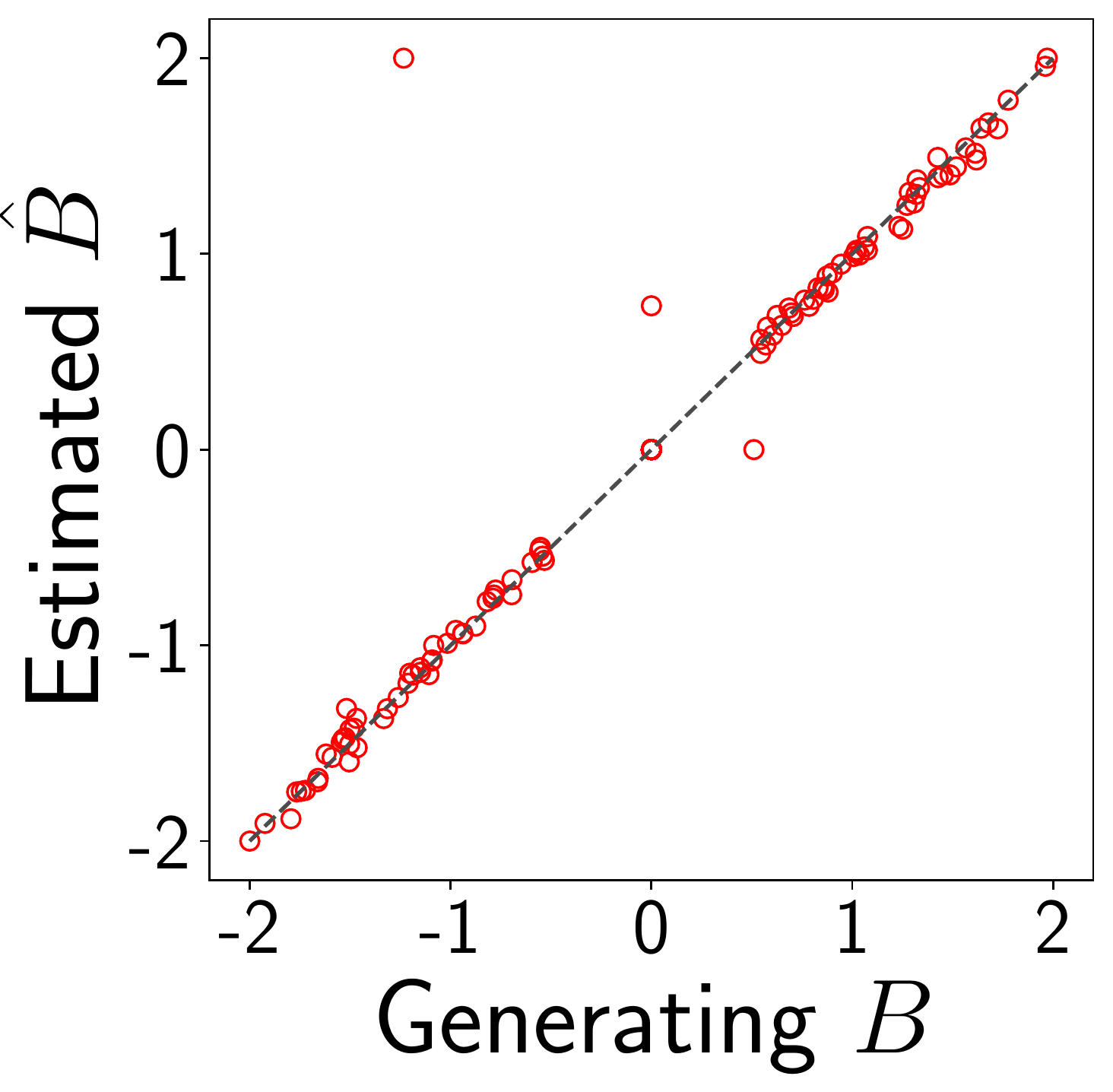}}
\subfigure[$\bm{G}$ ($q$=3)]{
\includegraphics[width=0.145\textwidth]{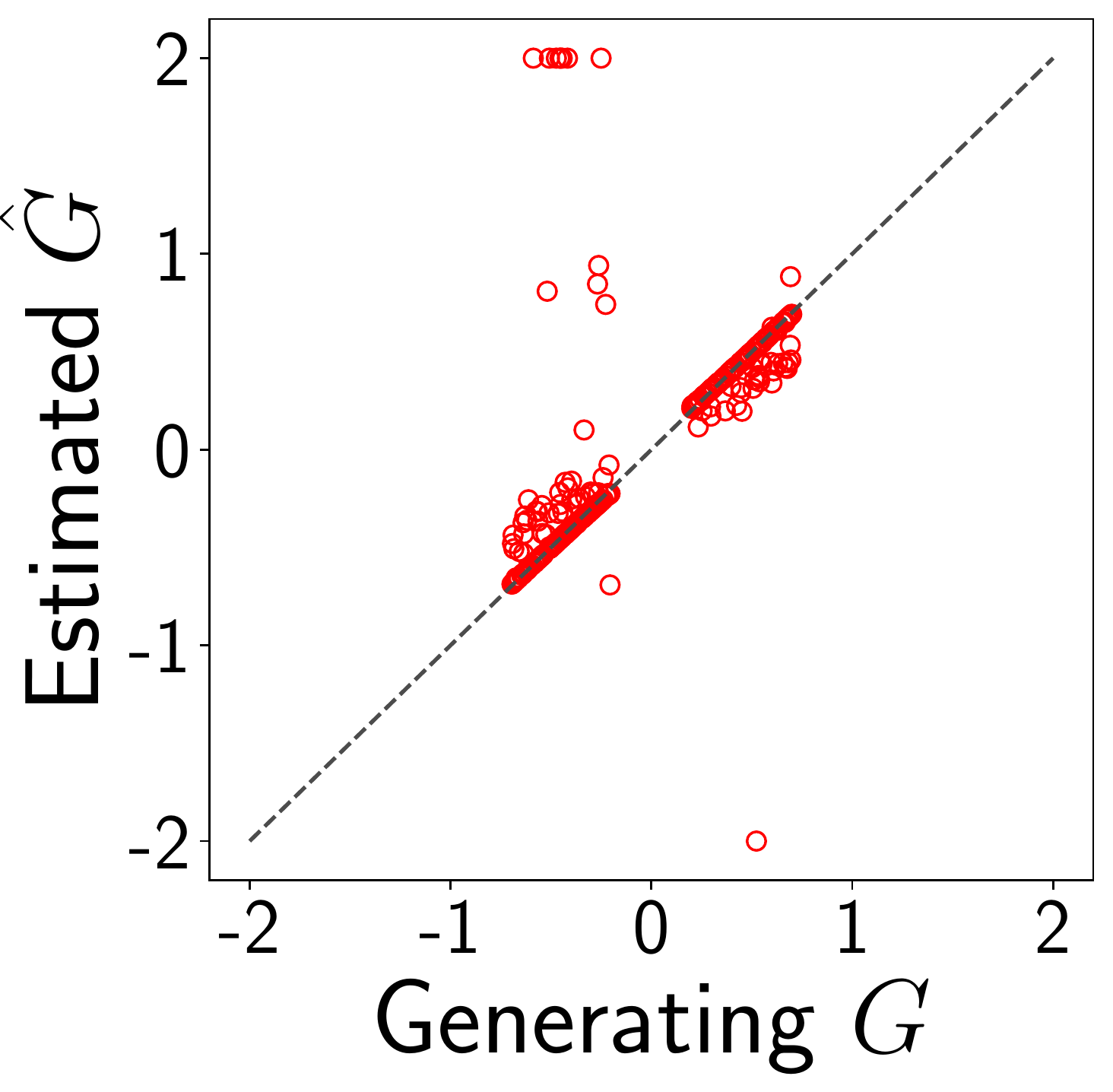}} 
\subfigure[$\bm{B}_{\mathrm{LiNA}}$ ($q$=3)]{
\includegraphics[width=0.145\textwidth]{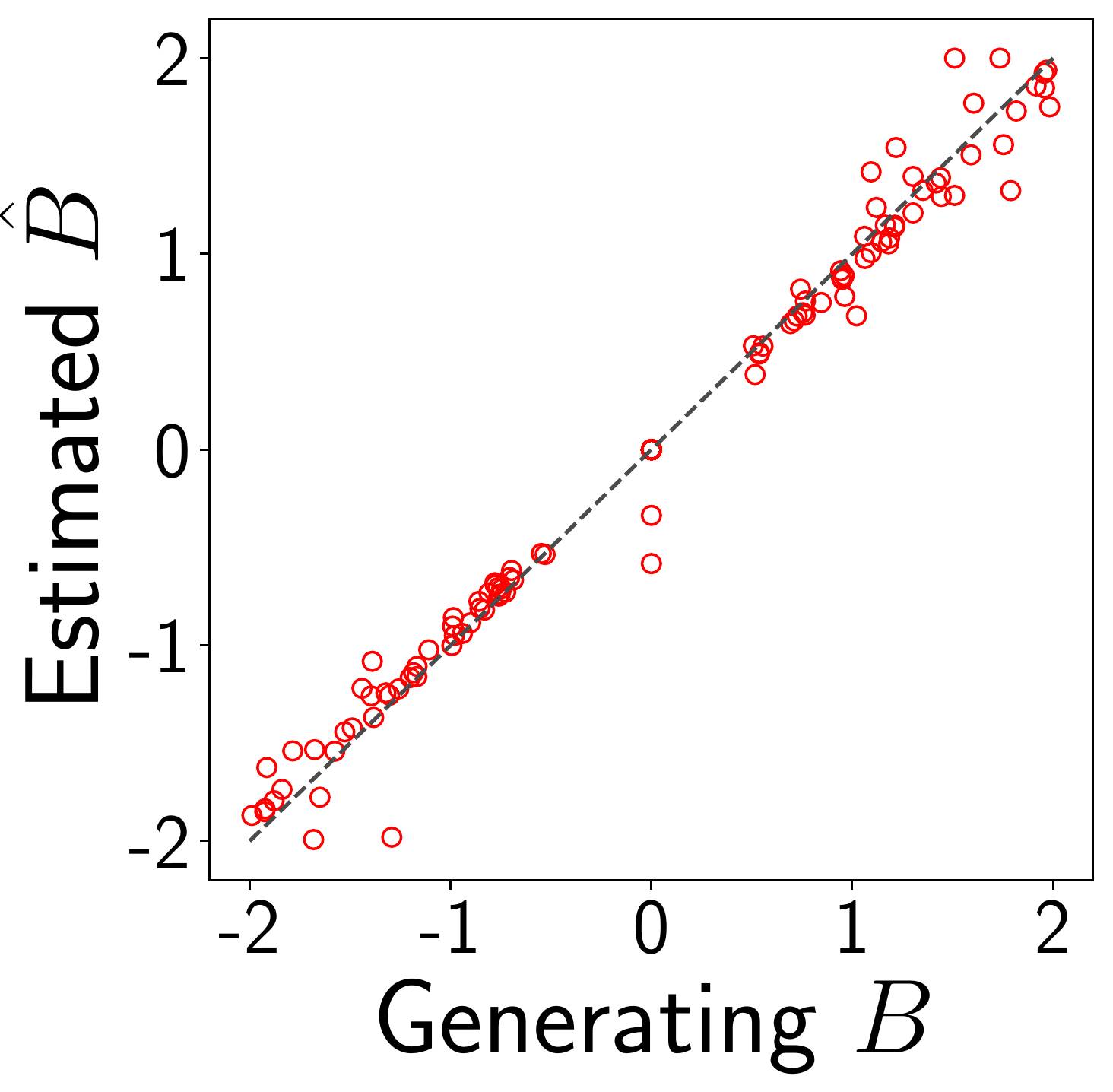}}
\subfigure[$\bm{B}_{\mathrm{NICA}}$ ($q$=3)]{
\includegraphics[width=0.145\textwidth]{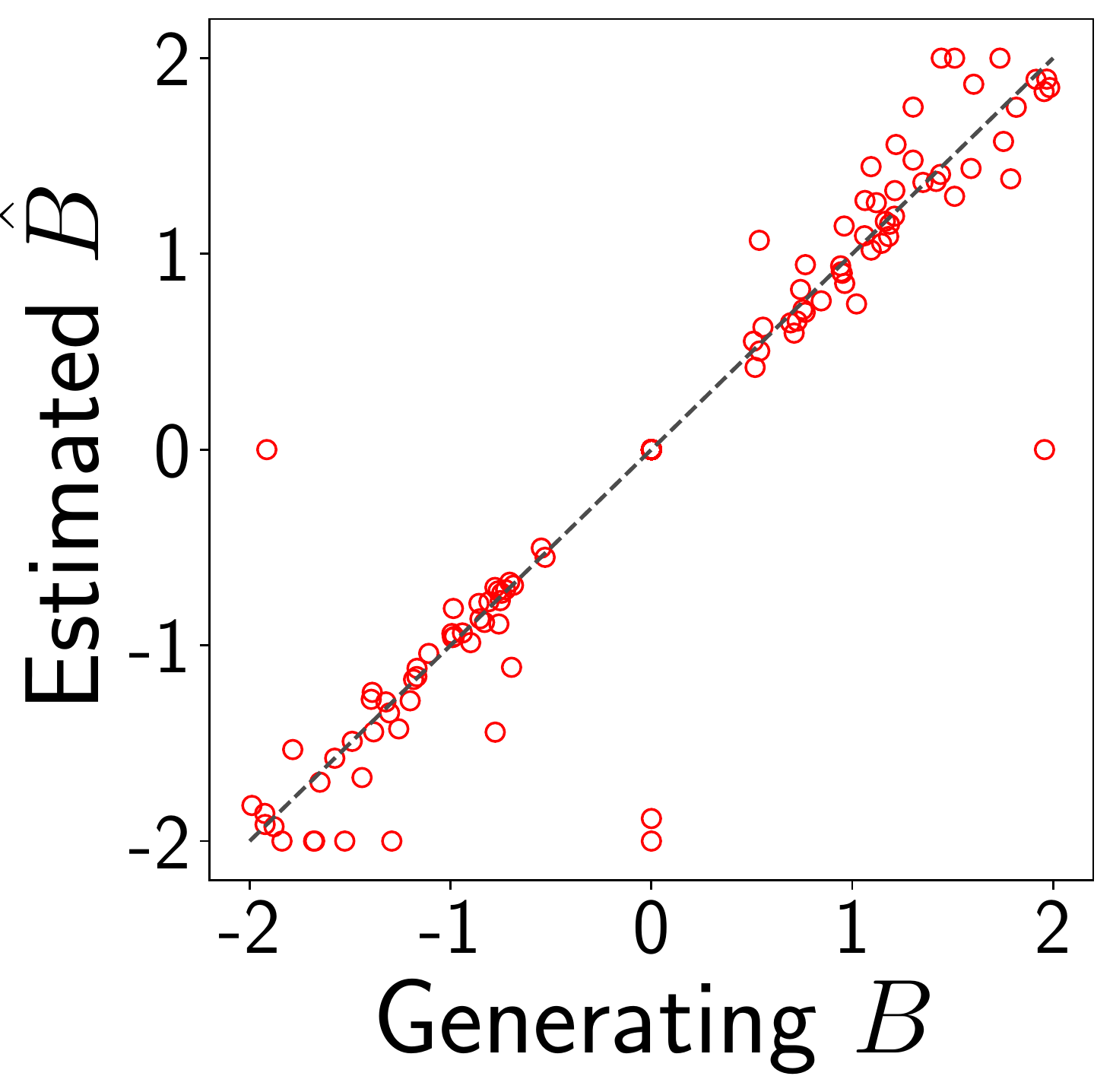}}
\subfigure[$\bm{G}$ ($q$=5)]{
\includegraphics[width=0.145\textwidth]{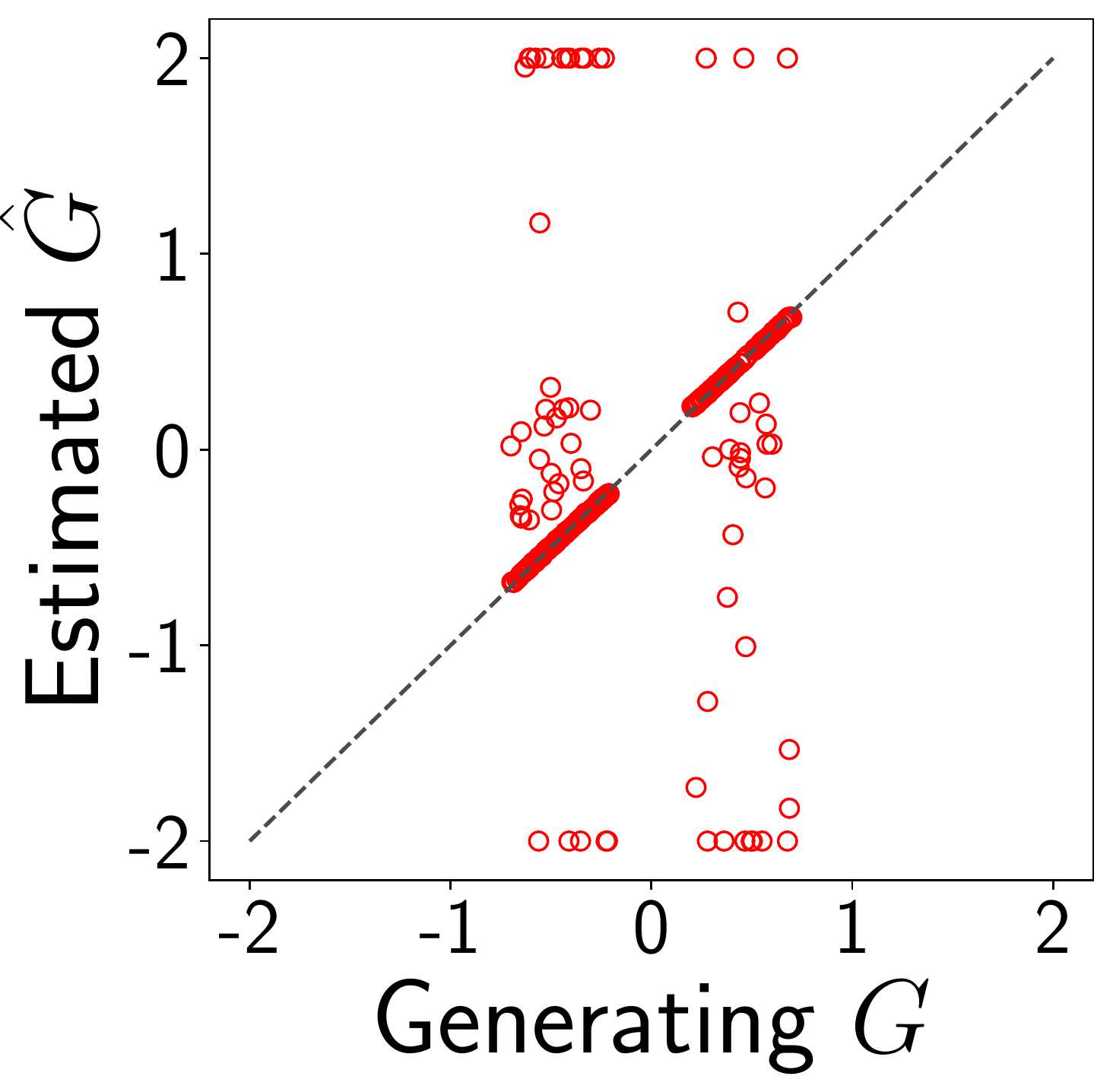}} 
\subfigure[$\bm{B}_{\mathrm{LiNA}}$ ($q$=5)]{
\includegraphics[width=0.145\textwidth]{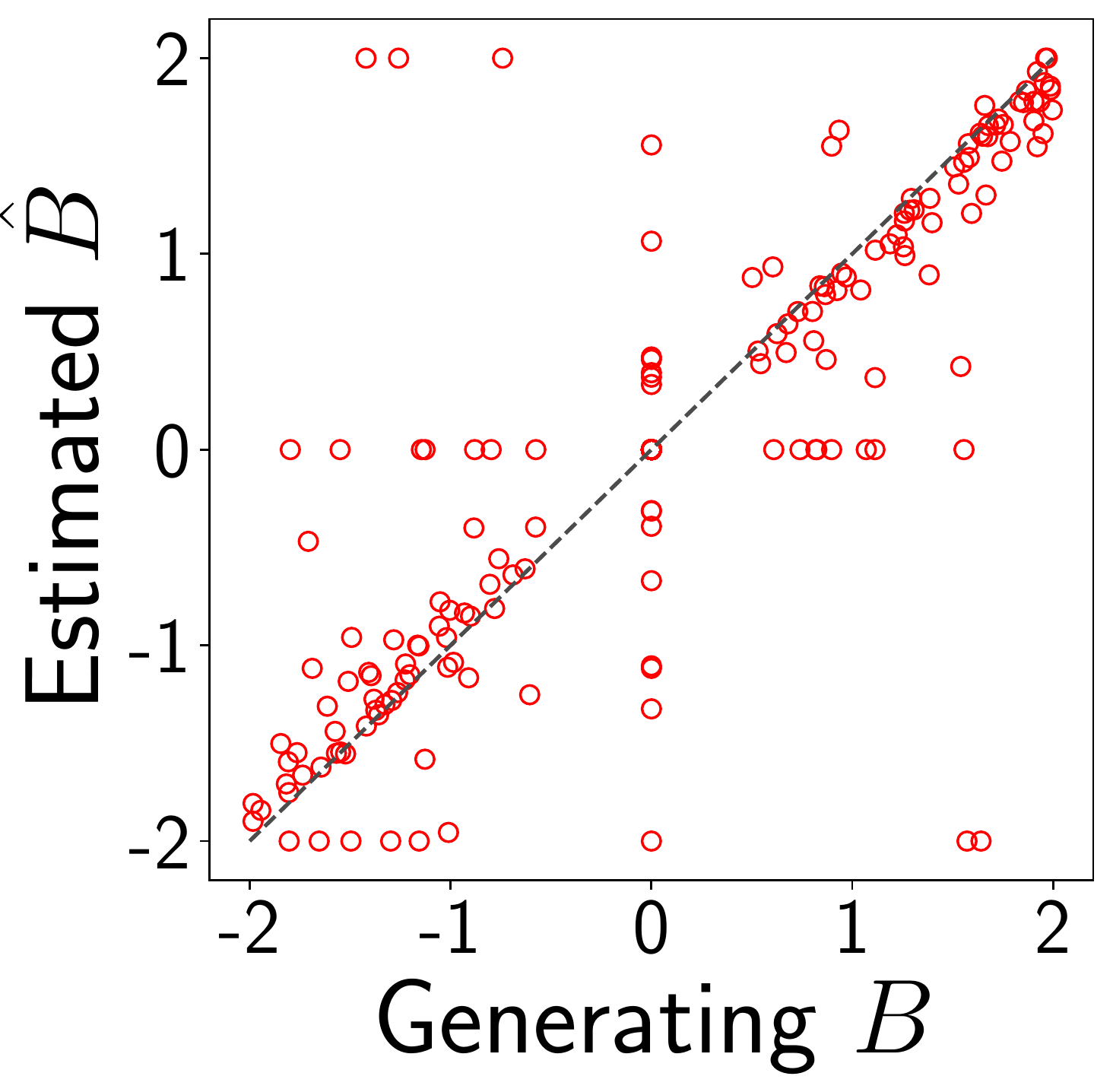}}
\subfigure[$\bm{B}_{\mathrm{NICA}}$ ($q$=5)]{
\includegraphics[width=0.145\textwidth]{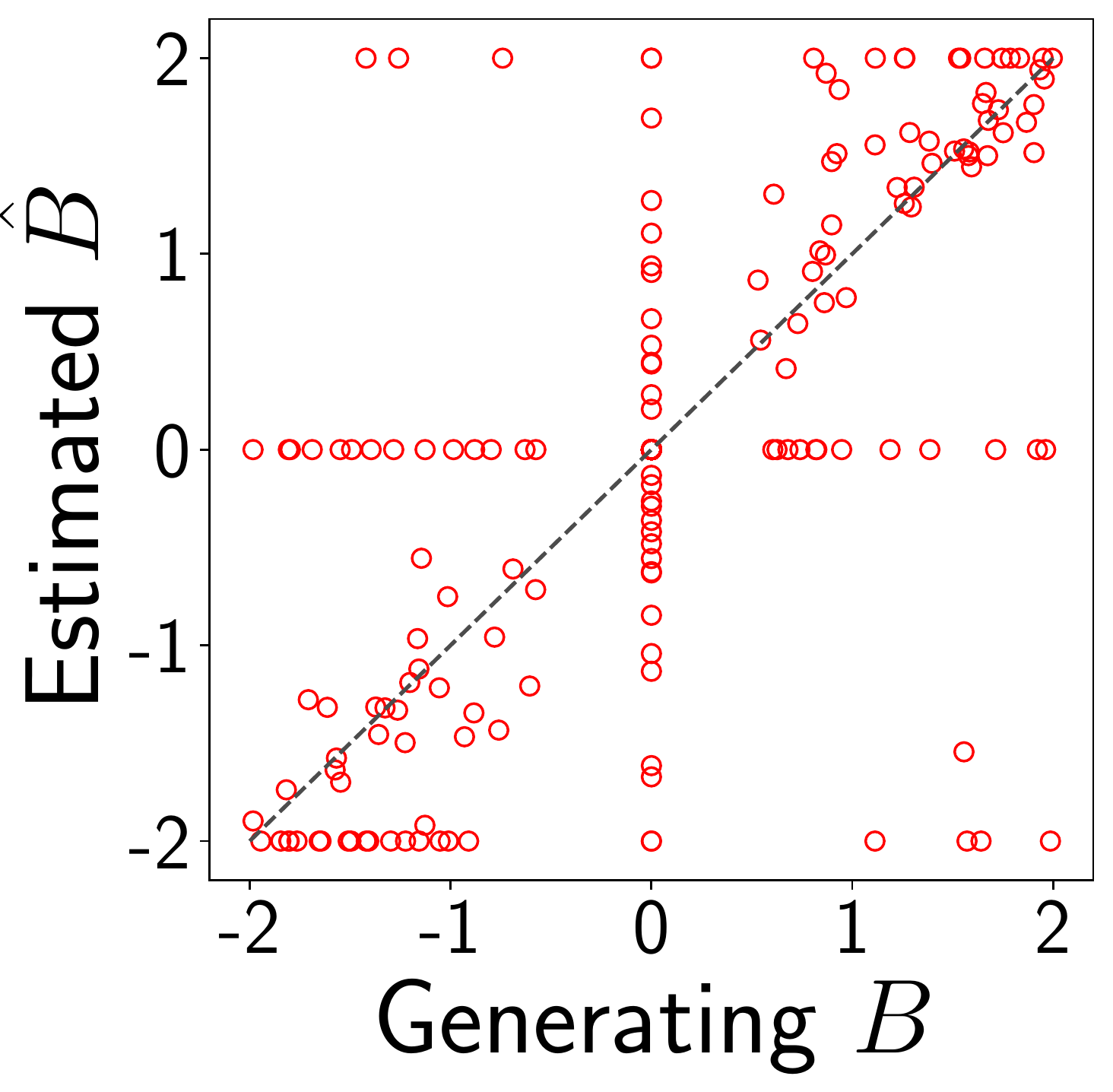}}
\subfigure[$\bm{G}$ ($q$=10)]{
\includegraphics[width=0.145\textwidth]{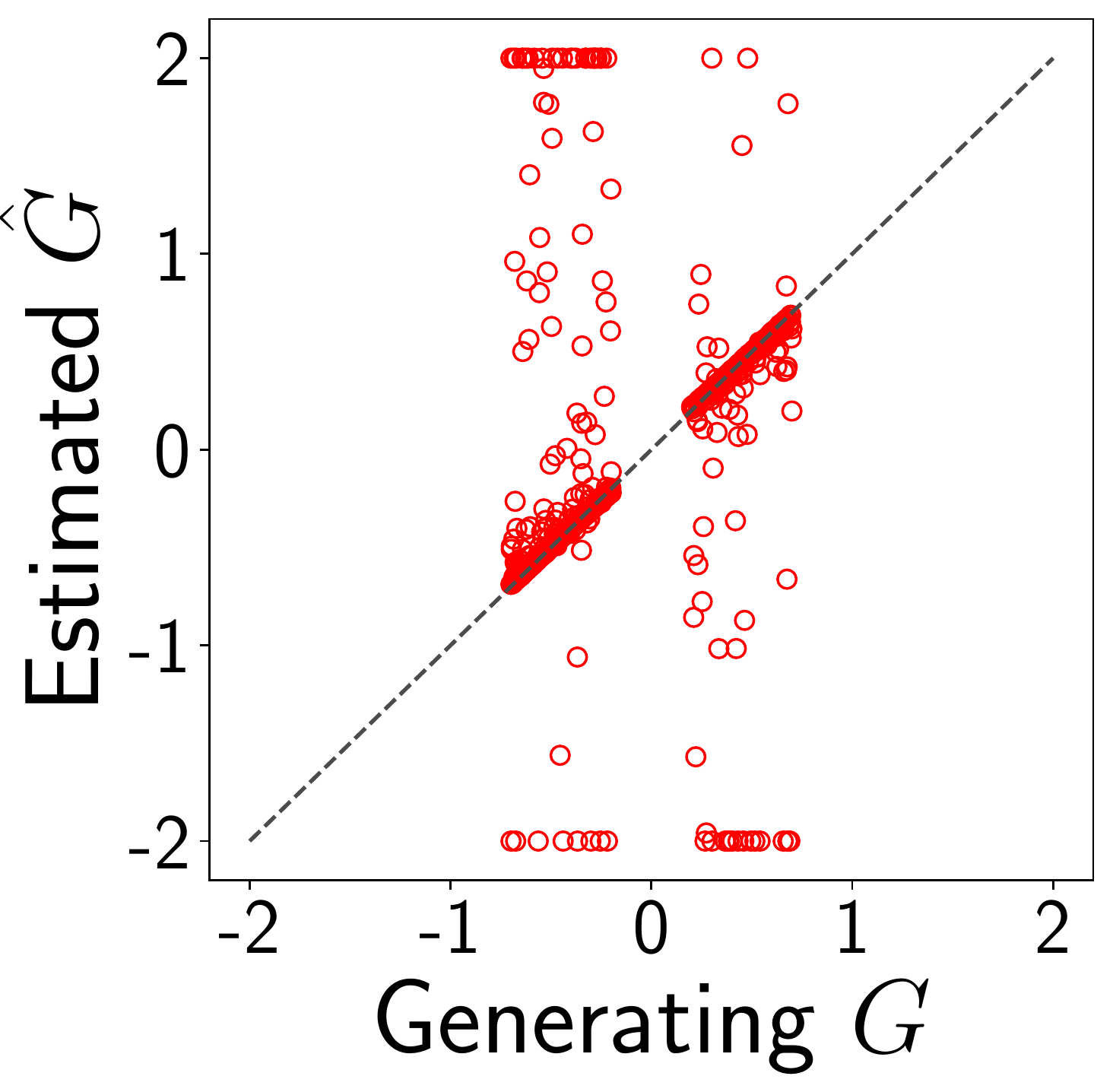}} 
\subfigure[$\bm{B}_{\mathrm{LiNA}}$ ($q$=10)]{
\includegraphics[width=0.145\textwidth]{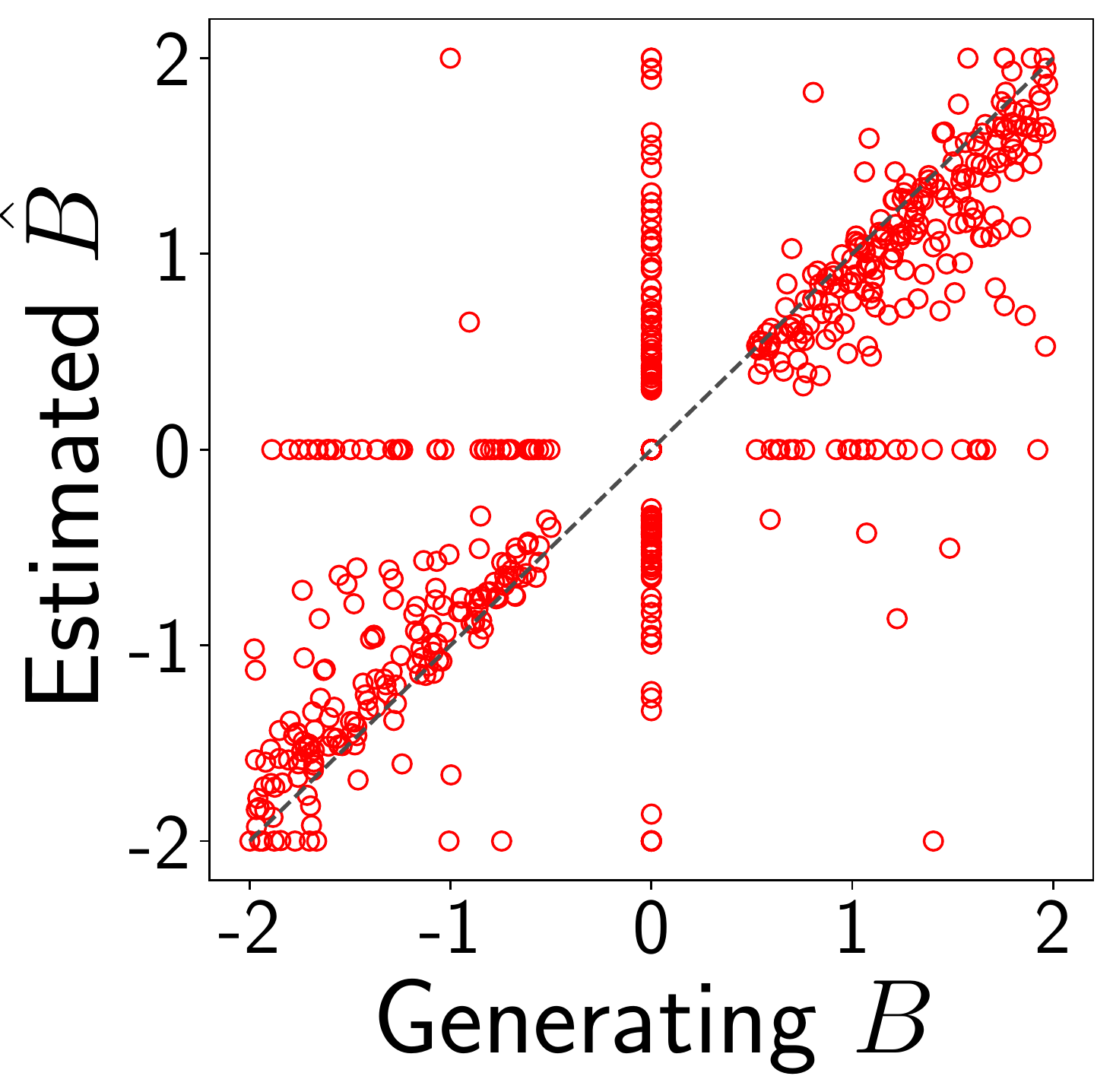}}
\subfigure[$\bm{B}_{\mathrm{NICA}}$ ($q$=10)]{
\includegraphics[width=0.145\textwidth]{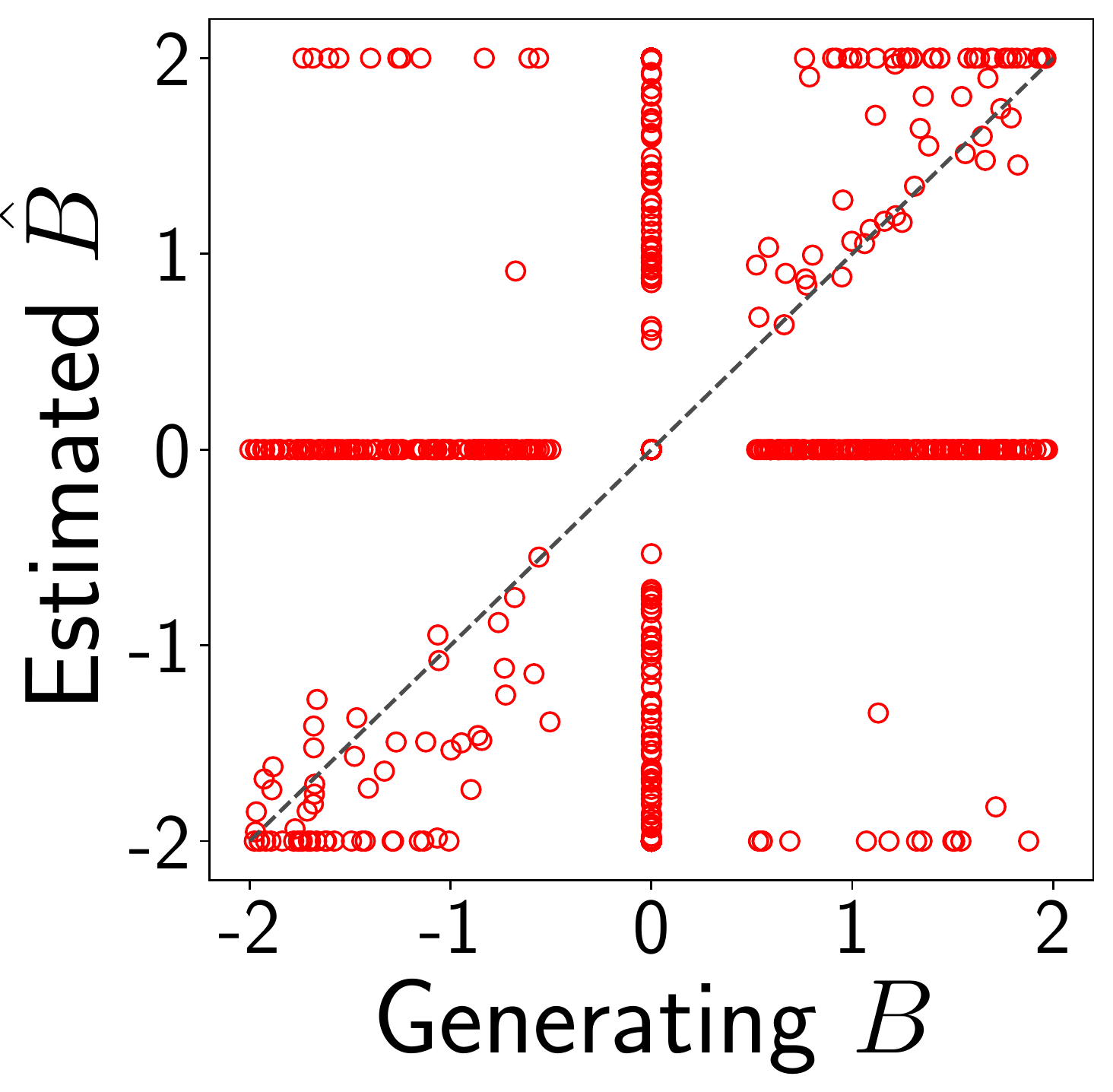}}
\caption{Scatter plots of estimated causal structures versus the true ones with different numbers of latent factors $q$. Note that estimates outside the interval [-2,2] are plotted at the edges of this interval.
(a), (d), (g) and (j) are scatter plots of the estimated factor loading matrix versus the true ones. (b), (e), (h) and (k) are scatter plots of our method's estimated adjacency matrix versus the true ones while (c), (f), (i) and (l) are of NICA method's estimated matrix. The x-axis is the generating $\bm{G}$ or $\bm{B}$ while the y-axis is the estimated $\hat{\bm{G}}$ or $\hat{\bm{B}}$. Closer to the main diagonal means higher accuracy.}\label{fig:diff_nf}
\end{figure}

\textbf{iii) Different numbers of latent factors} $q = 2,3,5,10$, to emphasize the capability of estimating causal effects, compared with NICA (Triad is not compared since it does not estimate effects).
We applied the same method, CFA, to estimate $\hat{\bm{G}}$ as the NICA.
Overall, in Figure~\ref{fig:diff_nf}, we found LiNA gives better performances in all cases. The accuracy decreases along with more latent factors. 
Specifically, NICA is comparable to ours when $q=$ 2. However, as $q=$ 10, accuracies both decrease, but LiNA decreases much more slowly than NICA. The reasons may be 1) more measurement variables with the fixed sample size results in reduced power of CFA to estimate $\hat{\bm{G}}$, propagating errors to learn $\hat{\bm{B}}$; 2) the sparsity constraint deals with small sample sizes while NICA does not. And NICA does not estimate the causal directions and effects simultaneously, which may lack statistical efficiency.

\textbf{iv) Multi-domain data} $M=$ 2, 3, 4, 5, through varying noises $\bm{\varepsilon}$'s distributions (sub-Gaussian or super-Gaussian). To obtain the true graph for evaluation, we generated the identical graphs of latent factors in each domain. We varied the number of latent factors, $q_m=$ 2, 3, 5.
In Figure~\ref{fig:diff_sam}(g)-(i), we found F1 scores of both methods tend to decrease with more domains or more latent factors increases. Specifically, in all cases MD-LiNA gives a better performance compared with MD*, in that MD* did neglect the problem that factors from different domains are represented by which factors of interest. Further, though we experimented with only $q_m=$ 2, 3, 5, the whole causal graph is much more complicated, which has totally $Mq_m$ latent factors and $2Mq_m$ observed variables.

\subsection{Real-World Applications}
\paragraph{Yahoo stock indices dataset.} We aimed to find the causal structure between different regions of the world, i.e., Asia, Europe, and the USA, each of which consisted of 2/3 stock indices. They were $ \mathrm{Asia:= \lbrace N225, 000001.SS} \rbrace$ from Japan and China, $\mathrm{ Europe:= \lbrace BUK100P, FCHI, N100\rbrace}$ from United Kingdom, France, and other European countries, and $\mathrm{USA:= \lbrace DJI, GSPC, NYA\rbrace}$ from the United States. We divided the data into two non-overlapping time segments such that their distributions varied across segments and are viewed as two different domains. We tested its multicollinearity and used 10-fold cross validation to select parameter values. Details are in SM H.1. Due to the different time zones, it is expected the ground truth is $\mathrm{Asia} \to \mathrm{Europe} \to \mathrm{USA}$~\cite{janzing2010telling,chen2014causal}, with which our recovered causal structure in Figure~\ref{fig:application} was in accordance.

\begin{figure}[t!]	
	\centering
	\resizebox{0.45\textwidth}{!}{
	\begin{tikzpicture}[scale=0.6, line width=0.5pt, inner sep=0.2mm, shorten >=.1pt, shorten <=.1pt]
	\draw (0, 0) node(1) [circle, minimum size=7mm, draw] {{\footnotesize\,Asia\,}};
	\draw (1.2, 1.3) node(5) [rectangle, minimum size=4mm, draw] {{\footnotesize\,N225\,}};
	\draw (-1.2, 1.3) node(6) [rectangle, minimum size=4mm, draw] {{\footnotesize\,000001.SS\,}};
	\draw (-1.5, -2.0) node(2) [circle, minimum size=7mm, draw] {{\footnotesize\,Eur.\,}};
	\draw (-2, 0) node(7) [rectangle, minimum size=4mm, draw] {{\footnotesize\,BUK100P\,}};
	\draw (-2.3, -3.5) node(8) [rectangle, minimum size=4mm, draw] {{\footnotesize\,FCHI\,}};
	\draw (-0.8, -3.5) node(9) [rectangle, minimum size=4mm, draw] {{\footnotesize\,N100\,}};
	\draw (1.5, -2.0) node(3) [circle, minimum size=7mm, draw] {{\footnotesize\,USA\,}};
	\draw (1.6, 0) node(10) [rectangle, minimum size=4mm, draw] {{\footnotesize\,GSPC\,}};
	\draw (0.75, -3.5) node(11) [rectangle, minimum size=4mm, draw] {{\footnotesize\,NYA\,}};
	\draw (2, -3.5) node(12) [rectangle, minimum size=4mm, draw] {{\footnotesize\,DJI\,}};
	
	\draw[-arcsq,blue,very thick] (1) -- (2) node[pos=0.5,sloped,above] {{\footnotesize\,-2.84\,}}; 
	\draw[-arcsq,blue,very thick] (2) -- (3)node[pos=0.5,sloped,above] {{\footnotesize\,1.85\,}}; 
	\draw[-arcsq,blue,very thick] (1) -- (3) node[pos=0.5,sloped,above] {{\footnotesize\,0.28\,}}; 
	\draw[-arcsq,very thick] (1) -- (5) node[pos=0.5,sloped,above] {};
	\draw[-arcsq,very thick] (1) -- (6) node[pos=0.5,sloped,above] {};
	\draw[-arcsq,very thick] (2) -- (9)node[pos=0.5,sloped,above] {};
	\draw[-arcsq,very thick] (2) -- (7)node[pos=0.5,sloped,above] {};
	\draw[-arcsq,very thick] (2) -- (8)node[pos=0.5,sloped,above] {};
	\draw[-arcsq,very thick] (3) -- (10)node[pos=0.5,sloped,above] {};
	\draw[-arcsq,very thick] (3) -- (11)node[pos=0.5,sloped,above] {};
	\draw[-arcsq,very thick] (3) -- (12)node[pos=0.5,sloped,above] {};
	\end{tikzpicture}~~~~~~~~~~~~~~~~~~
	\begin{tikzpicture}[scale=0.6, line width=0.5pt, inner sep=0.2mm, shorten >=.1pt, shorten <=.1pt]
	\draw (0, 0) node(1) [circle, minimum size=7mm, draw] {{\footnotesize\,Asia\,}};
	\draw (1.2, 1.3) node(5) [rectangle, minimum size=4mm, draw] {{\footnotesize\,N225\,}};
	\draw (-1.2, 1.3) node(6) [rectangle, minimum size=4mm, draw] {{\footnotesize\,000001.SS\,}};
	\draw (-1.5, -2.0) node(2) [circle, minimum size=7mm, draw] {{\footnotesize\,Eur.\,}};
	\draw (-2, 0) node(7) [rectangle, minimum size=4mm, draw] {{\footnotesize\,BUK100P\,}};
	\draw (-2.3, -3.5) node(8) [rectangle, minimum size=4mm, draw] {{\footnotesize\,FCHI\,}};
	\draw (-0.8, -3.5) node(9) [rectangle, minimum size=4mm, draw] {{\footnotesize\,N100\,}};
	\draw (1.5, -2.0) node(3) [circle, minimum size=7mm, draw] {{\footnotesize\,USA\,}};
	\draw (1.6, 0) node(10) [rectangle, minimum size=4mm, draw] {{\footnotesize\,GSPC\,}};
	\draw (0.75, -3.5) node(11) [rectangle, minimum size=4mm, draw] {{\footnotesize\,NYA\,}};
	\draw (2, -3.5) node(12) [rectangle, minimum size=4mm, draw] {{\footnotesize\,DJI\,}};
	
	\draw[-arcsq,blue,very thick] (1) -- (2) node[pos=0.5,sloped,above] {{\footnotesize\,-2.84\,}}; 
	\draw[-arcsq,densely dotted,red,very thick] (3) -- (2)node[pos=0.5,sloped,above] {{\footnotesize\,1.85\,}};
	\draw[-arcsq,blue,very thick] (1) -- (3) node[pos=0.5,sloped,above] {{\footnotesize\,0.28\,}}; 
	\draw[-arcsq,very thick] (1) -- (5) node[pos=0.5,sloped,above] {};
	\draw[-arcsq,very thick] (1) -- (6) node[pos=0.5,sloped,above] {};
	\draw[-arcsq,very thick] (2) -- (9)node[pos=0.5,sloped,above] {};
	\draw[-arcsq,very thick] (2) -- (7)node[pos=0.5,sloped,above] {};
	\draw[-arcsq,very thick] (2) -- (8)node[pos=0.5,sloped,above] {};
	\draw[-arcsq,very thick] (3) -- (10)node[pos=0.5,sloped,above] {};
	\draw[-arcsq,very thick] (3) -- (11)node[pos=0.5,sloped,above] {};
	\draw[-arcsq,very thick] (3) -- (12)node[pos=0.5,sloped,above] {};
	\end{tikzpicture}}\\
	~~~~~~~(a)~~~~~~~~~~~~~~~~~~~~~~~~~~~~~~~~~~~~~~~~~~~~~~~~(b) 
	\caption{Estimated stock indices networks using the (a) MD-LiNA and (b) MD* methods. Solid blue lines denote consistent edges with the ground truth while densely dotted red lines are not. }
	\label{fig:application}
\end{figure}
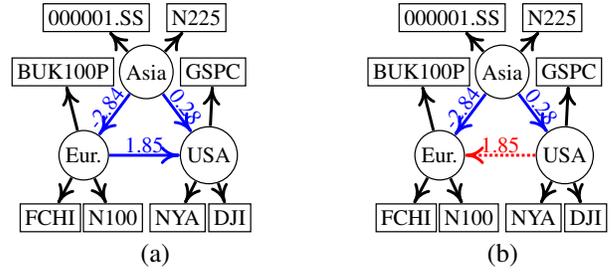
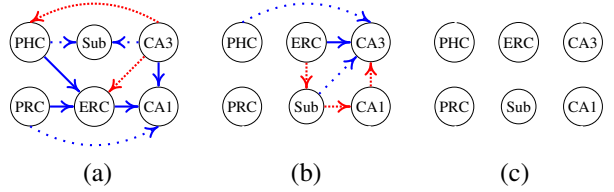
\begin{figure}[t!]  
	\centering
	\resizebox{0.45\textwidth}{!}{
	\begin{tikzpicture}[scale=1.0, line width=0.5pt, inner sep=0.2mm, shorten >=.1pt, shorten <=.1pt]
	\draw (0, -1.3) node(1) [circle, minimum size=7mm, draw] {{\footnotesize\,PRC\,}};
	\draw (2.6,0) node(2) [circle, minimum size=7mm, draw] {{\footnotesize\,CA3\,}};
	\draw (1.3, 0) node(3) [circle, minimum size=7mm, draw] {{\footnotesize\,Sub\,}};
    \draw (0, 0) node(4) [circle, minimum size=7mm, draw] {{\footnotesize\,PHC\,}};
	\draw (2.6, -1.3) node(5) [circle, minimum size=7mm, draw] {{\footnotesize\,CA1\,}};
	\draw (1.3, -1.3) node(6) [circle, minimum size=7mm, draw] {{\footnotesize\,ERC\,}};
	
	\draw[-arcsq,loosely dotted,blue,very thick] (4) -- (3) node[pos=1.5,sloped,above] {}; 
	\draw[-arcsq,loosely dotted,blue,very thick] (2) -- (3) node[pos=0.5,sloped,above] {} ; 
	\draw[-arcsq,densely dotted,red,very thick] (2) -- (6) node[pos=0.5,sloped,above] {} ; 
	\draw[-arcsq,blue,very thick] (2) -- (5) node[pos=0.5,sloped,above] {} ; 
	\draw[-arcsq,blue,very thick] (1) -- (6) node[pos=1.5,sloped,above] {}; 
	\draw[-arcsq,blue,very thick] (4) -- (6) node[pos=1.5,sloped,above] {};
	\draw[-arcsq,blue,very thick] (6) -- (5) node[pos=1.5,sloped,above] {};
	\draw[-arcsq,densely dotted, red,very thick] (2.6,0.4) to[bend right,densely dotted,red,very thick] (0,0.4);
	\draw[-arcsq,loosely dotted,blue,very thick] (0,-1.7) to[bend right,loosely dotted,blue,very thick] (2.6,-1.7);
	\end{tikzpicture}~~~~~~~~~ 
	\begin{tikzpicture}[scale=1.0, line width=0.5pt, inner sep=0.2mm, shorten >=.1pt, shorten <=.1pt]
    \draw (0, -1.3) node(1) [circle, minimum size=7mm, draw] {{\footnotesize\,PRC\,}};
    \draw (2.6,0) node(2) [circle, minimum size=7mm, draw] {{\footnotesize\,CA3\,}};
	\draw (1.3, 0) node(3) [circle, minimum size=7mm, draw] {{\footnotesize\,ERC\,}};
    \draw (0, 0) node(4) [circle, minimum size=7mm, draw] {{\footnotesize\,PHC\,}};
	\draw (2.6, -1.3) node(5) [circle, minimum size=7mm, draw] {{\footnotesize\,CA1\,}};
	\draw (1.3, -1.3) node(6) [circle, minimum size=7mm, draw] {{\footnotesize\,Sub\,}};
	
	\draw[-arcsq,blue,very thick] (3) -- (2) node[pos=1.5,sloped,above] {};
	\draw[-arcsq,densely dotted,red,very thick] (3) -- (6) node[pos=1.5,sloped,above] {};
	\draw[-arcsq,densely dotted,red,very thick] (6) -- (5) node[pos=1.5,sloped,above] {};
	\draw[-arcsq,densely dotted,red,very thick] (5) -- (2) node[pos=1.5,sloped,above] {};
	\draw[-arcsq,loosely dotted,blue,very thick] (6) -- (2) node[pos=1.5,sloped,above] {};
	\draw[-arcsq,loosely dotted,blue,very thick] (0,0.4) to[bend left,loosely dotted,blue,very thick] (2.6,0.4);
	\draw[-arcsq,loosely dotted,white,very thick] (0,-1.7) to[bend right,loosely dotted,blue,very thick] (2.6,-1.7);
	\end{tikzpicture}~~~~~~~~~ 
    \begin{tikzpicture}[scale=1.0, line width=0.5pt, inner sep=0.2mm, shorten >=.1pt, shorten <=.1pt]
    \draw (0, -1.3) node(1) [circle, minimum size=7mm, draw] {{\footnotesize\,PRC\,}};
    \draw (2.6,0) node(2) [circle, minimum size=7mm, draw] {{\footnotesize\,CA3\,}};
	\draw (1.3, 0) node(3) [circle, minimum size=7mm, draw] {{\footnotesize\,ERC\,}};
    \draw (0, 0) node(4) [circle, minimum size=7mm, draw] {{\footnotesize\,PHC\,}};
	\draw (2.6, -1.3) node(5) [circle, minimum size=7mm, draw] {{\footnotesize\,CA1\,}};
	\draw (1.3, -1.3) node(6) [circle, minimum size=7mm, draw] {{\footnotesize\,Sub\,}};
	\draw[-arcsq,densely dotted, white,very thick] (2.6,0.4) to[bend right,densely dotted,red,very thick] (0,0.4);
	\draw[-arcsq,loosely dotted,white,very thick] (0,-1.7) to[bend right,loosely dotted,blue,very thick] (2.6,-1.7);
	\end{tikzpicture} 
	}\\
	(a)~~~~~~~~~~~~~~~~~~~~~~~~~~~(b)~~~~~~~~~~~~~~~~~~~~~~~~~~~(c) 
	\caption{Causal structures of fMRI hippocampus data using (a) LiNA, (b) NICA, and (c) Triad methods. Note that Triad output an empty graph, which implied all brain regions are independent with each other. Solid blue lines are consistent edges with the anatomical connectivity while densely dotted red lines are spurious. Loosely dotted blue lines represent redundant edges.}
	\label{fig:application_fMRI}
\end{figure}

\paragraph{fMRI hippocampus dataset.} We investigated causal relations between six brain regions of an individual: perirhinal cortex (PRC), parahippocampal cortex (PHC), entorhinal cortex (ERC), subiculum (Sub), CA1,
and CA3/Dentate Gyrus (DG), each of which had left and right sides and were treated as measurements~\cite{poldrack2015long}. We used the anatomical connectivity between regions as a reference for evaluation~\cite{bird2008hippocampus,ghassami2018multi}. 
From Figure~\ref{fig:application_fMRI}, we see though our method estimated one more redundant edge, we obtained more consistent as well as less spurious edges than NICA, while Triad failed to learn the relations between these regions in this data. Besides, we found our results also coincide with some current findings, e.g., ERC $\to$ CA1 is supposed to correlate with memory loss~\cite{kerchner2012hippocampal}. Please refer to SM H.2 for more details.

\section{Conclusions}
We proposed \underline{M}ulti-\underline{D}omain \underline{Li}near \underline{N}on-Gaussian Acyclic Models for L\underline{A}tent Factors (MD-LiNA) with its identification results, which gave deeper interpretation for latent factors that count. To discover the underlying causal structure for shared latent factors of interest, we proposed an integrated two-phase approach along with its local consistency. 
Our experimental results on simulated data and real-world applications validated the efficacy and robustness of the proposed algorithm.

\section*{Acknowledgments}
This work was supported by the Grant ONR N00014-20-1-2501, the Natural Science Foundation of China (61876043, 61976052), Science and Technology Planning Project of Guangzhou (201902010058), and the Grant of China Scholarship Council. FX would like to acknowledge the support by China Postdoctoral Science Foundation (020M680225) and a research project of Huawei.


\clearpage
\section*{Supplementary Materials to "Causal Discovery with Multi-Domain LiNGAM for Latent Factors"}
In this section, we provide further explanations about,
\begin{itemize}[noitemsep,leftmargin=23.5pt]
    \item [A:] A Simple Coding for MD-LiNA Models;
    \item [B:] Discussions of Assumptions
    \item [C:] Proofs for Lemma 1 and Theorems 1-3;
    \item [D:] Detailed Procedures to Estimate Measurement Models;
    \item [E:] Derivation of the Log-Likelihood Function;
    \item [F:] Complexity Analysis;
    \item [G:] More Details about Synthetic Experiments;
    \item [H:] More Details about Real-World Experiments.
\end{itemize}
Note that Eqs.(1) to~\eqref{eq:optimal_multi_log} refer to the main paper. For brevity, we omit the superscripts of all notations for LiNA models.

\subsection*{A: A Simple Coding for MD-LiNA Models}
Here we show how to deal with the multi-domain data with a simple coding method~\cite{shimodaira2016cross}.

Denote $\bm{x}^{(m)} \in \mathbb{R}^{p_m}$ and $\bm{f}^{(m)} \in \mathbb{R}^{q_m}$ by the random vectors which collect observed variables and latent factors of domain $m$, respectively, where $p_m$ is the number of observed variables and $q_m$ is that of latent factors in domain $m$. Employing the coding method, $\bm{x}^{(m)}$ and $\bm{f}^{(m)}$ then are coded as augmented variables $\bar{\bm{x}} \in \mathbb{R}^{p}$, $p = \sum_{m=1}^M p_m$, and $\bar{\bm{f}} \in \mathbb{R}^{q}$, $q = \sum_{m=1}^M q_m$, which are defined as,
\begin{equation*} 
    \begin{aligned}
        \bar{\bm{x}}^{T} &=\lbrace \textbf{0}_{p_1}^T,...,\textbf{0}_{p_{m-1}}^T,(\bm{x}^{(m)})^T,\textbf{0}_{p_{m+1}}^T,...,\textbf{0}_{p_{M}}^T \rbrace, \\
        \bar{\bm{f}}^{T}&=\lbrace \textbf{0}_{q_1}^T,...,\textbf{0}_{q_{m-1}}^T,(\bm{f}^{(m)})^T,\textbf{0}_{q_{m+1}}^T,...,\textbf{0}_{q_{M}}^T \rbrace,
    \end{aligned}
\end{equation*} 
where $\textbf{0}_{p_i} \in \mathbb{R}^{p_i}$ is a vector with $p_i$ zeros. In principle, now all observations from different domains are coded into the same space with $p$ dimensions, which produce newly observed data $\bar{\bm{X}}^{(m)} \in \mathbb{R}^{p\times n_m}$ for domain $m$'s data matrix $\bm{X}^{(m)} \in \mathbb{R}^{p_m\times n_m}$. Hence, we have the whole observed data matrix $\bar{\bm{X}} \in \mathbb{R}^{p \times n}$, 
\begin{equation}
\label{Eq:augX}
\begin{aligned}
    \bar{\bm{X}} &=\lbrace \bar{\bm{X}}^{(1)},...,\bar{\bm{X}}^{(M)} \rbrace,\\
    &= \mathrm{Diag}(\bm{X}^{(1)},...,{\bm{X}^{(m)}}),
\end{aligned}
\end{equation}
where $\mathrm{Diag}()$ indicates a block diagonal matrix. The data matrix of latent factors for all domains is derived identically as in~Eq.\eqref{Eq:augX}.

Let us consider the MD-LiNA model of Figure~\ref{fig:MD-LiNA} to see the partitions of multi-domain variables. In~Eq.\eqref{eq:MD_LiNA}, we get,
\begin{equation} \tag{\ref{eq:MD_LiNA}}
    \begin{aligned}
        \bm{f}^{(m)}=\bm{B}^{(m)}\bm{f}^{(m)}+\bm{\varepsilon}^{(m)},\\
        \bm{x}^{(m)}=\bm{G}^{(m)}\bm{f}^{(m)}+\bm{e}^{(m)},
    \end{aligned}
\end{equation} 
where $\bm{G}^{(m)}\in \mathbb{R}^{p_m \times q_m} $, and $\bm{B}^{(m)} \in \mathbb{R}^{q_m \times q_m}$ are factor loadings, and causal effects matrix of domain $m$, respectively. Employing the above coding representation method, we treat $\bm{x}^{(m)}$ and $\bm{f}^{(m)}$ as the augmented ones $\bar{\bm{x}}$ and $\bar{\bm{f}}$, and let $\bar{\bm{G}}= \mathrm{Diag}( \bm{G}^{(1)},...,\bm{G}^{(M)}) \in \mathbb{R}^{p\times q}$ and $\bar{\bm{B}}=\mathrm{Diag} (\bm{B}^{(1)},...,\bm{B}^{(M)}) \in \mathbb{R}^{q\times q}$. In this way, we derive, 
\begin{equation} \tag{\ref{eq:augment}}
\begin{aligned} 
\bar{\bm{f}}&=\bar{\bm{B}}\bar{\bm{f}}+\bar{\bm{\varepsilon}},\\
\bar{\bm{x}}&=\bar{\bm{G}} \bar{\bm{f}}+ \bar{\bm{e}},
\end{aligned}
\end{equation}
where both observed variables and latent factors from different domains lie in a common space. Besides, its data matrix form can be derived similarly.

\subsection*{B: Discussions of Assumptions}
We discuss the plausibility of MD-LiNA's assumptions. Firstly, MD-LiNA is defined as,
\begin{repdefinition}{def:lina}[MD-LiNA] 
    An MD-LiNA model satisfies the following assumptions:
    \begin{itemize}[noitemsep,topsep=-3pt,leftmargin=20pt]
    \item [A1.] $\bm{f}^{(m)}$ are generated linearly from a Directed Acyclic Graph (DAG) with non-Gaussian distributed external variables $\bm{\varepsilon}^{(m)}$, as in~Eq.\eqref{eq:MD_LiNA};
    \item [A2.] $\bm{x}^{(m)}$ are generated linearly from $\bm{f}^{(m)}$ plus Gaussian distributed errors $\bm{e}^{(m)}$, as in~Eq.\eqref{eq:MD_LiNA};
    \item [A3.] Each ${f}_i$ has at least 2 pure measurement variables;
    \item [A4.] Each $\Bar{{f}_i}^{(m)}$ is linearly generated by only one latent in $\Tilde{\bm{f}}$ and each $\Tilde{{f}}_i$ generates at least one latent in $\Bar{\bm{f}}$, as in~Eq.\eqref{eq:fandf}.
\end{itemize}
\end{repdefinition}

Note that assumptions A1 and A2 are part of the assumptions from NICA~\cite{shimizu2009estimation}; A3 is from Triad~\cite{cai2019triad}, which is milder than previous models (e.g., NICA), since it only needs 2 pure variables while NICA needs at least 3. Assumptions A1 - A3 guarantee the full identification of causal effects matrix $\bm{B}$ in LiNA model; A4 is derived to assure the full identification of the MD-LiNA model, which facilitates the learning of matrix $\tilde{\bm{B}}$, i.e., it helps find the shared information from latent factors $\bar{\bm{f}}$ of different domains and clarify the concepts of factors of interest $\tilde{\bm{f}}$. 

Assumption A4 implies that each row of $\bm{H}$ has only one non-zero element and each column has at least one non-zero element, which is explicable and mild. In particular, in the context of the ordinary factor analysis, using rotation techniques, it estimates factor loadings that are interpretable, and the interpretability is defined as simplicity. The perfect simple structure, which is the simplest possible structure, is almost identical to the assumption A4. References about simplicity and perfect simple structure include~\citeauthor{browne2001overview}~\shortcite{browne2001overview} and~\citeauthor{jennrich2006rotation}~\shortcite{jennrich2006rotation}.
Thus,  MD-LiNA's assumptions are plausible and explicable.

The key differences to the mostly common latent factor models are that $\bm{f}^{(m)}$ here may be causally related rather than being independent;
further, we introduce $\tilde{\bm{f}}$ and assumption A4, which ensure the identification of causal structure among shared latent factors of interest.

\subsection*{C: Proofs}
\subsubsection{C.1 Proof of Lemma 1}
We first quote the definition of Triad constraints~\cite{cai2019triad} since it is used in Lemma~\ref{theo:triad}. 
\begin{definition}[Triad constraints] \label{{def:triad}}
    Suppose ${x}_i$, ${x}_j$ and ${x}_k$ are distinct and correlated variables and that all noise variables are non-Gaussian. Define the pseudo-residual of $\lbrace {x}_i, {x}_j \rbrace$ relative to ${x}_k$, which is called a reference variable, as
    \begin{equation} \label{eq:def_triad}
    {E}_{(i,j|k)} = {x}_i-\frac{\mathrm{cov} ({x}_i,{x}_k)}{\mathrm{cov} ({x}_j,{x}_k) } \cdot {x}_j.
\end{equation}
    We say that $\lbrace {x}_i, {x}_j \rbrace$ and $\lbrace {x}_k \rbrace$ satisfy Triad constraints if and only if ${E}_{(i,j|k)} \!\perp\!\!\!\perp {x}_k$, i.e., $\lbrace {x}_i, {x}_j \rbrace$ and $\lbrace {x}_k \rbrace$ violate Triad constraints if and only if ${E}_{(i,j|k)} \not\!\perp\!\!\!\perp {x}_k$.
\end{definition}
Note that the Triad constraints assume that all noises in Eq.\eqref{eq:MD_LiNA} including $\bm{e}$ are non-Gaussian distributed while in our models, noises $\bm{\varepsilon}$ are assumed to be non-Gaussian but $\bm{e}$ are Gaussian.

\begin{figure}[t]  
	\centering
	\begin{tikzpicture}[scale=0.8, line width=0.5pt, inner sep=0.2mm, shorten >=.1pt, shorten <=.1pt]
        \draw (-2, 0) node(1) [circle, minimum size=7mm, draw] {{\footnotesize\,${f}_1$\,}};
        \draw (0, 0) node(2) [circle, minimum size=7mm, draw] {{\footnotesize\,${f}_2$\,}};

        \draw (-2, -1.4) node(3) [rectangle, minimum size=4mm, draw] {{\footnotesize\,${x}_i$\,}};
        \draw (-0.5, -1.4) node(4) [rectangle, minimum size=4mm, draw] {{\footnotesize\,${x}_j$\,}};
        \draw (0.5, -1.4) node(5) [rectangle, minimum size=4mm, draw] {{\footnotesize\,${x}_k$\,}};
        
        \draw[-arcsq] (1) -- (2) node[pos=0.5,sloped,above] {$\beta$};
        \draw[-arcsq] (1) -- (3) node[pos=0.5,sloped,above] {$\gamma$};
        \draw[-arcsq] (2) -- (4) node[pos=0.5,sloped,above] {$\theta$};
        \draw[-arcsq] (2) -- (5) node[pos=0.5,sloped,above] {$\eta$};
    \end{tikzpicture}~~~~~~~~~ 
    \begin{tikzpicture}[scale=0.8, line width=0.5pt, inner sep=0.2mm, shorten >=.1pt, shorten <=.1pt]
        \draw (-2, 0) node(1) [circle, minimum size=7mm, draw] {{\footnotesize\,${f}_1$\,}};
        \draw (0, 0) node(2) [circle, minimum size=7mm, draw] {{\footnotesize\,${f}_2$\,}};

        \draw (-2, -1.4) node(3) [rectangle, minimum size=4mm, draw] {{\footnotesize\,${x}_i$\,}};
        \draw (-0.5, -1.4) node(4) [rectangle, minimum size=4mm, draw] {{\footnotesize\,${x}_j$\,}};
        \draw (0.5, -1.4) node(5) [rectangle, minimum size=4mm, draw] {{\footnotesize\,${x}_k$\,}};
        
        \draw[-arcsq] (2) -- (1) node[pos=0.5,sloped,above] {$\beta$};
        \draw[-arcsq] (1) -- (3) node[pos=0.5,sloped,above] {$\gamma$};
        \draw[-arcsq] (2) -- (4) node[pos=0.5,sloped,above] {$\theta$};
        \draw[-arcsq] (2) -- (5) node[pos=0.5,sloped,above] {$\eta$};
	\end{tikzpicture} \\
	(a)~~~~~~~~~~~~~~~~~~~~~~~~~~~~~~~~~~~~~(b) 
	\caption{Identification of causal structures with two latent factors using Triad constraints, when errors $\bm{e}$ of observed variables $\bm{x}$ are Gaussian distributed.}
	\label{fig:theorem1} 
\end{figure}
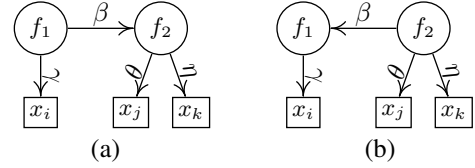

\begin{replemma}{theo:triad}
    Assume that the input data $\bm{X}$ strictly follow the LiNA model. 
    Then the factor loading matrix ${\bm{G}}$ is identifiable up to permutation and scaling of columns and the causal effects matrix ${{\bm{B}}}$ is fully identifiable.
\end{replemma}
\begin{proof}

We prove the lemma by proving the corollary firstly that Triad constraints also hold even when the $\bm{e}$ are normally distributed in our models. That is,
\begin{corollary} \label{coro:triad} 
    Assume that the input data $\bm{X}$ strictly follow the LiNA model. Let ${f}_1$ and ${f}_2$ be directed connected latent factors without confounders and let $\lbrace {x}_i \rbrace$ and $\lbrace {x}_j, {x}_k\rbrace$ be their pure measurement variables.
    Then $\lbrace {x}_i, {x}_j\rbrace$ and $\lbrace {x}_k \rbrace$ violate the Triad constraint if and only if ${f}_1 \to {f}_2$ holds.
\end{corollary}

We prove by two aspects. i) If ${f}_1 \to {f}_2$ holds, then  $\lbrace {x}_i, {x}_j\rbrace$ and ${x}_k$ violate the Triad constraints, as shown in Figure~\ref{fig:theorem1} (a). According to the assumptions of our models, we have,
\begin{equation}\label{eq:theom_triad0}
    \begin{aligned}
    {f}_1  &= {\varepsilon}_1,\\
    {f}_2 &= \beta{\varepsilon}_1 + {\varepsilon}_2,\\
    {x}_i &= \gamma {\varepsilon}_1 + {e}_i, \\
    {x}_j &= \beta \theta {\varepsilon}_1 + \theta {\varepsilon}_2 + {e}_j , \\
    {x}_k &= \beta \eta {\varepsilon}_1 + \eta {\varepsilon}_2 + {e}_k, \\
    \end{aligned}
\end{equation}
and the pseudo-residual of $\lbrace {x}_i, {x}_j\rbrace$ relative to ${x}_k$ defined in~Eq.\eqref{eq:def_triad}
is,
\begin{equation} \label{eq:theom_triad1}
    \begin{aligned}
    {E}_{(i,j|k)} &= {x}_i-\frac{\mathrm{cov} ({x}_i,{x}_k)}{\mathrm{cov} ({x}_j,{x}_k) } \cdot {x}_j, \\
    &= \gamma {\varepsilon}_1 + {e}_i - t \cdot (\beta \theta {\varepsilon}_1 + \theta {\varepsilon}_2 + {e}_j) ,\\
    &= (\gamma-t\beta \theta) \cdot{\varepsilon}_1 + t\theta {\varepsilon}_2 + {e}_i + t {e}_j,\\
    \end{aligned} 
\end{equation}
where $t=\frac{\gamma \beta \cdot\mathrm{var}({\varepsilon}_1)}{\beta ^2 \theta  \cdot \mathrm{var}({\varepsilon}_1) + \theta  \cdot \mathrm{var}({\varepsilon}_2) } \neq 0$, since all the causal effects $\beta,\gamma, \theta$ and $\eta$ are not equal to 0 in non-trivial cases. Considering ${E}_{(i,j|k)}$ and ${x}_k$ in Eqs.(\ref{eq:theom_triad0}) and (\ref{eq:theom_triad1}), we find that they both obtain at least one common non-Gaussian component ${\varepsilon}_2$, where $t \theta \neq 0$ and $\eta \neq 0$. Thus, by virtue of the Darmois-Skitovich theorem~\cite{darmois1953analyse,skitovitch1953property}, we get ${E}_{(i,j|k)} \not\!\perp\!\!\!\perp {x}_k$, which implies that $\lbrace {x}_i, {x}_j\rbrace$ and $\lbrace {x}_k \rbrace$ violate the Triad constraint. We see here that the dependence is owing to the non-Gaussianity from noises $\bm{\varepsilon}$, rather than errors $\bm{e}$.

ii) If ${f}_2 \to {f}_1$ holds, then $\lbrace {x}_i, {x}_j\rbrace$ and ${x}_k$ satisfy the Triad constraints, as shown in (b) of Figure~\ref{fig:theorem1}. Similarly, we obtain,
\begin{equation} \label{eq:theom_triad2}
    \begin{aligned}
    {f}_2  &= {\varepsilon}_2,\\
    {f}_1 &= \beta{\varepsilon}_2 + {\varepsilon}_1,\\
    {x}_i &= \beta \gamma {\varepsilon}_2 + \gamma {\varepsilon}_1 + {e}_i, \\
    {x}_j &= \theta {\varepsilon}_2 + {e}_j , \\
    {x}_k &= \eta {\varepsilon}_2 + {e}_k, \\
    \end{aligned}
\end{equation}
and the pseudo-residual of $\lbrace {x}_i, {x}_j\rbrace$ relative to ${x}_k$ is,
\begin{equation} \label{eq:theom_triad3}
    \begin{aligned}
    {E}_{(i,j|k)} &= {x}_i-\frac{\mathrm{cov} ({x}_i,{x}_k)}{\mathrm{cov} ({x}_j,{x}_k) } \cdot {x}_j, \\
    &= \beta \gamma {\varepsilon}_2 + \gamma {\varepsilon}_1 + {e}_i - \frac{\beta \gamma}{\theta} \cdot (\theta {\varepsilon}_2 + {e}_j ) ,\\
    &= \gamma {\varepsilon}_1 -  \frac{\beta \gamma}{\theta}{e}_j.\\
    \end{aligned} 
\end{equation}
Due to the mutual independence between $\bm{\varepsilon}$ and $\bm{e}$, i.e., there are no common components shared by ${E}_{(i,j|k)}$ and ${x}_k$, we come to the conclusion of ${E}_{(i,j|k)} \!\perp\!\!\!\perp {x}_k$. It implies that the Triad constraint is satisfied. Thus, the corollary is proved.

Corollary~\ref{coro:triad} implies that using the asymmetry from Triad constraints into LiNA, one can find pure measurement variables and locate the latent factors. Thus, based on Corollary~\ref{coro:triad} and the identifiability from~\citeauthor{shimizu2009estimation}~\shortcite{shimizu2009estimation}, the identifiability of LiNA is guaranteed.
\end{proof}

\subsubsection{C.2 Proof of Theorem~\ref{theo:MD-LiNA}}
\begin{reptheorem} {theo:MD-LiNA} 
	Assume that the input multi-domain data $\bm{X}$ with ${\bm{X}}^{(m)}$ of domain $m$, strictly follow the MD-LiNA model.
	Then the underlying factor loading matrix $\bar{\bm{G}}$ is identifiable up to permutation and scaling of columns and the causal effects matrix ${\tilde{\bm{B}}}$ is fully identifiable.
\end{reptheorem}
\begin{proof}
First note that if the number of domains is $M=1$, i.e., the observed data are single-domain, the identifiability of the model has been proved by~\citeauthor{shimizu2009estimation}~\shortcite{shimizu2009estimation}, assuming that each latent factor has at least three pure measurement variables. Fortunately, due to the presence of Triad constraints~\cite{cai2019triad}, this assumption is relaxed to the one which needs only two pure measurement variables for each latent, as demonstrated in Corollary 1. In other words, it is proved that if the single-domain data ${\bm{X}}$ strictly follow the LiNA model, the underlying factor loading matrix ${\bm{G}}$ is identifiable up to permutation and scaling of columns and the causal effects matrix ${\bm{B}}$ is fully identifiable.
With the single-domain LiNA model's identification, here we ought to prove the case with multiple domains, which is the generalization of single-domain ones.

Firstly,
~Eq.\eqref{eq:augment} demonstrates,
\begin{equation}\tag{\ref{eq:augment}}
\begin{aligned}
\bar{\bm{f}}&=\bar{\bm{B}}\bar{\bm{f}}+\bar{\bm{\varepsilon}},\\
\bar{\bm{x}}&=\bar{\bm{G}} \bar{\bm{f}}+ \bar{\bm{e}},
\end{aligned}
\end{equation}
where $\bar{\bm{G}}= \mathrm{Diag}( \bm{G}^{(1)},...,\bm{G}^{(M)}) \in \mathbb{R}^{p\times q}$ and $\bar{\bm{B}}=\mathrm{Diag} (\bm{B}^{(1)},...,\bm{B}^{(M)}) \in \mathbb{R}^{q\times q}$. 
Since we estimate one measurement model for all domains simultaneously using the augmented multi-domain data, $\bar{\bm{G}}$ is identifiable up to permutation and scaling of columns.
However, since causal relationships are embedded by the factors of interest $\Tilde{\bm{f}}$, our focus is on estimating the causal structure among $\tilde{\bm{B}}$. We suppose in~Eq.\eqref{eq:fandf}, 
\begin{equation} \tag{\ref{eq:fandf}}
\begin{aligned}
\bar{\bm{f}}=\bm{H}\tilde{\bm{f}},
\end{aligned}
\end{equation}
where $\bm{H} \in \mathbb{R}^{{q} \times \tilde{q}}$ tells that which factors from different domains are supposed to stand for which same concepts. Combining~Eqs.\eqref{eq:augment} and~\eqref{eq:fandf}, we obtain,
\begin{equation}
\begin{aligned}
\Tilde{\bm{f}}=\Tilde{\bm{B}} \Tilde{\bm{f}}+\Tilde{\bm{\varepsilon}},
\end{aligned}
\end{equation}
where $\Tilde{\bm{B}} = (\bm{H}^T\bm{H})^{-1}\bm{H}^T\bar{\bm{B}}\bm{H} \in \mathbb{R}^{\tilde{q} \times \tilde{q}}$ and $\Tilde{\bm{\varepsilon}}= (\bm{H}^T\bm{H})^{-1}\bm{H}^T\bar{\bm{\varepsilon}}\in \mathbb{R}^{\tilde{q}}$. It is worth noting that the inverse matrix of $\bm{H}^T\bm{H}$ always exists due to the assumption A4. We see that this is a structure model between the actual latent factors of interest. To prove it is in accordance with the single-domain LiNA model such that $\Tilde{\bm{B}}$ is identifiable, we have to ensure that $\Tilde{\bm{B}}$ can be permuted to a strictly lower triangular matrix and $\Tilde{\bm{\varepsilon}}$ are independent with each other. Fortunately, due to assumption A1, $\bm{B}^{(m)}$ as well as $\bar{\bm{B}}$ can be permuted to a strictly lower triangular matrix. Then because of $\Tilde{\bm{B}} = (\bm{H}^T\bm{H})^{-1}\bm{H}^T\bar{\bm{B}}\bm{H}$ and assumption A4, $\Tilde{\bm{B}}$ also satisfies the condition since it follows the acyclicity constraint as suggested in~\citeauthor{zheng2018dags}~\shortcite{zheng2018dags}, i.e., $h(\tilde{\bm{B}})= \mathrm{tr}(e^{\tilde{\bm{B}} \circ \tilde{\bm{B}}}) - \tilde{q}=0 $. Note that each row of $\bm{H}$ has only one non-zero element and each column has at least one non-zero element due to the assumption A4, we know that actually $\tilde{\bm{\varepsilon}}$ are linear combinations of $\bar{\bm{\varepsilon}}$ and each external variable $\tilde{{\varepsilon}}_i$ has different components in $\bar{\bm{\varepsilon}}$. So, by virtue of the independence between $\bar{\bm{\varepsilon}}$, its non-Gaussianity and the Darmois-Skitovich theorem~\cite{kagan1973characterization}, $\Tilde{\bm{\varepsilon}}$ are also independent with each other. In this way,~Eq.(\ref{eq:MD_LiNA}) is rewritten as,
\begin{equation}\label{eq:tilde1}
\begin{aligned}
\Tilde{\bm{f}}=\Tilde{\bm{B}}\Tilde{\bm{f}}+\Tilde{\bm{\varepsilon}},\\ 
\tilde{\bm{x}}=\tilde{\bm{G}} \tilde{\bm{f}}+ \tilde{\bm{e}},
\end{aligned}
\end{equation}
where $\tilde{\bm{x}} = \bar{\bm{x}}$, $ \tilde{\bm{e}}=\bar{\bm{e}}$ and $\tilde{\bm{G}} = \bar{\bm{G}}\bm{H}$. It implies that 
${\tilde{\bm{B}}}$ is fully identifiable.
Thus, the theorem is proved.
\end{proof}

\subsubsection{C.3 Proof of Theorem~\ref{theo:consistent1}}
The quadratic penalty function from QPM is given by
\begin{equation} 
    \begin{aligned}
        \mathcal{S}(\bm{B}) = \mathcal{F}(\bm{B},\hat{\bm{G}}) + \frac{\rho}{2}h(\bm{B})^2.
    \end{aligned} \label{eq:qpm}
\end{equation}
We give the following mild conditions, which are used in Theorem~\ref{theo:consistent1}.
\begin{itemize}
    \item [] {\it{Mild Conditions:}}
\end{itemize}
\begin{itemize}[noitemsep,leftmargin=27pt]
    \item [C0.] The sequence of non-negative tolerances $\lbrace \tau_k \rbrace$ is bounded, where the gradient of Eq.~\eqref{eq:qpm} satisfies $\Arrowvert \nabla_{\bm{B}} \mathcal{S}(\bm{B}) \Arrowvert_{F} \leq \tau_k$ and $k$ is the iteration time.
    \item [C1.] The acyclicity constraint $h(\bm{B})$ satisfies
    \begin{equation*}
        h(\bm{B}) = 0 \Longleftrightarrow \nabla_{\bm{B}}h(\bm{B})=0.
    \end{equation*} 
\end{itemize}

\begin{reptheorem}{theo:consistent1}
    Assume that the input single-domain data $\bm{X}$ strictly follow the LiNA model. Suppose the sample size $n$, the number of observed variables $p$ and the penalty coefficient $\rho$ satisfy $n,p,\rho \to \infty$. Then under conditions given in C0 \& C1, our method using QPM with
    ~Eq.\eqref{eq:optimal_elastic}, is consistent and locally consistent to learn ${{\bm{G}}}$ and ${{\bm{B}}}$, respectively. 
\end{reptheorem}
\begin{proof}
We prove the theorem by proving that as $n, p, \rho \to \infty$, under conditions C0 \& C1, our method guarantees to learn the factor loading ${{\bm{G}}}$ consistently and the weighted adjacency matrix ${{\bm{B}}}$ locally consistently up to its DAG solution.

First we show that our score-based constrained optimization function in~Eq.\eqref{eq:optimal_elastic} basically contains three parts: the log-likelihood of measurement models, denoted by $\mathcal{C}_{0}= \sum_{t=1}^n  \frac{1}{2} \left \| \bm{X}(t)-\hat{\bm{G}}\Tilde{\hat{\bm{G}}}^T\bm{X}(t)\right \| ^2_{\Sigma^{-1}}$; the log-likelihood of LiNGAM, additionally with an acyclicity constraint $h(\bm{B})=0$, denoted by $\mathcal{C}_{1}=\sum_{t=1}^n \sum_{i=1}^{q} \log \hat{p}_i(\bm{g}_i^T\bm{X}(t)-\bm{b}_i^T\Tilde{\hat{\bm{G}}}^T\bm{X}(t))$; and the sparse or elastic net regularization, denoted by $\mathcal{C}_{2}$. Note that as $p\to \infty$, estimations of latent factors in line~\ref{algo:f} of Algorithm 1 are validated~\cite{hyvarinen2001independent}.

Since confirmatory factor analysis is consistent with unweighted least squares estimators, 
when $n, p \to \infty$~\cite{long1983confirmatory}, $\mathcal{C}_{0}$ as a score function is consistent to estimate the factor loadings. 
Furthermore,~\citeauthor{ng2020convergence}~\shortcite{ng2020convergence} has shown that if conditions C0 \& C1 are satisfied and as $\rho \to \infty$, the estimators $\hat{{\bm{B}}}$  of $\mathcal{C}_{1}$ using QPM are guaranteed to be the DAG solutions, which is locally consistent.
The adaptive sparsity constraint $\Arrowvert \bm{B} \Arrowvert_{1*}$ is a special case of adaptive Lasso, which has oracle properties, including the consistency in variable selection~\cite{hyvarinen2010estimation,zou2006adaptive}. Furthermore, the elastic net regularization gains consistency as well~\cite{zou2005regularization}. Thus, $\mathcal{C}_{2}$ in our score function is consistent for estimation. Eventually, according to the Slutsky’s theorem~\cite{goldberger1964econometric}, our whole score function in~Eq.\eqref{eq:optimal_elastic} is consistent and locally consistent to learn $\hat{{\bm{G}}}$ and $\hat{{\bm{B}}}$, respectively. The theorem is proven.
\end{proof}

\subsubsection{C.4 Proof of Theorem~\ref{theo:consistent2}}

Let $\tilde{\bm{f}}(t)$ be the $t^{th}$ column (observation) of the true factors $\tilde{\bm{f}}$ and $\bm{H}$ be the true factor loadings. $\bm{H}_i$ is the $i^{th}$ column of $\bm{H}^T$. Let $\Arrowvert \bm{X} \Arrowvert = \lbrack \mathrm{tr} (\bm{X}^T\bm{X}) \rbrack ^{1/2}$ denote the norm of matrix $\bm{X}$.

We give the following mild conditions, which are used in Theorem~\ref{theo:consistent2}.
\begin{itemize}
    \item [] {\it{Mild Conditions:}}
\end{itemize}
\begin{itemize}[noitemsep,leftmargin=27pt]
    \item [C2.] $\mathop{{}\mathbb{E}}\Arrowvert \tilde{\bm{f}}(t) \Arrowvert^4 < \infty$ and $n^{-1}\sum_{t=1}^{n} \tilde{\bm{f}}(t) \tilde{\bm{f}}(t)^T \to \Sigma_{\tilde{\bm{f}}}$ as $n \to \infty$ for some $\tilde{q} \times \tilde{q}$ positive definite matrix $\Sigma_{\tilde{\bm{f}}}$.
    \item [C3.] $\Arrowvert \bm{H}_i \Arrowvert < \infty$, and $\Arrowvert \bm{H}^T\bm{H} /p - \Sigma_{\bm{H}} \Arrowvert \to 0$ as $p\to \infty$ for some $\tilde{q}\times \tilde{q}$ positive definite matrix $\Sigma_{\bm{H}}$.
    \item [C4.] There exists a positive constant $C < \infty$ such that $\mathop{{}\mathbb{E}} \arrowvert (\bar{\bm{f}}-\bm{H}\tilde{\bm{f}})_{ij}\arrowvert ^{8} \leq C$ ($i=1,...,q$; $j=1,...,n$).
    \item [C5.] The eigenvalues of the matrix $(\Sigma_{\bm{H}} \Sigma_{\tilde{\bm{f}}})$ are distinct. 
\end{itemize}

\begin{reptheorem}{theo:consistent2} 
    Assume that the input multiple-domain data $\bm{X}$ with ${\bm{X}}^{(m)}$ strictly follow the MD-LiNA model.
    Suppose the sample size $n_m$, the number of observed variables $p_m$ of each domain $m$ and the penalty coefficient $\rho$ satisfy
    $n_m, p_m, \rho \to \infty$. Then under conditions given in C0-C5,
    our method using QPM with
    ~Eq.\eqref{eq:optimal_multi}, is consistent to learn $\bar{\bm{G}}$ and ${{\bm{H}}}$, and locally consistent to learn ${\tilde{\bm{B}}}$.
\end{reptheorem}

\begin{proof}
We prove the theorem by proving that as $n_m, p_m, \rho \to \infty$, under conditions C0 - C5, our method guarantees to learn the factor loadings $\bar{{\bm{G}}}$ and the transformation matrix ${{\bm{H}}}$ consistently, as well as the weighted adjacency matrix ${\tilde{\bm{B}}}$ among latent factors of interest locally consistently up to its DAG solution.

First we show that apart from the three parts in~Eq.\eqref{eq:optimal_elastic} in Theorem~\ref{theo:consistent1} ($\mathcal{C}_{0}$, $\mathcal{C}_{1}$ and $\mathcal{C}_{2}$), our score-based constrained optimization function for MD-LiNA model in~Eq.\eqref{eq:optimal_multi} also consists of another part: the construction error $\mathcal{E}(\bm{H})$, denoted by $\mathcal{C}_{3}$. Note that $\mathcal{C}_{0}$ and $\mathcal{C}_{1}$ here are similar with those in Theorem~\ref{theo:consistent1} while $\mathcal{C}_{2}$ is the same. That is, $\mathcal{C}_{0} = \sum_{t=1}^n \left \| \bar {\bm{X}}(t)-\bar{\bm{G}}\Tilde{\bar{\bm{G}}}^{T}\bar {\bm{X}}(t)\right \| ^2_{\Sigma^{-1}}$ while $\mathcal{C}_{1}$ equals to $\sum_{t=1}^n \sum_{i=1}^{\tilde{q}} \log \hat{p}_i(\bm{h}_i^T\bar{\bm{f}}(t)-\tilde{\bm{b}}_i^T\Tilde{\hat{\bm{H}}}^T\bar{\bm{f}}(t))$. Since the measurement models are processed using augmented data $\bar{\bm{X}}$, which can be seen as an augmented single-domain measurement model, it implies that $\mathcal{C}_{0}$ is consistent to estimate $\bar{\bm{G}}$ as $n_m, p_m \to \infty$ ($n, p \to \infty$)~\cite{long1983confirmatory}. As $p \to \infty$, estimations of latent factors $\bar{\bm{f}}$ are validated~\cite{hyvarinen2001independent}.
Furthermore, $\mathcal{C}_1$ with the acyclicity constraint also characterizes the log-likelihood function of LiNGAM between $\tilde{\bm{f}}$, which is identical to that of Theorem~\ref{theo:consistent1} substantially. Thus,~\citeauthor{ng2020convergence}~\shortcite{ng2020convergence} has shown that if conditions C0 \& C1 are satisfied and as $\rho \to \infty$, the estimators $\hat{\tilde{\bm{B}}}$ of $\mathcal{C}_{1}$ using QPM are guaranteed to be the DAG solutions, which is locally consistent.
The consistency proof of $\mathcal{C}_{2}$ is the same as that of Theorem~\ref{theo:consistent1}. 

We focus on the consistency of $\mathcal{C}_{3}$. Since minimizing the reconstruction errors $\mathcal{E}(\bm{H})$ in a factor model is consistent under the condition $n,p \to \infty$ as $n_m, p_m \to \infty$ and those given in C2 - C5~\cite{bai2003inferential}, we know $\mathcal{C}_{3}$ is consistent to estimate $\hat{\bm{H}}$. 
Thus, according to the Slutsky’s theorem~\cite{goldberger1964econometric}, our whole score function in~Eq.\eqref{eq:optimal_multi} is (locally) consistent to learn $\bar{{\bm{G}}}$, ${\bm{H}}$ and ${\tilde{\bm{B}}}$. To be concluded, the theorem is proven.
\end{proof}

\subsection*{D: Detailed Procedures to Estimate Measurement Models}
We have several approaches for estimating measurement models, i.e., factor loading matrix $\bar{\bm{G}}$.
Firstly, we can employ the Confirmatory Factor Analysis (CFA) approach. In this case, the existence of pure variables is assumed. Given the instrument design and/or empirical domain knowledge, existing methods, e.g., Triad constraints~\cite{cai2019triad} can be applied to find the pure measurement variables and the number of latent factors. Once pure variables are identified, we subsequently do CFA to estimate the measurement model, which constrains the pure measurement variables to be pure~\cite{asparouhov2009exploratory,reilly1996identification}. Secondly, more exploratory approaches are advocated, e.g., Exploratory Structural Equation Modeling (ESEM) approach, which is a general and frequently used multivariate analysis technique in statistics~\cite{asparouhov2009exploratory}. It enables us to use fewer restrictions on estimating factor loadings, with which we could name factors, i.e., which factor influences which observed variables substantially or not.

In our paper, we take the first approach, but we can use the second approach as well in our framework. That is, as illustrated in lines 1 to 3 of Algorithm~\ref{alg:Framwork_b}, we utilize Triad constraints to locate the latent factors. In practice, we often directly assume that the pure measurement variables have been known a priori by virtue of the background domain knowledge or empirical research~\cite{bollen1989structural} or
have been investigated correctly by Triad constraints in the experiments (with reasons in the following paragraph).
Thereafter we get the estimated latent factors $\bar{\bm{f}}$. Note that for multi-domain data, we estimate the measurement models for all domains simultaneously.

In reality, the assumption that the pure measurement variables have been known a priori by virtue of the background domain knowledge or empirical research~\cite{bollen1989structural} is ubiquitous in various applications~\cite{byrne2010structural}. For example, psychologists would argue for the loading of indicators designed to measure the specific factor like coping, but not on the depression or anxiety. Besides, incorporating a priori substantive knowledge in the form of restrictions on the measurement model makes the definition of the latent variables better grounded in subject-matter theory and leads to parsimonious models~\cite{asparouhov2009exploratory}.

\subsection*{E: Derivation of the Log-Likelihood Function}
\subsubsection{E.1 The log-likelihood function for LiNA models}
From~Eq.(\ref{eq:MD_LiNA}) with $M=1$, we can obtain a reduced model,
\begin{equation}
    \begin{aligned}
        \bm{x} = \bm{A} \bm{\varepsilon} + \bm{e},
    \end{aligned}
\end{equation}
where $\bm{A}=\bm{G}(\bm{I}-\bm{B})^{-1}$. This is a noisy Independent Component Analysis (noisy ICA) model, which is characterized that  the $p$-variate vector of observations $\bm{X}$ are generated from a set of $q$ statistically independent non-Gaussian external influences $\bm{\varepsilon}$ via a linear instantaneous mixing matrix $\bm{A}$ plus additive errors $\bm{e}$. 

We now derive the log-likelihood. Assume the errors $\bm{e}$ to be Gaussian with known covariance matrix $\Sigma$ and vanishing means, which are $\bm{P}(\bm{e}) = \mathcal{G}(\bm{0},\Sigma)$. Since $\bm{e}$ are independent with each other, $\Sigma$ is diagonal. Given the external influences $\bm{\varepsilon}$ and the errors $\bm{e}$, the conditional distribution of the data matrix $\bm{X}$ thus is, 
\begin{equation} 
    \begin{aligned}
        \bm{P} ( \bm{X}|\bm{\varepsilon},\bm{e}) = \delta(\bm{X}-(\bm{A}\bm{\varepsilon}+\bm{e})),
    \end{aligned}
\end{equation}
which is a delta distribution in that the data $\bm{X}$ are full determined by the given quantities. Marginalizing out the errors $\bm{e}$, the conditional distribution of $\bm{X}$ given $\bm{\varepsilon}$ subsequently becomes,
\begin{equation} 
    \begin{aligned}
        \bm{P} ( \bm{X}|\bm{\varepsilon}) &= \int \delta(\bm{X}-(\bm{A}\bm{\varepsilon}+\bm{e})) \mathcal{G}(\bm{0},.\Sigma)\,\mathrm{d}e,\\
        &=\mathcal{G}(\bm{X}-\bm{A}\bm{\varepsilon},\Sigma).
    \end{aligned}
\end{equation}
Denote the density function of the independent external influences ${\varepsilon}_i$ by $p_i$. Due to the independent relationships between $\bm{\varepsilon}$, we have $\bm{P} (\bm{\varepsilon}) = \prod_{i=1}^{q} p_i({\varepsilon}_i)$.  Thus one obtains the joint distribution of the data $\bm{X}$ and the external influences $\bm{\varepsilon}$:
\begin{equation} 
    \begin{aligned}\label{eq:joint_dis}
        \bm{P} (\bm{X}, \bm{\varepsilon}) &= \bm{P} ( \bm{X}|\bm{\varepsilon}) \bm{P} (\bm{\varepsilon}),\\
        &=\mathcal{G}(\bm{X}-\bm{A}\bm{\varepsilon},\Sigma)\prod_{i=1}^{q} p_i({\varepsilon}_i).
    \end{aligned}
\end{equation}
Taking the logarithm provides us with the log-likelihood of the whole model, which is,
\begin{equation} \label{eq:L0}
    \begin{aligned}
        &\log (\bm{P} (\bm{X}, \bm{\varepsilon}))\\
        =& \sum_{t=1}^n \bigg [ \log \mathcal{G}(\bm{X}(t)-\bm{A}\bm{\varepsilon}(t),\Sigma) + \sum_{i=1}^{q} \log {p_i}(\varepsilon_i(t)) \bigg ],\\
        = &\sum_{t=1}^n \bigg [ \frac{1}{2} \left \| \bm{X}(t)-\bm{A}\bm{\varepsilon}(t)\right \| ^2_{\Sigma^{-1}}  + \sum_{i=1}^{q} \log {p_i}(\varepsilon_i(t)) \bigg ] \\
        & + C, \\
    \end{aligned}
\end{equation}
where $\left \| \bm{x} \right \|^2_{\Sigma^{-1}} = \bm{x}^T \Sigma^{-1}\bm{x} $, $\bm{X}(t)$ and $\bm{\varepsilon}(t)$ are realizations of the data and the external influences, i.e., the $t^{th}$ column of $\bm{X}$ and $\bm{\varepsilon}$, respectively, and $\bm{\varepsilon}(t) = \lbrace \varepsilon_1(t),\varepsilon_2(t),...,\varepsilon_q(t) \rbrace^T $. $C$ is a constant. This likelihood is essentially identical as the one proposed by~\cite{hyvarinen1998independent}. It not only belongs to the noisy ICA model, but includes the measurement and the structure models. Noisy ICA model estimates $\bm{A}$ and $\bm{\varepsilon}$ while ours estimates the measurement loadings $\bm{G}$ as well as the  structure matrix $\bm{B}$. Thus,~Eq.(\ref{eq:L0}) has to be rewritten as a likelihood of $\bm{G}$ and $\bm{B}$, represented as $\mathcal{L}(\bm{B},\bm{G})$. Details are as follows.

Given the structure model in~Eq.(\ref{eq:MD_LiNA}) (the second equation) and the estimated factor loadings $\hat{\bm{G}}$, we have the estimated latent factors $\hat{\bm{f}} \doteq (\hat{\bm{G}}^T \hat{\bm{G}})^{-1}\hat{\bm{G}}^{T} \bm{X}$~\footnote{This is validated when the measurement errors $\bm{e}_i$ are small enough or there are lots of measurement variables $\bm{X}_i$ for each latent factor by~\cite{hyvarinen2001independent}} and $\hat{\bm{\varepsilon}} = (\bm{I}-\bm{B})\hat{\bm{f}}$. For simplicity, we re-express the estimated latent factors as $\hat{\bm{f}} \triangleq \Tilde{\hat{\bm{G}}}^T \bm{X}$, where $\Tilde{\hat{\bm{G}}}^T = (\hat{\bm{G}}^T \hat{\bm{G}})^{-1}\hat{\bm{G}}^{T}$. Given estimated latent factor samples,
\begin{equation}
    \begin{aligned}
        \hat{\bm{f}} &= \lbrace \hat{\bm{f}}(1),\hat{\bm{f}}(2),...,\hat{\bm{f}}(n)\rbrace \\
        &= \lbrace \Tilde{\hat{\bm{G}}}^T\bm{X}(1),\Tilde{\hat{\bm{G}}}^T\bm{X}(2),...,\Tilde{\hat{\bm{G}}}^T\bm{X}(n) \rbrace.
    \end{aligned}
\end{equation}
with each sample $\bm{f}(t)$ as a $q$-dimensional vector,
\begin{equation}
    \begin{aligned}
        \hat{\bm{f}}(t) &= \lbrace \hat{f}_1(t),\hat{f}_2(t),...,\hat{f}_q(t)\rbrace\\
        &= \lbrace \bm{g}_1^T\bm{X}(t),\bm{g}_2^T\bm{X}(t),...,\bm{g}_q^T\bm{X}(t) \rbrace,
    \end{aligned}
\end{equation}
where $\Tilde{\hat{\bm{G}}} = \lbrace \bm{g}_1,\bm{g}_2,...,\bm{g}_q\rbrace$, we have,
\begin{equation}
\begin{aligned}
    \bm{\varepsilon}(t) &= (\bm{I}-\bm{B})\Tilde{\hat{\bm{G}}}^T\bm{X}(t), \\
    \varepsilon_i(t) &= \bm{g}_i^T\bm{X}(t)-\bm{b}_i^T\Tilde{\hat{\bm{G}}}^T\bm{X}(t).
\end{aligned}
\end{equation}
Hence, the log-likelihood of our model with regard to $\bm{B}$ and $\hat{\bm{G}}$ is derived as,
\begin{equation}\label{eq:likelihood}
    \begin{aligned}
        &\mathcal{L}(\bm{B},\hat{\bm{G}})\\
        &= \sum_{t=1}^n \bigg [ \frac{1}{2} \left \| \bm{X}(t)-\bm{A}\bm{\varepsilon}(t)\right \| ^2_{\Sigma^{-1}} + \sum_{i=1}^{q} \log {p_i}(\varepsilon_i(t)) \bigg ] \\
        &+ C, \\
        &= \sum_{t=1}^n \bigg [ \frac{1}{2} \left \| \bm{X}(t)-\hat{\bm{G}}\Tilde{\hat{\bm{G}}}^T\bm{X}(t)\right \| ^2_{\Sigma^{-1}} \\ &+ \sum_{i=1}^{q} \log {p_i}(\bm{g}_i^T\bm{X}(t)-\bm{b}_i^T\Tilde{\hat{\bm{G}}}^T\bm{X}(t)) \bigg ] + C, 
    \end{aligned}
\end{equation}
where $\bm{b}_i$ is the $i^{th}$ column of the weighted matrix $\bm{B}^{T}$. Note that, in~Eq.(\ref{eq:likelihood}), the first part shows the mechanism of generating $\bm{X}$ of the measurement model, while the second part characterizes the independence relationships between external influences $\bm{\varepsilon}$, which can be optimized systematically. 

The log-likelihood function enjoys the following merits: i) A dominating benefit of our formulation is the binding of likelihood with $\bm{\varepsilon}$. It enables us to company $\bm{\varepsilon}$ with the independence constraints, so that they can be directly dealt with as an optimization problem by manipulating $\bm{\varepsilon}$'s estimation~\cite{cai2018self}. ii) It belongs to a semi-parametric problem since the densities of $\bm{\varepsilon}$ are not specified. But, there are various available estimation methods for approximating $\log \hat{p}_i$~\cite{pham1997blind}. It has been realized that even when fixing $\hat{p}_i$ to be a single function, one can still receive a consistent and satisfactory estimator~\cite{hyvarinen2010estimation}. 

\subsubsection{E.2 The log-likelihood function for MD-LiNA models}

Given MD-LiNA models with $M>1$, we derive from~Eq.\eqref{eq:augment},
\begin{equation}
    \begin{aligned}
        \bar{\bm{x}} = \bar{\bm{A}} \bar{\bm{\varepsilon}} + \bar{\bm{e}},
    \end{aligned}
\end{equation}
where $\bar{\bm{A}}=\bar{\bm{G}}(\bm{I}-\bar{\bm{B}})^{-1}$. Considering the fact that this model aims to estimate $\bar{\bm{G}}$ and $\tilde{\bm{B}}$ between $\tilde{\bm{f}}$, rather than $\bar{\bm{B}}$, and based on the joint distribution $\bm{P} ({\bm{X}}, {\bm{\varepsilon}})$ for LiNA models in~Eq.\eqref{eq:joint_dis}, we define a newly $\bm{P} (\bar{\bm{X}}, \bar{\bm{\varepsilon}})$ for MD-LiNA models, 
\begin{equation} 
    \begin{aligned}
        & \bm{P} (\bar{\bm{X}}, \bar{\bm{\varepsilon}}) \triangleq  \bm{P} (\bar{\bm{X}}|\bar{\bm{\varepsilon}})\bm{P} (\tilde{\bm{\varepsilon}}),\\
        &= \mathcal{G}(\bar{\bm{X}}-\bar{\bm{A}} \bar{\bm{\varepsilon}},\Sigma)\prod_{i=1}^{\tilde{q}} p_i({\tilde{\varepsilon}}_i).
    \end{aligned}
\end{equation}
where $\tilde{\bm{\varepsilon}}$ are random vectors that collect external influences of $\tilde{\bm{f}}$ and $\Sigma$ is the covariance matrix of $\bar{\bm{e}}$. Taking the logarithm yields,
\begin{equation} \label{eq:log-p}
    \begin{aligned}
        &\log (\bm{P} (\bar{\bm{X}}, \bar{\bm{\varepsilon}}))\\
        =& \sum_{t=1}^n \bigg [ \log \mathcal{G}(\bar{\bm{X}}(t)-\bar{\bm{A}}\bar{\bm{\varepsilon}}(t),\Sigma) 
        + \sum_{i=1}^{\tilde{q}} \log {p_i}(\tilde{\varepsilon}_i(t)) \bigg ],\\
        = & \sum_{t=1}^n \bigg [ \frac{1}{2} \left \| \bar{\bm{X}}(t)-\bar{\bm{A}}\bar{\bm{\varepsilon}}(t)\right \| ^2_{\Sigma^{-1}} 
        + \sum_{i=1}^{\tilde{q}} \log {p_i}(\tilde{\varepsilon}_i(t)) \bigg ] + C, \\
    \end{aligned}
\end{equation}
where $\left \| \bm{x} \right \|^2_{\Sigma^{-1}} = \bm{x}^T \Sigma^{-1}\bm{x} $, $\bar{\bm{X}}(t)$ and $\bar{\bm{\varepsilon}}(t)$ are realizations of the data and the external influences in $\bar{\bm{X}}$, i.e., the $t^{th}$ column of $\bar{\bm{X}}$ and $\bar{\bm{\varepsilon}}$, respectively.

Similarly with the derivation of the log-likelihood of LiNA models, we have from~Eqs.\eqref{eq:MD_LiNA} and~\eqref{eq:tilde1},
\begin{equation} \label{eq:epsilon}
\begin{aligned} 
    \bar{\bm{\varepsilon}}(t) &= (\bm{I}-\bar{\bm{B}})\Tilde{\bar{\bm{G}}}^{T}\bar{\bm{X}}(t), \\
    \tilde{\bm{\varepsilon}}(t) &= (\bm{I}-\tilde{\bm{B}})\Tilde{\hat{\bm{H}}}^T\bar{\bm{f}}(t), \\
    \tilde{\varepsilon}_i(t) &= \bm{h}_i^T\bar{\bm{f}}(t)-\tilde{\bm{b}}_i^T\Tilde{\hat{\bm{H}}}^T\bar{\bm{f}}(t),
\end{aligned}
\end{equation}
where $\Tilde{\bar{\bm{G}}}^{T} = (\bar{\bm{G}}^{T} \bar{\bm{G}})^{-1}\bar{\bm{G}}^{T}$, $\Tilde{\hat{\bm{H}}}^T = ({\bm{H}}^T {\bm{H}})^{-1}{\bm{H}}^{T}$, and $\bm{h}_i$ is the $i^{th}$ column of $\Tilde{\hat{\bm{H}}}$.  $\tilde{\bm{b}}_i$ denotes the $i^{th}$ column of $\tilde{\bm{B}}^{T}$.

Hence, combining~Eqs.\eqref{eq:log-p} and~Eq.\eqref{eq:epsilon}, we get the log-likelihood function for MD-LiNA models,
\begin{equation} 
\begin{aligned}
&\mathcal{F}(\tilde{\bm{B}},{\bm{H}}) = -\mathcal{L} (\tilde{\bm{B}},{\bm{H}}) + \lambda_1 \Arrowvert \tilde{\bm{B}} \Arrowvert_{1*} + \lambda_2 \Arrowvert \tilde{\bm{B}} \Arrowvert^2,\\
&=- \sum_{t=1}^n \bigg [ \frac{1}{2} \left \| \bar {\bm{X}}(t)-\bar{\bm{G}}\Tilde{\bar{\bm{G}}}^{T}\bar {\bm{X}}(t)\right \| ^2_{\Sigma^{-1}} \\ 
&+ \sum_{i=1}^{\tilde{q}} \log \hat{p}_i(\bm{h}_i^T\bar{\bm{f}}(t)-\tilde{\bm{b}}_i^T\Tilde{\hat{\bm{H}}}^T\bar{\bm{f}}(t)) \bigg ] - C \\
&+ \lambda_1 \Arrowvert \tilde{\bm{B}} \Arrowvert_{1*} + \lambda_2 \Arrowvert \tilde{\bm{B}} \Arrowvert^2,
\end{aligned}
\end{equation}
Note that compared with the log-likelihood for LiNA models in~Eq.\eqref{eq:likelihood}, we replace $\bm{B}$ with $\tilde{\bm{B}}$, since $\tilde{\bm{B}}$ is our basic focus. Subsequently, part of our optimization becomes, 
\begin{equation} 
\begin{aligned}
&\mathcal{F}(\tilde{\bm{B}},{\bm{H}}) = -\mathcal{L} (\tilde{\bm{B}},{\bm{H}}) + \lambda_1 \Arrowvert \tilde{\bm{B}} \Arrowvert_{1*} + \lambda_2 \Arrowvert \tilde{\bm{B}} \Arrowvert^2,\\
&=- \sum_{t=1}^n \bigg [ \frac{1}{2} \left \| \bar {\bm{X}}(t)-\bar{\bm{G}}\Tilde{\bar{\bm{G}}}^{T}\bar {\bm{X}}(t)\right \| ^2_{\Sigma^{-1}} \\ 
&+ \sum_{i=1}^{\tilde{q}} \log \hat{p}_i(\bm{h}_i^T\bar{\bm{f}}(t)-\tilde{\bm{b}}_i^T\Tilde{\hat{\bm{H}}}^T\bar{\bm{f}}(t)) \bigg ] - C \\
&+ \lambda_1 \Arrowvert \tilde{\bm{B}} \Arrowvert_{1*} + \lambda_2 \Arrowvert \tilde{\bm{B}} \Arrowvert^2,
\end{aligned}
\end{equation}

\subsection*{F: Complexity Analysis}
We analyze the complexity of Algorithm~\ref{alg:Framwork_b}. In phase I, we estimate the factor loading matrix $\bar{\bm{G}}$ mainly by CFA with BFGS method, whose computational complexity is $O(pqt_1)$~\cite{nocedal2006numerical}; and estimate then the augmented latent factors $\bar{\bm{f}}$ with complexity $O(pqn)$, where $p$ and $q$ are the number of all observed variables in $\bm{X}$ and latent factors in $\bar{\bm{f}}$, respectively. $t_1$ is the number of iterations in CFA and $n$ is the sample sizes of all domains. 

In phase II, since we estimate the transformation matrix $\bm{H}$ and the causal effects matrix $\tilde{\bm{B}}$ iteratively for MD-LiNA, the computational complexities for $\bm{H}$ and $\tilde{\bm{B}}$ at each iteration are $O(q\tilde{q}t_2)$ and $O(m^2|S|t_3 + m^3t_3 + m|S|t_3t_4)$, respectively~\cite{nocedal2006numerical,zheng2018dags}. $\tilde{q}$ is the number of latent factors of interests $\tilde{\bm{f}}$, $m$ is the memory size of L-BFGS ($m \ll \tilde{q}^2$), and $S$ is the active set of $\tilde{\bm{B}}$ ($|S| < \tilde{q}^2$). $t_2$ is the number of inner iterations for $\bm{H}$ while $t_3$ and $t_4$ are those for $\tilde{\bm{B}}$. In particular, $t_4$ is the number of inner iterations of L-BFGS update when estimating $\tilde{\bm{B}}$~\cite{zheng2018dags}.

\subsection*{G: More Details about Synthetic Experiments}
\subsubsection{G.1 Data generation and experimental settings}
In simulation, the data were generated according to~Eq.(\ref{eq:MD_LiNA}), taking a single-domain LiNA model as an example ($M=1$), i.e., 
\begin{itemize}
  \item [1)] 
  Firstly, we randomly generated the latent factors $\bm{f}$ from the first equation in~Eq.(\ref{eq:MD_LiNA}) as in~\citeauthor{zheng2018dags}~\shortcite{zheng2018dags}. Specifically, we simulated a random unweighted DAG, according to the graph models, namely Er{\"d}os–{\'R}enyi (ER) model~\cite{newman2018networks}. Given the DAG, we subsequently assigned uniformly the edge weights from $[-2,-0.5] \cup [0.5,2]$ to get an adjacency matrix $\bm{B}$. Thereafter, we independently drew samples for the external influence variables $\bm{\varepsilon}$, which followed non-Gaussian distributions having zero means and unit variances. Having obtained $\bm{\varepsilon}$ and $\bm{B}$, we generated factors from the first equation in~Eq.(\ref{eq:MD_LiNA}) and normalized them to be of unit variances.    
  \item [2)]
  Secondly, we generated the measurement variables $\bm{x}$ from the second equation in~Eq.(\ref{eq:MD_LiNA}) as in~\citeauthor{shimizu2009estimation}~\shortcite{shimizu2009estimation}. Particularly, we independently sampled the measurement errors $\bm{e}$ from Gaussian distributions with zero means and unit variances, and we sampled the loading matrix $\bm{G}$ uniformly at random from an interval $[-0.7,-0.2] \cup [0.2,0.7]$. In order to constrain the variances of variables $\bm{X}$ to units when measurement errors were added, we normalized the rows of matrix $\bm{G}$ and errors $\bm{e}$. Finally, we employed the second equation in~Eq.(\ref{eq:MD_LiNA}) to create the observed data $\bm{X}$. 
\end{itemize}

Unless specified, the following detailed settings are used in each experiment. Each generated structure has 5 latent factors and each factor measures 2 pure variables in one domain. The sample size is fixed as 1000. In addition, we sampled the noises $\bm{\varepsilon}$ from Laplace distributions, setting the noise ratio to be 0.1. We ran our methods with regularization parameters $\lambda_1=\lambda_2=0.1$ and an effect threshold $\epsilon=0.3$. Especially, for the iv) multi-domain data experiment, we set different thresholds $\epsilon$ for different numbers of latent factors $q_m$, i.e., we set $\epsilon=0.3$ when $q_m=5$; $\epsilon=0.1$ when $q_m=3$; and $\epsilon=0.05$ when $q_m=2$. For each experimental setting, we conducted 100 independent trials.

We compared our method with NICA~\cite{shimizu2009estimation} and Triad~\cite{cai2019triad}. Since Triad outputs causal directions of latent factors with no causal effects, we used the recall, precision, and F1 score as measurements to evaluate the difference between the estimated skeleton and the true one. F1 score is a weighted average of recall and precision, with F1 score = 2$\times$recall$\times$precision/(recall+precision). Besides, since NICA aimed to estimate both causal directions and causal effects between latent factors but not simultaneously, we exploited scatter plots between the estimated matrices and the true ones for another accuracy measurement.

\subsubsection{G.2 Additional experiments}
To account for the sensitivity of noise ratios, regularization parameters and non-Gaussian distributions, we carried out additional experiments with \textbf{v) different noise ratios}, \textbf{vi) different regularization parameters} as well as \textbf{vii) different non-Gaussian distributions} of the noises of latent factors.

\begin{figure}[t!]
\centering
\subfigure[Re. with no.] {
\includegraphics[width=0.145\textwidth]{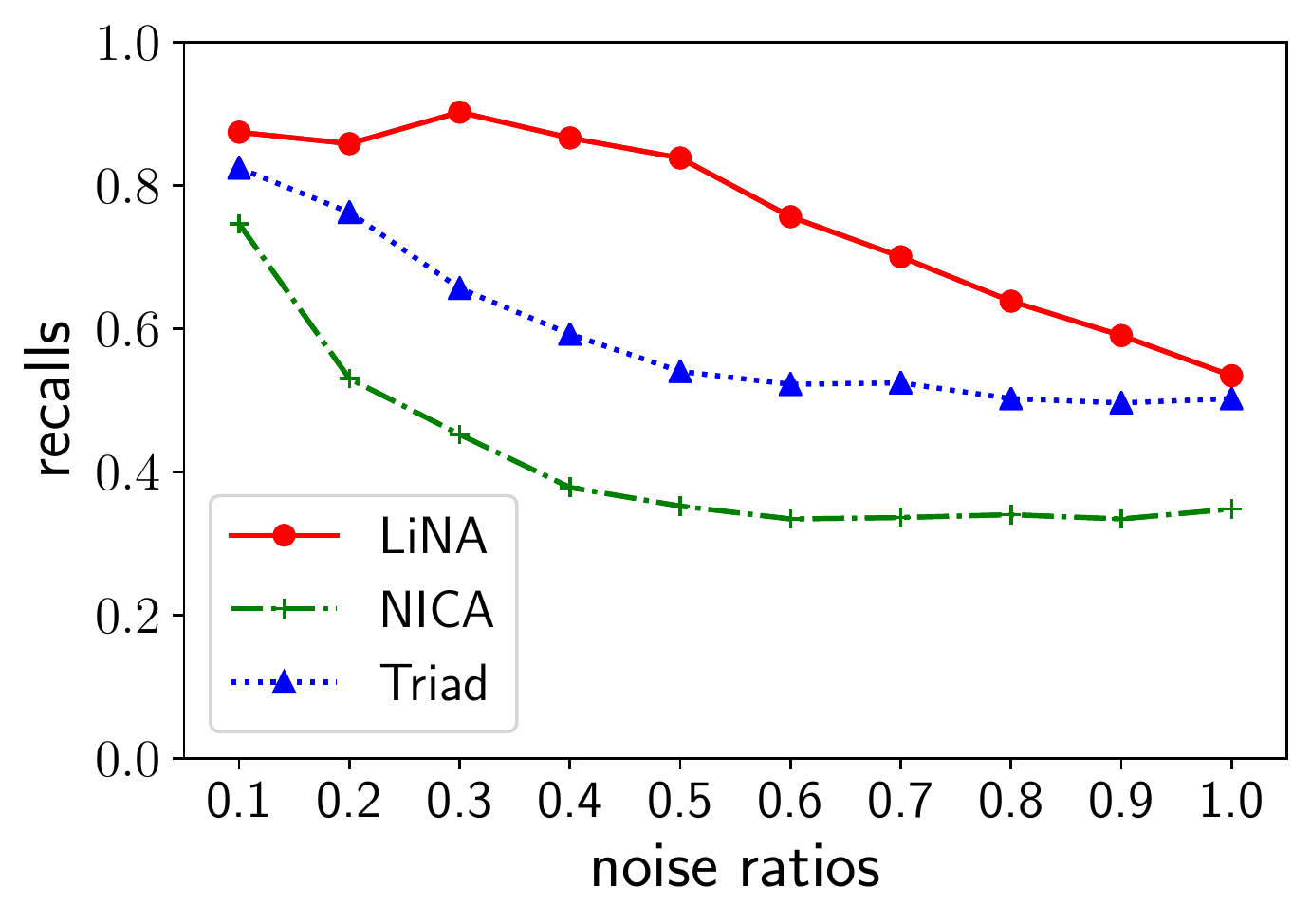}}
\subfigure[Pre. with no.]{
\includegraphics[width=0.145\textwidth]{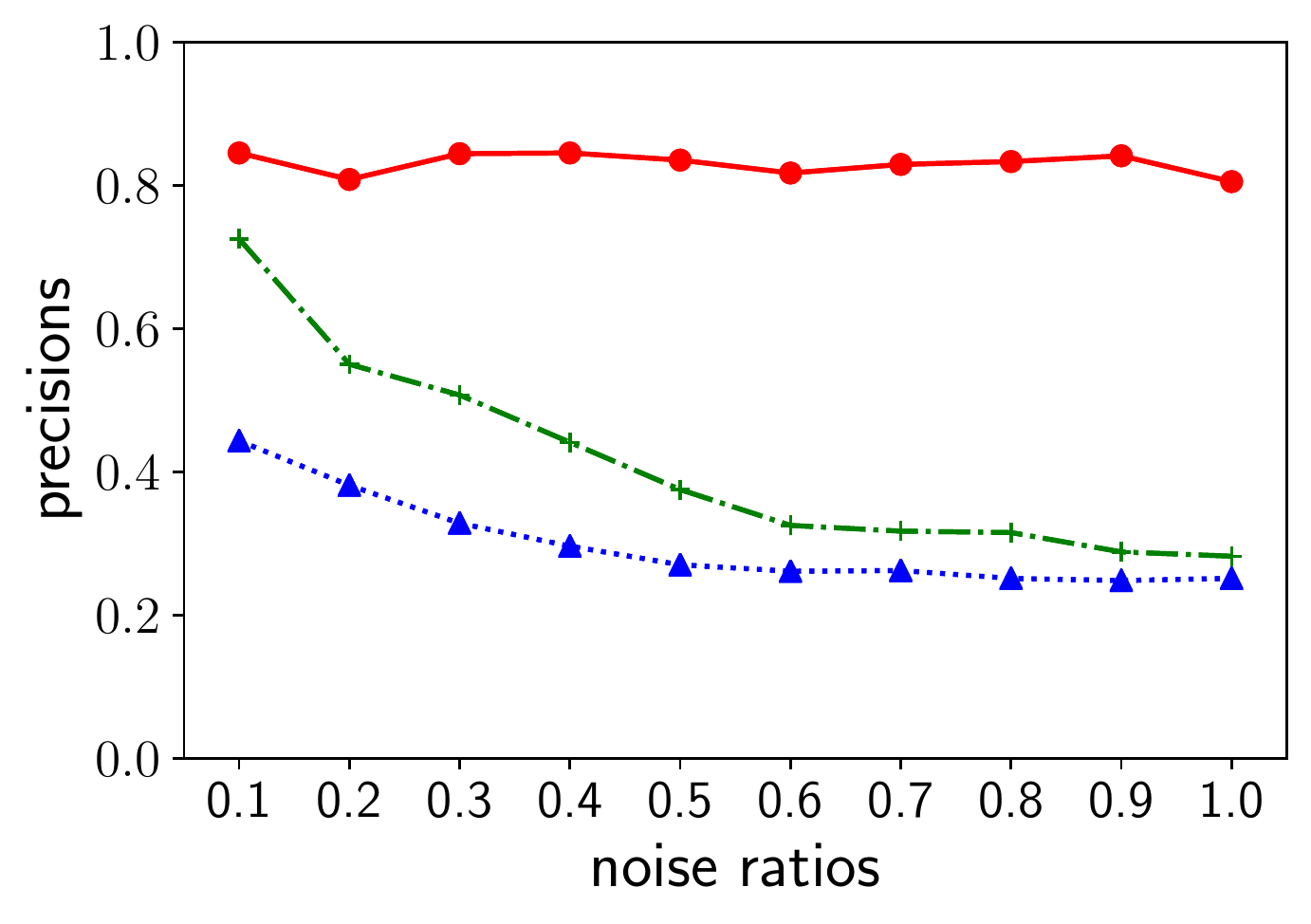}}
\subfigure[F1. with no.]{
\includegraphics[width=0.145\textwidth]{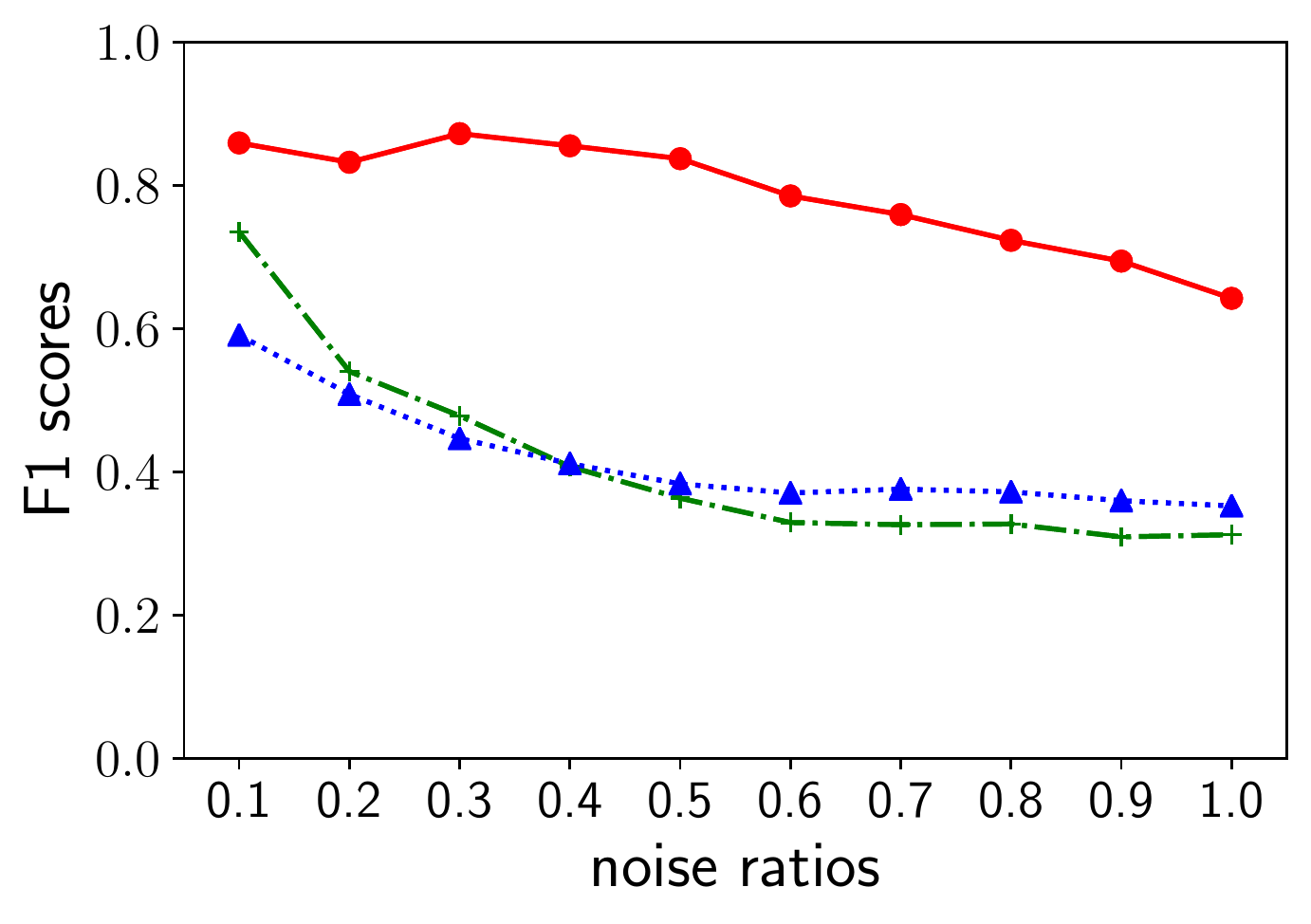}}
\caption{The sensitivity performance of different noise ratios. The x-axes are noise ratios while y-axes are recalls (Re.), precisions (Pre.), and F1 scores (F1.), respectively.}\label{fig:diff_noise}
\end{figure}

We demonstrated in Figure~\ref{fig:diff_noise} the sensitivity of \textbf{v) different noise ratios}, which are $\lbrace 0.1,...,1.0\rbrace$. As shown, we observed that all methods exhibit distinct slopes of decreases as the noise ratios increase, among which our LiNA's performance is less sensitive. These decreases are all reasonable. For LiNA and NICA methods, such decreases are due to the approximation errors, which have been declared by~\citeauthor{hyvarinen2001independent}~\shortcite{hyvarinen2001independent} that approximations are reliably credible when noise ratios are small enough. For Triad method, this decrease is owing to the fact that Triad highly relies on the independence tests. When noise ratios increase, the non-Gaussianity of measurement variables decreases resulting in the independence tests' unreliability.

\begin{figure}[t!]
\centering
\subfigure[F1. with $\lambda_1$]{
\includegraphics[width=0.145\textwidth]{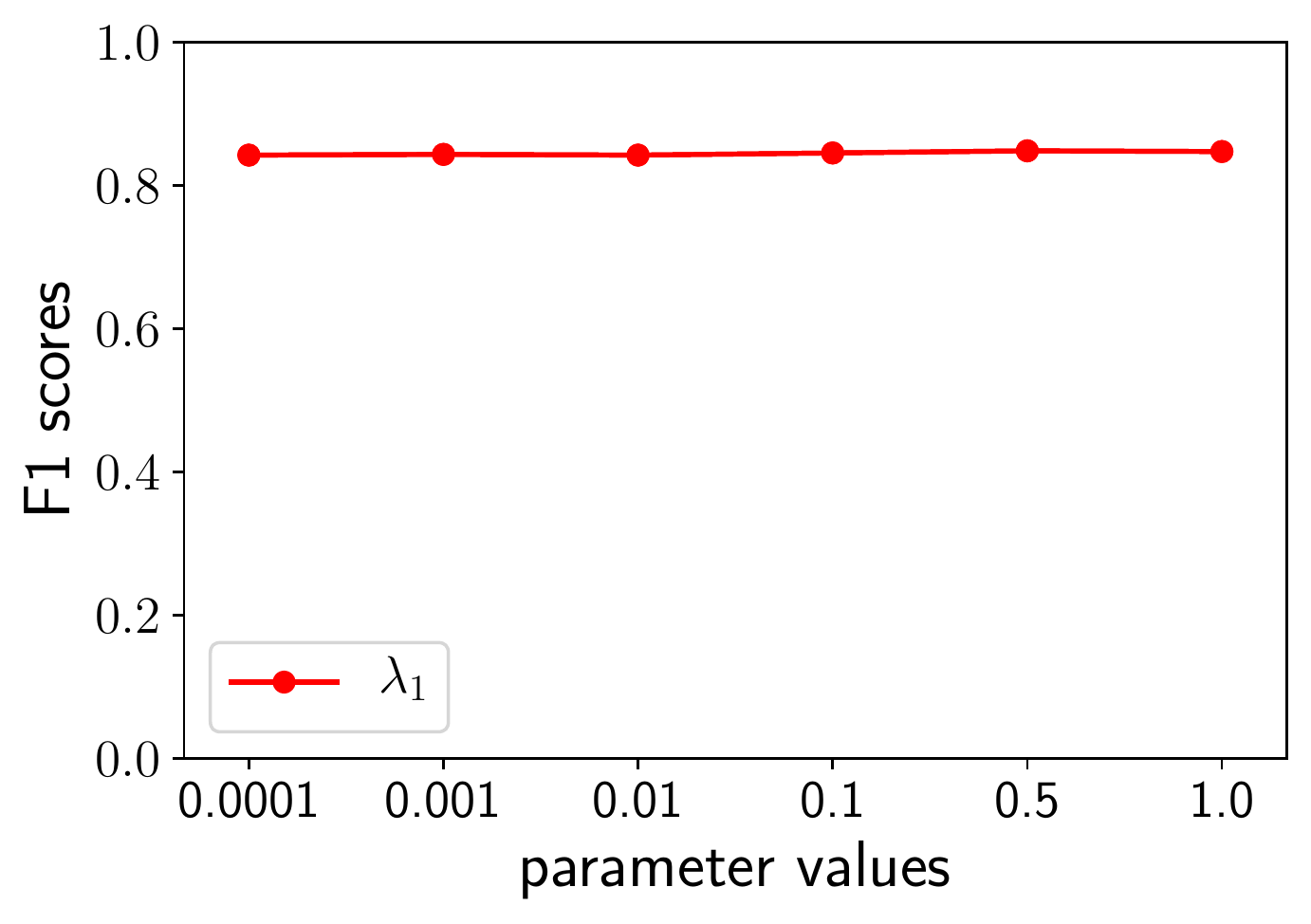}}
\subfigure[F1. with $\lambda_2$]{
\includegraphics[width=0.145\textwidth]{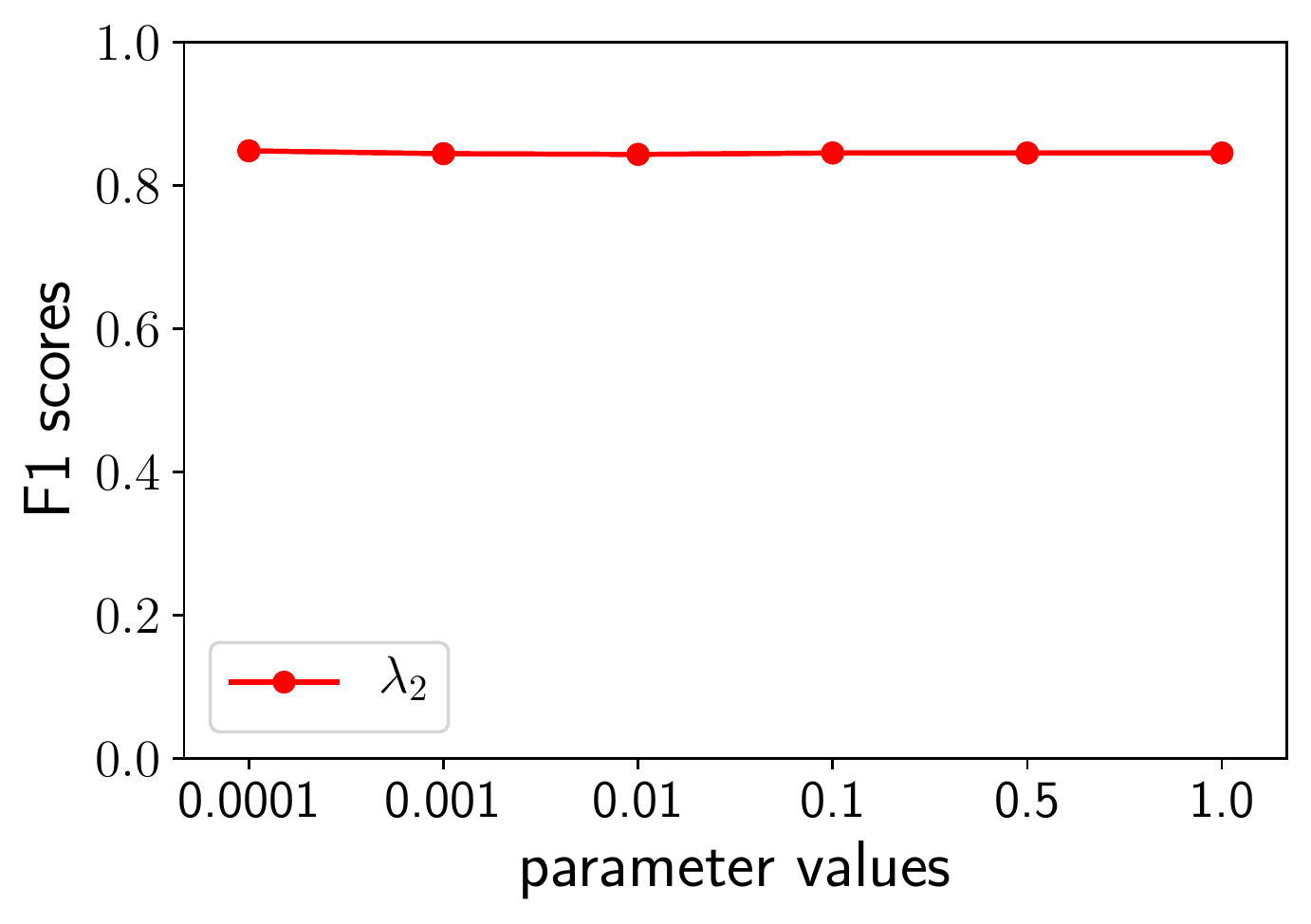}}
\subfigure[F1. with $\epsilon$]{
\includegraphics[width=0.145\textwidth]{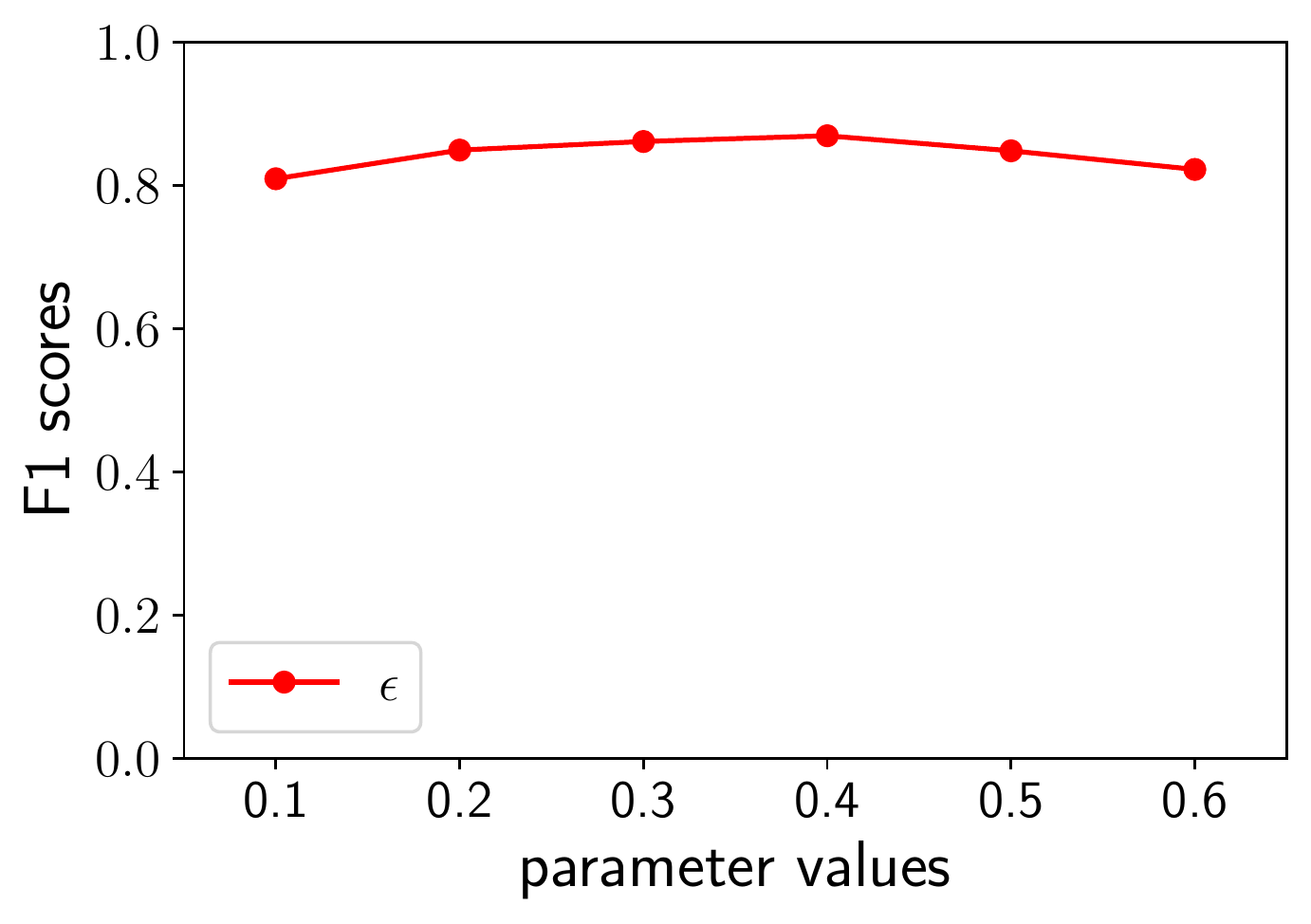}}
\caption{The sensitivity performance of parameters $\lambda_1, \lambda_2$ and $\epsilon$. 
The x-axes are parameter values while y-axes are F1 scores (F1.).}\label{fig:diff_para}
\end{figure}
We examined the sensitivity of \textbf{vi) different regularization parameters}, i.e., $\lambda_1, \lambda_2$ and $\epsilon$. We firstly fix $\lambda_2=0.1$ and $\epsilon=0.3$, and set $\lambda_1$ to be different values, i.e., 0.0001, 0.001, 0.01, 0.1, 0.5 and 1.0, as shown in the (a) of Figure~\ref{fig:diff_para}. As in (b), we fix $\lambda_1=0.1$ and $\epsilon=0.3$ in turn, and set $\lambda_2$ to be different values. We see that F1 scores of both $\lambda_1$ and $\lambda_2$ are stable with varying values of parameters. To investigate the effects of $\epsilon$, we set $\lambda_1=\lambda_2=0.1$, and vary $\epsilon=\lbrace 0.1,...,0.6\rbrace$ with results demonstrated in (c). Overall, we observe that performances will slightly drop when using smaller or larger $\epsilon$. It is expectedly reasonable, as smaller $\epsilon$ may leave more spurious edges and larger $\epsilon$ may cut down true edges. Thus, an optimal choice that balances the F1 scores and real causal effects in this case, may be 0.3, since it is small moderately and gains satisfactory F1 scores as well.

\begin{figure}[t!]
\centering
\subfigure[Re. with Sub.]{
\includegraphics[width=0.145\textwidth]{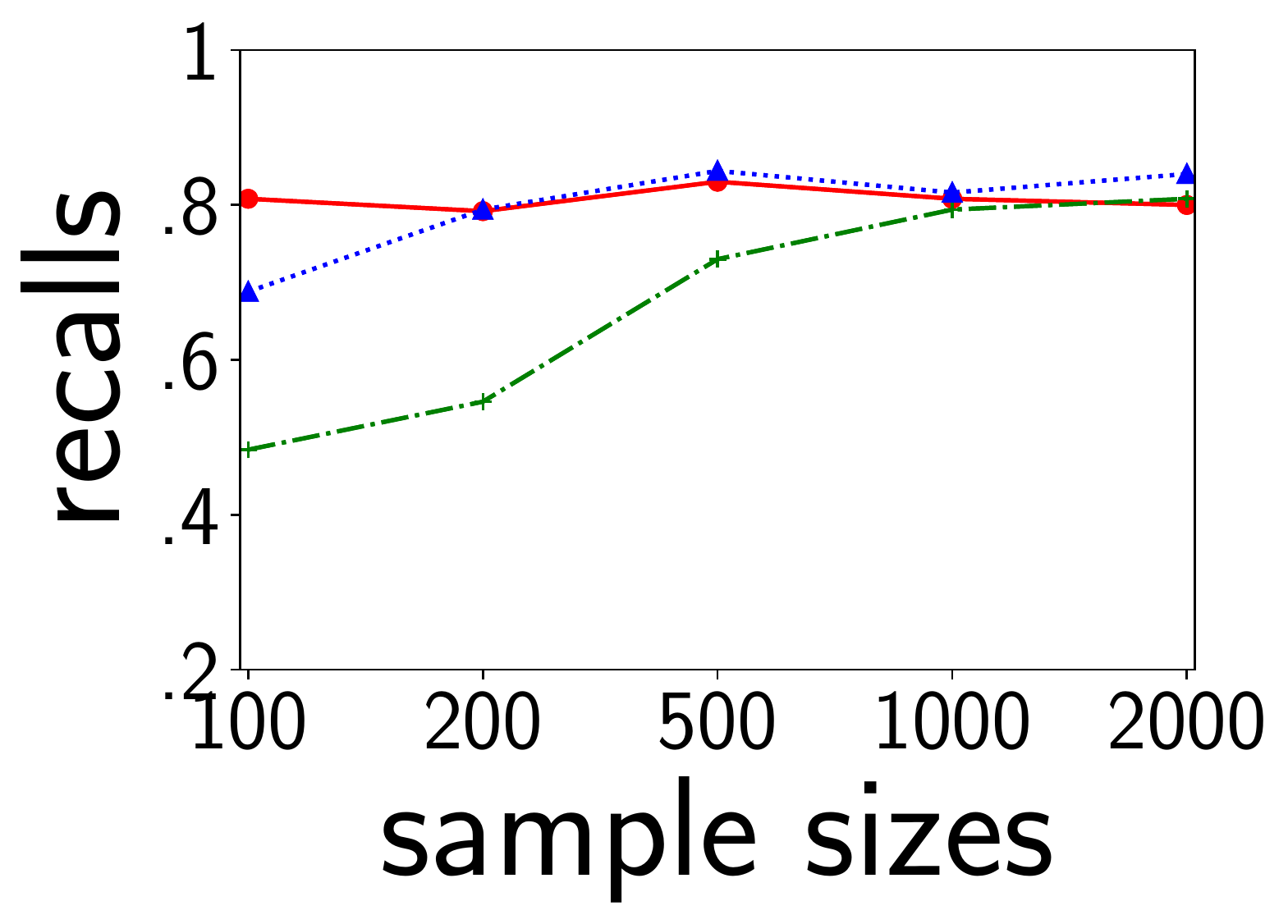}}
\subfigure[Pre. with Sub.]{
\includegraphics[width=0.145\textwidth]{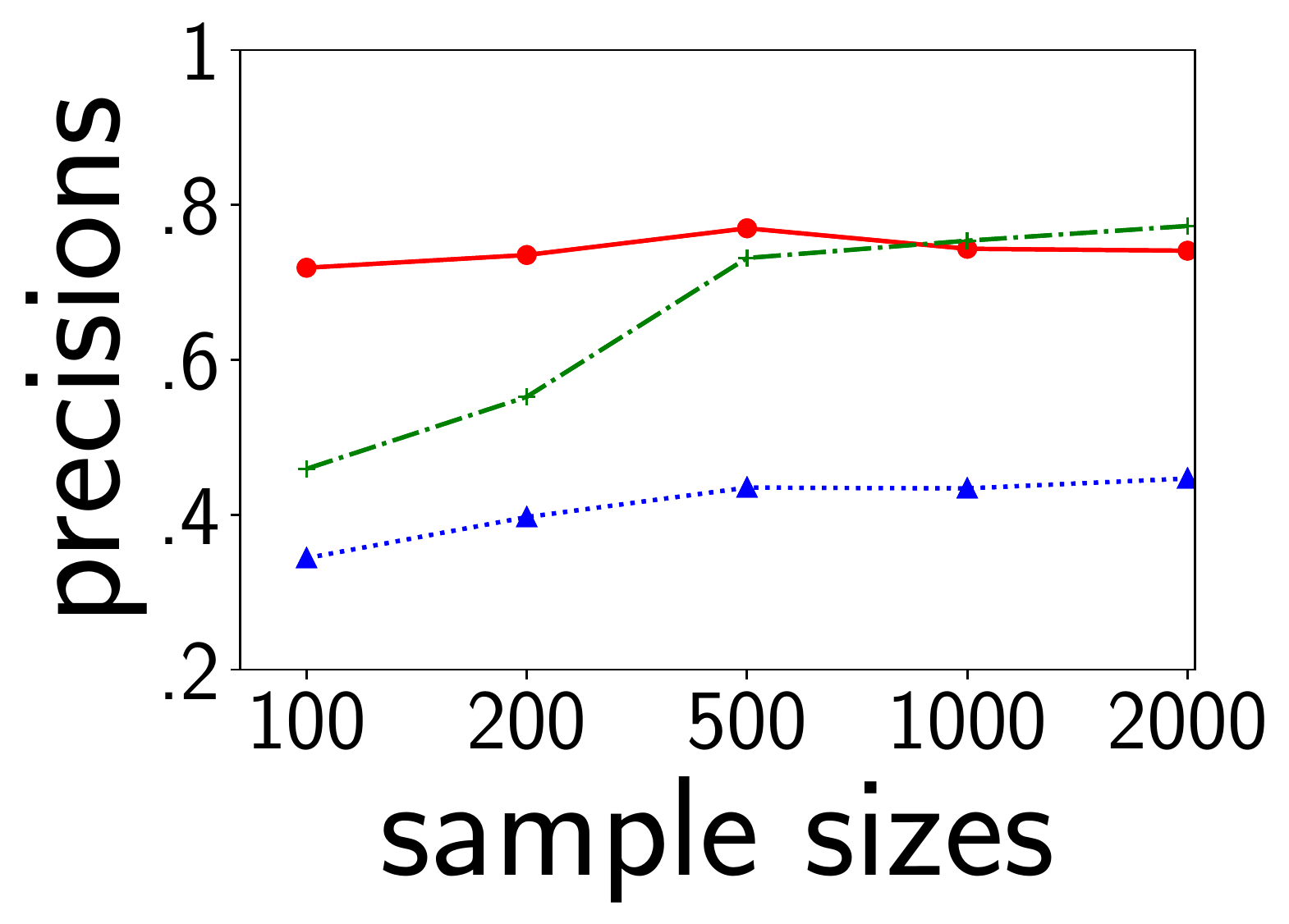}}
\subfigure[F1. with Sub.]{
\includegraphics[width=0.145\textwidth]{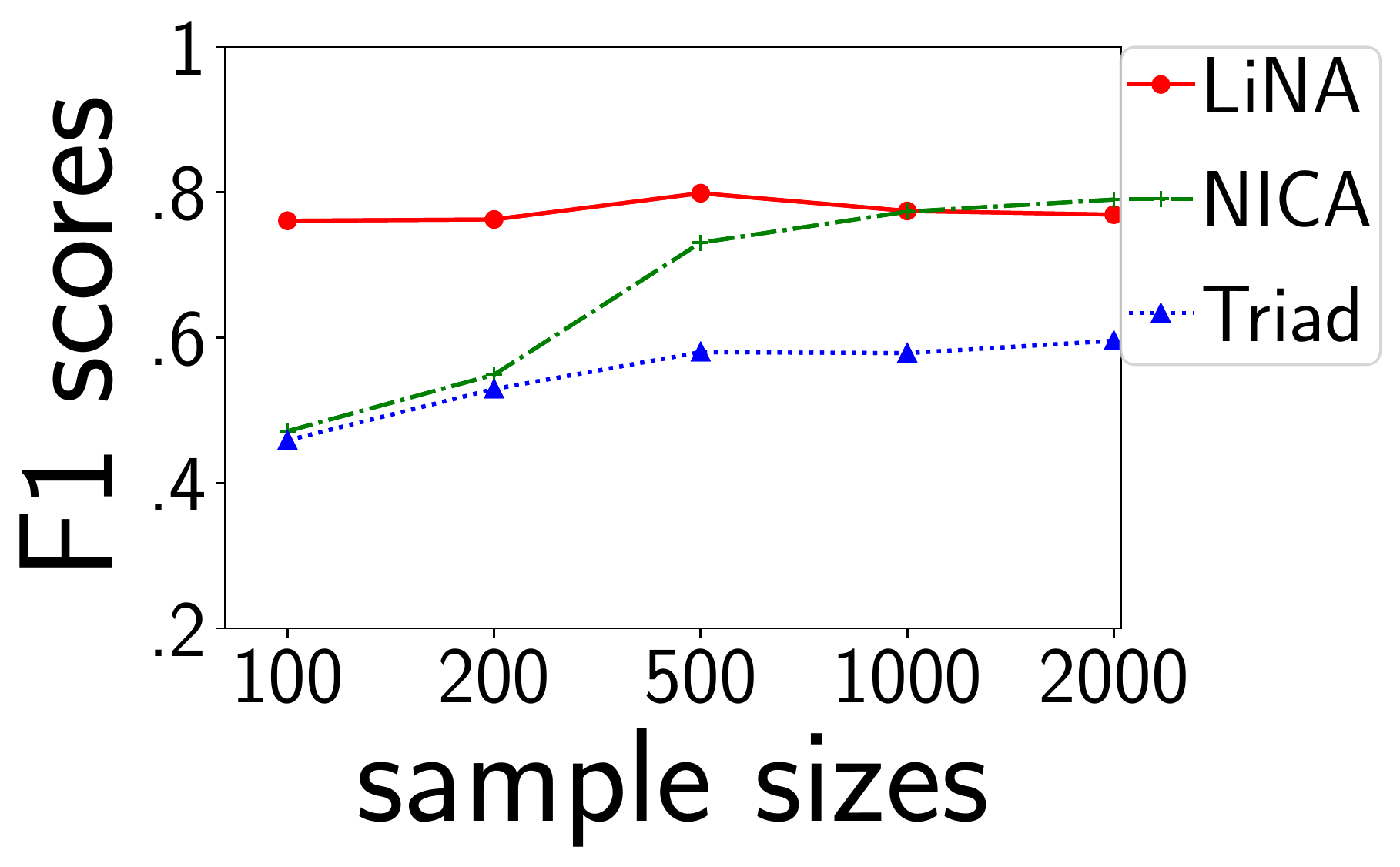}}
\subfigure[Re. with Sup.]{
\includegraphics[width=0.145\textwidth]{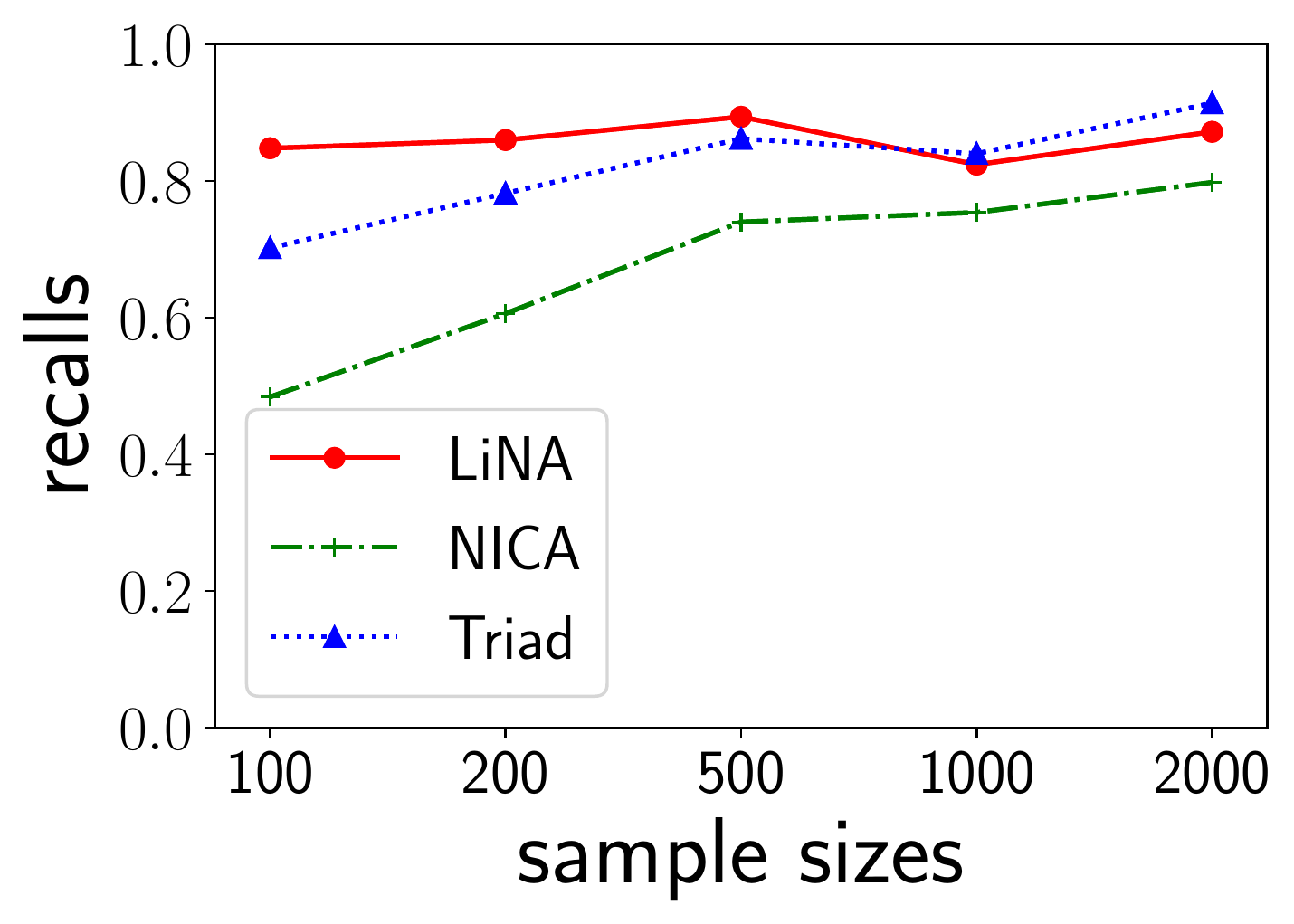}}
\subfigure[Pre. with Sup.]{
\includegraphics[width=0.145\textwidth]{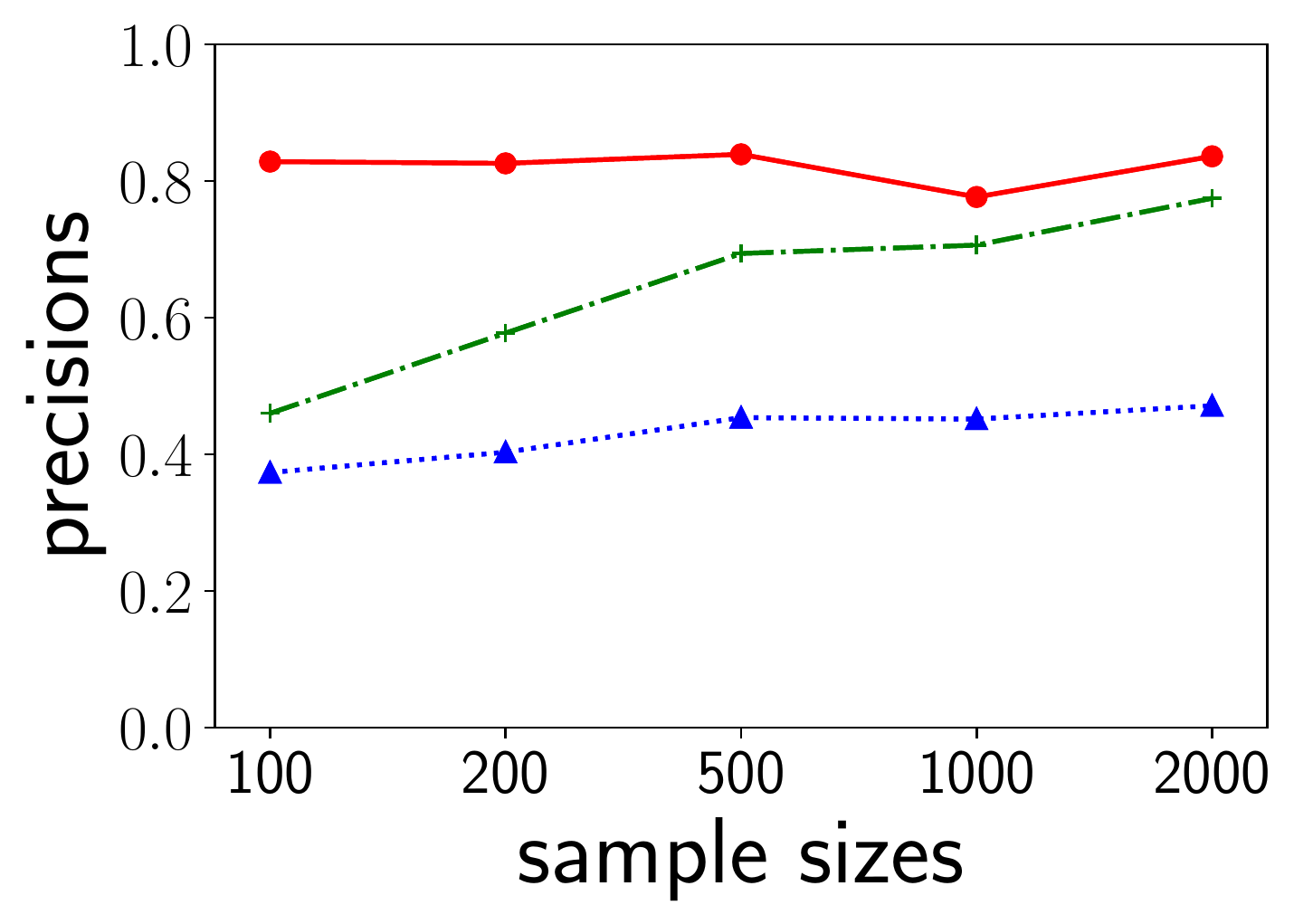}}
\subfigure[F1. with Sup.]{
\includegraphics[width=0.145\textwidth]{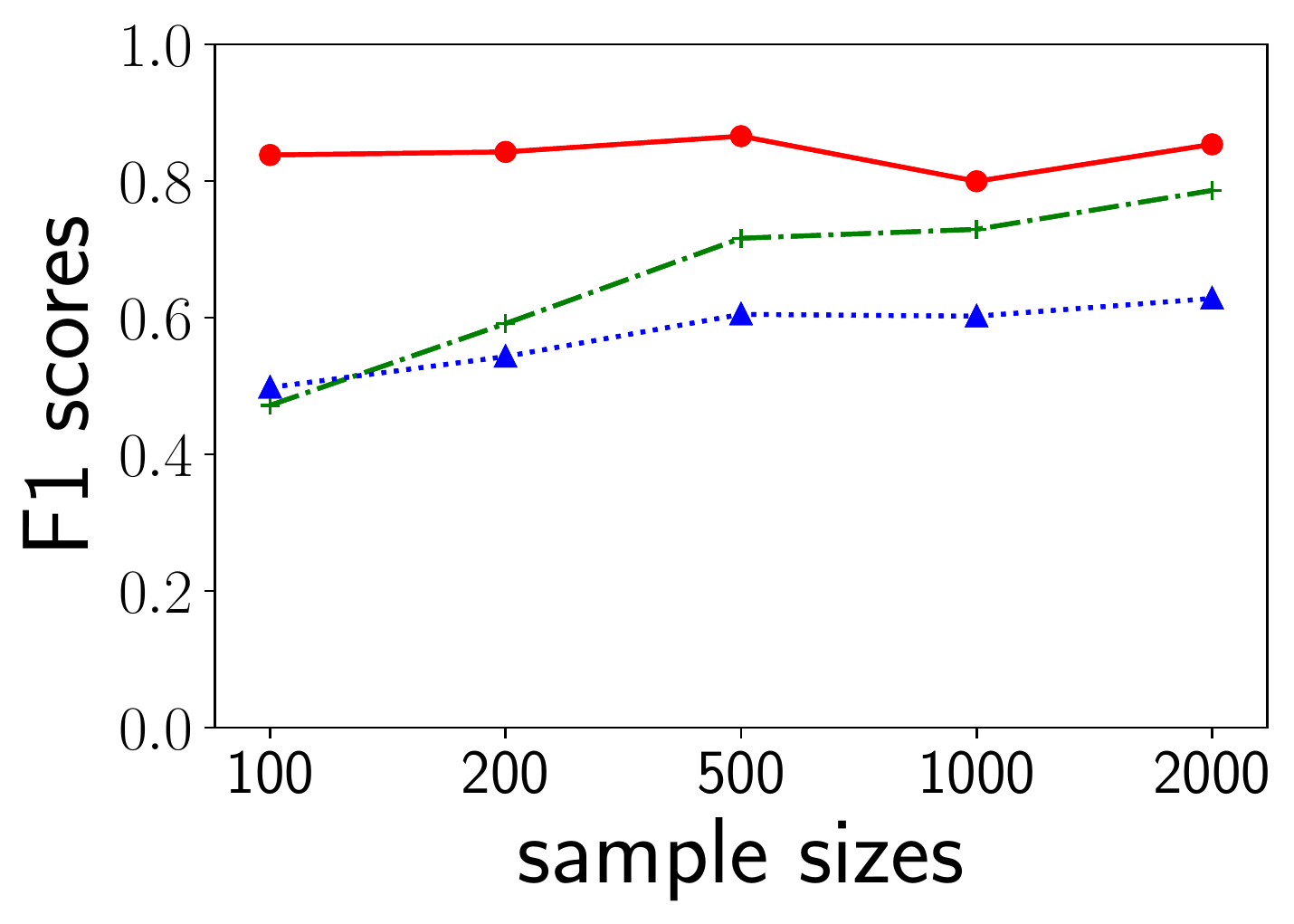}}
\caption{The recall (Re.), precision (Pre.) and F1 scores (F1.) of the recovered causal graphs between latent factors with different non-Gaussian distributions of the noises of latent factors. In (a), (b) and (c), noises follow sub-Gaussian distributions (Sub.) while those in (d), (e) and (f) follow super-Gaussian distributions (Sup.). The x-axis is the sample size and the y-axes are Re., Pre., and F1., respectively. Higher F1 score represents higher accuracy.}
\label{fig:diff_dis}
\end{figure}

In the main paper, we fixed distributions of noises latent factors to be Laplace ones. Thus, here we examined the sensitivity of 
\textbf{vii) 
different non-Gaussian distributions}, i.e., sub-Gaussian and super-Gaussian\footnote{In addition to the Laplace distributions, we generally generated both sub-Gaussian and super-Gaussian distributions in simulations through the following way: Firstly, a Gaussian variable $z$ with zero mean and unit variance is generated and transformed to a non-Gaussian variable by 
$\varepsilon_i=\sign (z) \left| z \right|^q$. In particular, we control the nonlinear exponent $q$ to lie in $\left[ 0.5,0.8\right]$ so that $\varepsilon_i$ is sub-Gaussian; while $\left[ 1.2,2.0\right]$ so that $\varepsilon_i$ is super-Gaussian. Eventually, the transformed $\varepsilon_i$ is standardized to have zero mean and unit variance.}. 
Overall, we found that our LiNA method achieves the best performance in most cases. In detail, when the noises are sub-Gaussian and there are more than 1000 samples (Figure~\ref{fig:diff_dis} (c)), NICA is comparable to ours. However, as sample sizes decrease, performances of the compared methods decrease, especially for NICA method who decreases dramatically for both distributions, whereas our LiNA remains incredibly stable. It verifies our robustness to different non-Gaussian distributions.

\subsection*{H: More Details about Real-World Experiments}

\subsubsection{H.1 Yahoo stock indices dataset}
The daily stock data were downloaded from the Yahoo finance database at 1078 days (from February 10, 2015, to January 10, 2020), and we used the adjusted closing prices for the stocks. We normalized each stock so that they have zero means and unit variances. To verify the effectiveness of our proposed method in multiple-domain schemes, we divided the data into two non-overlapping time segments such that their distributions varied across segments and they can be viewed as two different domains. The first domain had 631 observations (from February 10, 2015, to December 29, 2017) while the other 447 (from January 3, 2018, to January 10, 2020). Firstly, we found that the VIFs for each region are all lower than 10, which implies that such data do not encounter the multicollinearity problem. Hence, we employed the sparsity constraint and acyclicity constraint, without the $\ell _2$ regularization.  We applied $\lambda_1 = 0.01$ and $\epsilon = 0.1$, which were automatically selected via grid search with details in the following.

\begin{figure}[h!]
\centering
\includegraphics[width=0.45\textwidth]{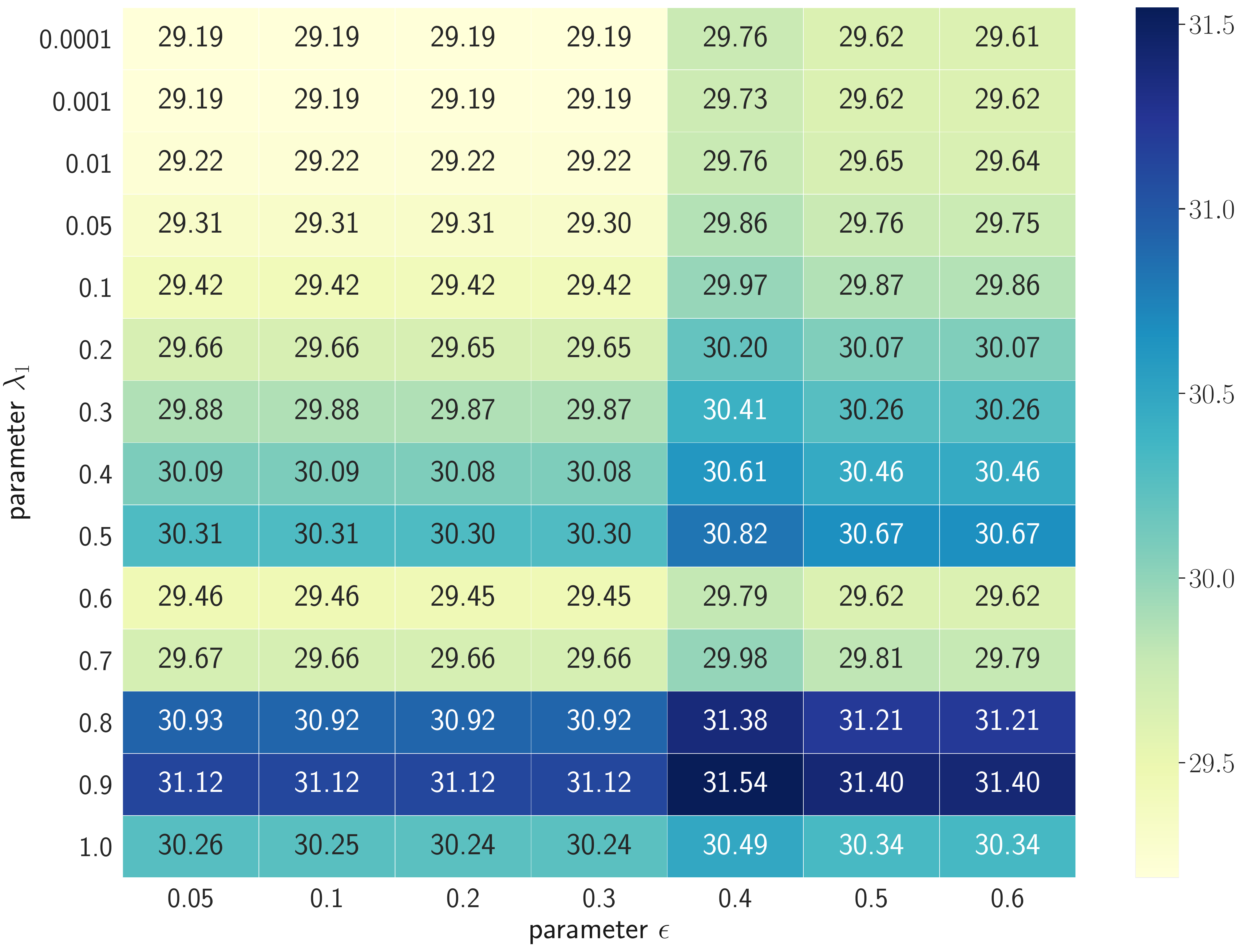}
\caption{The Heatmap of cross validation negative likelihoods for the Yahoo stock indices data over a range of parameter values $\lambda_1$ and $\epsilon$.}\label{fig:para_stock}
\end{figure}

We used 10-fold cross validation to select the regularization parameters $\lambda_1$ and $\epsilon$ for the Yahoo stock indices dataset. In particular, we evaluated in the validation our performance using the negative likelihood function in~Eq.\eqref{eq:unconstrained} over a range of parameter values. The result is reported in Figure~\ref{fig:para_stock}. We found that as $\epsilon$ increases or $\lambda_1$ increases, negatively likelihoods tend to increase as well (except when $\lambda_1 = 0.6$ or $0.7$), implying declining performances. We select $\lambda_1=0.001$ and $\epsilon=0.1$, since they correspond to one of the smallest values in the grid.
Due to the different time zones, it is expected the ground truth is $\mathrm{Asia} \to \mathrm{Europe} \to \mathrm{USA}$~\cite{janzing2010telling,chen2014causal}

\subsubsection{H.2 fMRI hippocampus dataset}
\begin{figure}[t]
\centering
\includegraphics[width=0.45\textwidth]{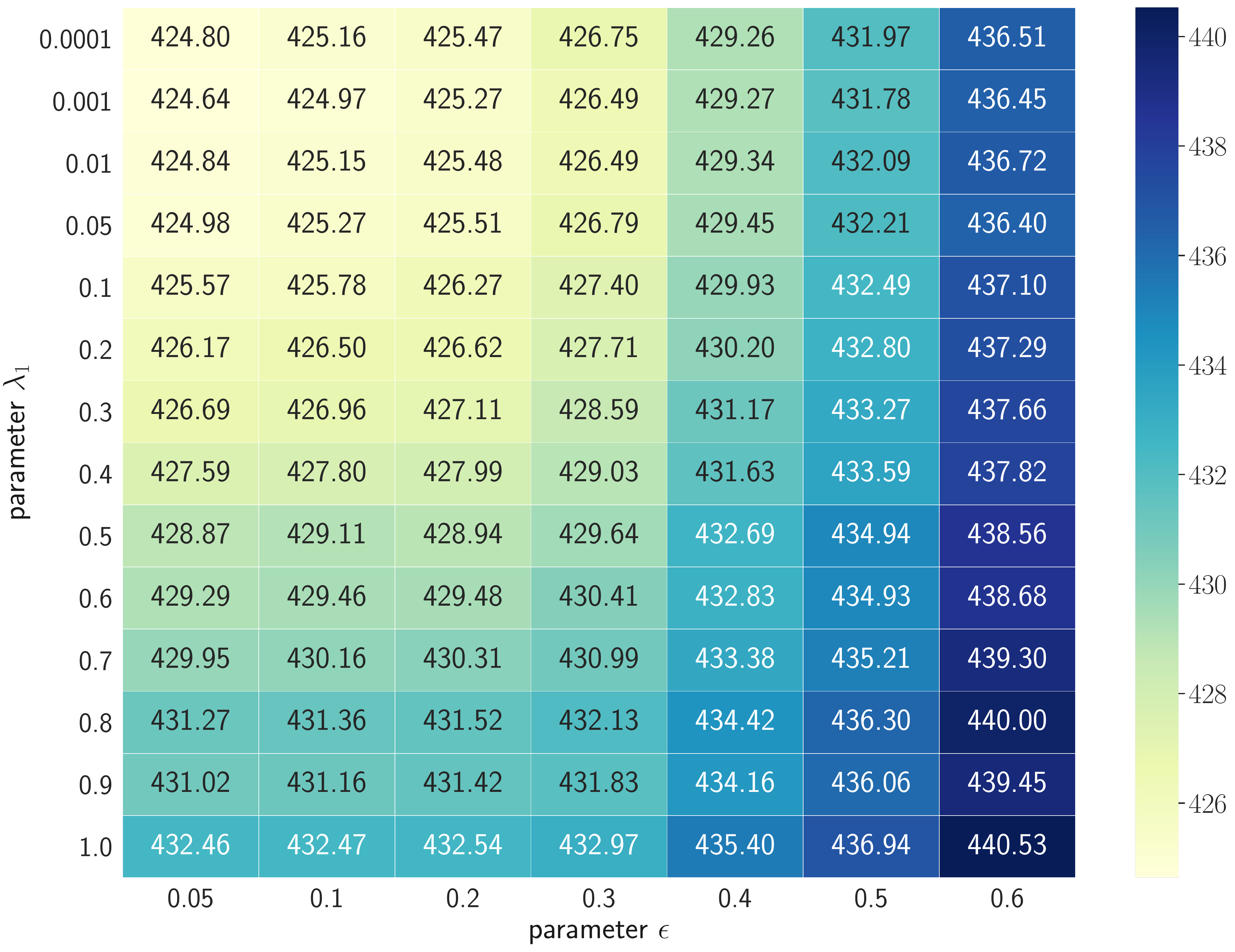}
\caption{The Heatmap of cross validation negative likelihoods for the fMRI hippocampus data over a range of parameter values $\lambda_1$ and $\epsilon$.}
\label{fig:para_fMRI}
\end{figure}

This fMRI hippocampus dataset~\cite{poldrack2015long} consisted of resting-state signals from six brain regions of an individual, which was collected from the same individual for 84 successive days. We used the anatomical connectivity between regions as a reference~\cite{bird2008hippocampus,ghassami2018multi}. Though this reference is cyclic, it was used to evaluate the estimated structures. Before employing our LiNA method for the estimation, we first tested whether this data encounter the multicollinearity problem or not. That is, the VIFs for brain regions are 1.696336, 2.145155, 2.418151, 1.157435, 1.375819, and 1.485547, respectively, which are all lower than 10. It means such data have no multicollinearity problems. Subsequently, we selected the parameter values $\lambda_1$ and $\epsilon$ through cross validation. Similarly, we also used 10-fold cross validation to select the regularization parameters $\lambda_1$ and $\epsilon$. The resulting heatmap is demonstrated in Figure~\ref{fig:para_fMRI}. We found that the performances decline with larger $\epsilon$. On the other hand, with decreasing $\lambda_1$, performances tend to improve. As such, we set $\lambda_1$ to be 0.001 and $\epsilon$ to be 0.05 for this dataset.

We utilized the single-domain LiNA method to estimate latent structures among six brain regions, compared with the NICA method. In particular, we performed experiments for 84 trials, which correspond to 84 days. If an edge was discovered in more than 50\% of trials, we took it into the final structure eventually.

Besides, we found our results also coincide with some current findings, e.g., the relation CA3 $\to$ CA1 is declared connective in neuroscience~\cite{song2015hippocampal}; PRC $\to$ ERC and PHC $\to$ ERC are usually related to episodic memories, which suggest distinct roles in memory formation and retrieval~\cite{kuhn2018temporal}; and ERC $\to$ CA1 is supposed to correlate with memory loss~\cite{kerchner2012hippocampal}.

{\small
\bibliographystyle{named}
\bibliography{ijcai21}
}

\end{document}